\documentclass[10pt,journal,compsoc]{IEEEtran}

\usepackage{graphicx} % for pdf, bitmapped graphics files
\usepackage{subfig}
\usepackage{epsfig} % for postscript graphics files
\usepackage{amsmath} % assumes amsmath package installed
\usepackage{amssymb}  % assumes amsmath package installed
\usepackage{booktabs}
\usepackage{placeins}
\usepackage{multicol}
\usepackage{multirow}
\usepackage{graphicx}
\usepackage{textcomp}
\usepackage[dvipsnames]{xcolor}
\usepackage{textcase}
\usepackage[tablename=TABLE]{caption}

\usepackage{enumitem}
\usepackage{diagbox}
\usepackage{caption}
\usepackage{float}
\usepackage[breaklinks=true,letterpaper=true,colorlinks,bookmarks=false]{hyperref}
\usepackage{packages/varmathbb}
\usepackage[font=small,labelfont=bf]{caption}
\usepackage{algorithm}
\usepackage{algpseudocode}
\usepackage{xspace}

\makeatletter
\DeclareRobustCommand\onedot{\futurelet\@let@token\@onedot}
\def\@onedot{\ifx\@let@token.\else.\null\fi\xspace}
\def\eg{e.g\onedot} 
\def\ie{i.e\onedot}

\def\wrt{wrt\onedot}

\makeatother

\renewcommand{\eqref}[1]{Eq.~\ref{#1}}

\newcommand{\boldparagraph}[1]{\vspace{0.2cm}\noindent{\bf #1:} }

\newif\ifcomment
\commenttrue
\ifcomment
	\newcommand{\yl}[1]{ \noindent {\color{cyan} {\bf Yiyi:} {#1}} }
	
\else
	\newcommand{\yl}[1]{}
\fi

\ifcomment
	\newcommand{\jc}[1]{ \noindent {\color{magenta} {\bf Jessy:} {#1}} }
\else
	\newcommand{\jc}[1]{}
\fi

\usepackage[dvipsnames]{xcolor}
\definecolor{armygreen}{rgb}{0.0, 0.26, 0.15}

\begin{document}

% Macros
\title{DPCN++: Differentiable Phase Correlation Network for Versatile Pose Registration}

\author{Zexi~Chen, 
        Yiyi Liao,
        Haozhe~Du, 
        Haodong~Zhang,
        Xuecheng~Xu,
        Haojian~Lu,
        Rong~Xiong,
        and~Yue~Wang% <-this % stops a space
% \thanks{All authors are with the State Key Laboratory of Industrial Control Technology and Institute of Cyber-Systems and Control, Zhejiang University, Zhejiang, China. Yue Wang is the corresponding author {\tt\small wangyue@iipc.zju.edu.cn}.}
}



\IEEEtitleabstractindextext{%
\parbox{0.918\textwidth}{
\begin{abstract}
Pose registration is critical in vision and robotics. This paper focuses on the challenging task of initialization-free pose registration up to 7DoF for homogeneous and heterogeneous measurements. While recent learning-based methods show promise using differentiable solvers, they either rely on heuristically defined correspondences or are prone to local minima. We present a differentiable phase correlation (DPC) solver that is globally convergent and correspondence-free. When combined with simple feature extraction networks, our general framework DPCN++ allows for versatile pose registration with arbitrary initialization. Specifically, the feature extraction networks first learn dense feature grids from a pair of homogeneous/heterogeneous measurements. These feature grids are then transformed into a translation and scale invariant spectrum representation based on Fourier transform and spherical radial aggregation, decoupling translation and scale from rotation. Next, the rotation, scale, and translation are independently and efficiently estimated in the spectrum step-by-step using the DPC solver. The entire pipeline is differentiable and trained end-to-end. We evaluate DCPN++ on a wide range of registration tasks taking different input modalities, including 2D bird's-eye view images, 3D object and scene measurements, and medical images. Experimental results demonstrate that DCPN++ outperforms both classical and learning-based baselines, especially on partially observed and heterogeneous measurements.
\end{abstract}
}
}

\maketitle

% \fi
\IEEEpeerreviewmaketitle

\section{Introduction}
\label{sec:Introduction}

\IEEEPARstart{P}{ose} registration aims to estimate the relative pose given a pair of measurements. It stems as a core competence for numerous applications, including object pose estimation~\cite{wang2019densefusion}, scene reconstruction~\cite{pan2020gem}, localization~\cite{Pan2019}, and medical imaging~\cite{zhong2015adaptive}.
Pose registration remains a challenging task in the initialization-free setting, in particular for partially observed or heterogeneous measurements.
%\yl{mention challenges, multi-modal?}
%\jc{}
%

Recently, learning-based pose registration methods have shown promise in addressing these challenges. 
%Recent progresses on deep learning empower the pose registration through providing more expressive features that leverage the accuracy and convergence. \yl{what are the challenges of the pose registration? what should be learned in pose registration?}
The pioneering work directly learns pose regression~\cite{regression_tmi}, however, the lack of interpretability hinders generalization. More recent approaches incorporate a differentiable and explicit solver into deep neural networks. This allows for enhancing the interpretability by splitting the task into two stages, trainable feature extraction and classical pose solving, while being able to be trained end-to-end. Despite usually being parameter-free, the pose solver is a key component as it provides gradients for the feature extraction network. In this work, we investigate a key question, \textit{what is an ideal solver for learning-based pose registration?}

%The pioneer of deep pose registration is intuitively the direct regression~\cite{regression_tmi} with neural networks, which is not inductive in architecture design. The lack of interpretability results in unsatisfying performance in generalization. More recent deep pose registration can be divided to two lines: i) correspondence-based methods and correspondence free methods. 
%\red{feature extraction + a differentiable classical solver. Key question: what is a good solver?}

Existing learning-based methods leverage two types of solvers depending on whether they rely on explicit correspondences. 
Correspondence-based solvers, e.g., an SVD solver, infer the pose based on a set of matched points~\cite{wang2019deep,yuan2020deepgmr, zeng20173dmatch}. This line of approaches can obtain a global optimum given good correspondences. However, it requires the network to explicitly learn intermediate (typically redundant) keypoints and their matching relationships, thereby forcing the network to address a higher-dimensional problem than the pose registration itself. Moreover, it heavily relies on outlier elimination to obtain robust pose registration, and the network can be misguided when the solver fails given bad correspondences.

%exclusively utilize deep learning for feature extraction and matching than designing a differentiable SVD solver. While providing better performance, they explicitly introduce intermediate keypoint representations which are obscure in definition, and thus inappropriate for direct supervision. We consider such feature matching actually increase the problem dimension since they call for multiple extra assumptions, such as the number or the distribution of feature points, and further outlier elimination.

\begin{figure*}[t]
\vspace{-15pt}
        \centering
            \includegraphics[width=\linewidth]{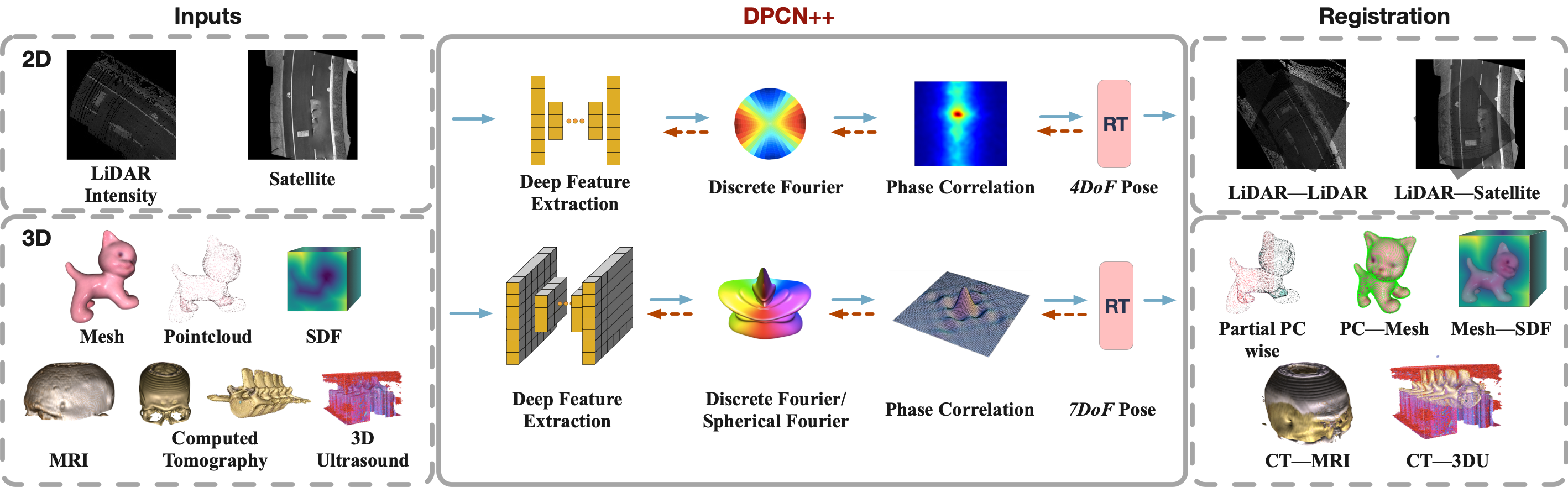}
        \caption{A general registration framework that can solve from 2D registration to 3D registration without requiring any initial guess. \textbf{Left:} Examples of input that can be pose estimated by DPCN++. \textbf{Middle:} We reformulate the registration problem as grid matching by rasterizing different inputs into either 2D or 3D grids. Considering the heterogeneous information contained in the two rasterized grids, the equivalent features are further extracted by trainable extractors, e.g. UNets. If well trained, the pose ($4DoF$ in 2D case or $7DoF$ in 3D) between such domain-aligned features is estimated by the later on differentiable solver, phase correlation. The estimated pose is then supervised by the ground truth and the gradient of which is back-propagated to the feature extractors for better performance. \textbf{Right:} Demonstration of the pose-aligned results.}
        \label{fig:teaser}
%\vspace{-20pt}
\end{figure*}

Another line of work avoids intensive feature matching by leveraging correspondence-free solvers.
Given a pair of extracted features instead of explicit correspondences, this type of solver iteratively updates the pose to maximize the feature similarity along the gradients~\cite{aoki2019pointnetlk}. This gradient-based solver requires a good initialization and easily gets stuck in the local minimum. The suboptimal solution can also mislead the feature extraction network.
%They~\cite{aoki2019pointnetlk,wang2019deep} exploit the existed dimension of the pose registration without referring to additional intermediate feature representations by introducing gradient descent into the deep neural networks. The main fall back of this line of methods is the usage of the gradient based solver, which is local optimal. Such optimal may mislead the feature extraction. Consequently, these methods do have better convergence basin but still, highly dependent on good initialization.

%This leads to a key question, \textit{what is an ideal solver for learning-based pose registration?}
Analyzing these two lines of work, we identify two key properties of an ideal solver for learning-based pose registration:
%\textit{being global convergence} and allowing for \textit{simplifying feature extraction}: 
i) It should avoid explicit correspondence, thus relieving the burden of learning heuristically defined features and preventing outlier elimination. ii) It should avoid local minimum solutions and ideally reach the global optimum in a non-iterative manner, hence providing direct supervision to the feature extraction network towards the best outcome.  
%i) A non-iterative global convergent solver is prone to local minima and does not rely on initialization; ii) Being correspondence-free simplifies the design of feature extraction from explicitly defining feature points, lines, or edges. 
In this paper, we propose to use the classical phase correlation~\cite{bulow2018scale}, \cite{SrinivasaReddy1996} as a differentiable solver that satisfies both criteria, \ie, being correspondence-free and globally convergent. When combined with simple feature extraction networks, our deep phase correlation network 
%In this paper, based on the findings of the two lines of methods for deep pose registration, we set to address the issues by a correspondence-free yet global convergent pose registration network with deep feature extractors, namely, DPCN++. With a differentiable and interpretable non-iterative global solver serves as the central insight, DPCN++ is also embraced by the advantages of the two aspects: i) being correspondence-free simplifies the design of feature extraction from explicitly defining feature points, lines, or edges, and ii) the interpretable non-iterative global convergent solver simplifies the feature training from supervising the progressive iterative results. With these in mind, we design a deep pose registration framework that: 
can perform global registration without any initialization and generalizes well on unseen objects, achieving superior performance compared to existing learned-based methods. In contrast to the classical phase correlation, our learning-based approach can estimate the relative pose given partially observed measurements with fewer overlaps as well as heterogeneous measurements, e.g., CT and MRI. Moreover, we provide a general pose registration framework for both, bird's eye view (BEV) 2D images and 3D measurements.
%solve pose registration not only on homogeneous orthogonal 2D representations and 3D representations, but also on heterogeneous measurements. 
%
%\yl{Lastly, it estimates pose registration in real-time., better place for this sentence? need it??} 

Fig.~\ref{fig:teaser} illustrates our general pose registration framework by leveraging a simple deep feature extraction network and a differentiable correspondence-free global solver. Specifically, we reformulate the pose registration problem as feature grid matching. Given a pair of homogeneous/heterogeneous measurements, we adopt feature extraction networks to learn the dense feature grids, respectively. At the core of our method is a correspondence-free, differentiable solver that decouples the estimation of rotation, translation, and scale. Based on the feature grids, we build the translation and scale-invariant spectrum representation to decouple the translation and scale from rotation, using Fourier transform and spherical radial aggregation. Next, the rotation, scale and translation are efficiently estimated in the spectrum step-by-step independently. In each step, the estimator is built upon the whole sub-solution space (translation, scale or rotation) correlation, thus is globally convergent. By modeling the estimator in a probabilistic manner, our solver is fully differentiable and allows for updating the feature extraction network from pose loss end to end. We summarize our contributions as follows:
\begin{itemize}

    \item Our core contribution is to incorporate a globally convergent, correspondence-free, up to $7$ degree of freedom ($7DoF$) phase correlation solver into learning-based pose registration. This allows for learning a simple feature matching network guided by pose estimation error instead of manually designed correspondences.
    %Deriving from Fourier Transform based phase correlation, the network is interpretable step by step, from the decoupling of translation, rotation and scale, to the phase correlation. With the interpretability and differentiability of the network, we leave network to learn equivalent features of the inputs instead of learning a direct pose regression. The learning process is accomplished by directly back propagating the pose error all the way to the feature extractors knowing that their perfect state is to obtains the common features that can be recognized by the correspondence-free and global convergent solver, i.e. phase correlation.

    \item Our framework is generally applicable to both 3D measurements and gravity aligned BEV images. Moreover, our method can serve for inputs across different modalities by simply leveraging two different feature extraction networks.
    %This can be achieved by adopting Cartesian Fourier transform in 2D log-polar coordinates and $SO(3)$ Fourier transform in spherical coordinates.
    
    \item We validate the effectiveness of our method via extensive experiments. Our method achieves superior performance compared to both classical and learning-based baselines, especially on partially observed or heterogeneous measurements.
    
    %\item Moreover, our method can serve for inputs across different modalities by simply leveraging two different feature extraction networks.\yl{update?}
    %To be precise, DPCN++ is able to i) register $7$ Degrees of Freedom ($DoF$) poses in the 3D case including $3Dof$ translation, $3Dof$ rotation and $1Dof$ scale, and ii) register $4DoF$ poses in the 2D case, including $2DoF$ translation, $1Dof$ rotation and $1DoF$ scale.
    
   % \item The framework can serve across modalities. By allowing the end-to-end training of the pose registration, DPCN++ is able to learn to register inputs of different modalities without explicitly supervising the feature extractors. Each modality has a corresponding feature extractor after the training so that different modalities are domain aligned after the extraction, which can be further registered by the phase correlation. Moreover, benefit from the aspect that it is the common domain of features that it is learning, the network performs well in generalization.

   %\item DPCN++ is customization friendly for future development. It is compatible with multiple feature extractors, e.g. UNet2D, UNet3D, Pointnet. With the flexibility in autoencoder plugins, the framework is scalable and is easy to develope base on the specific tasks later on.
    
    \item We release a multi-modal Aero-Ground Dataset, intending to facilitate cooperative localization between ground mobile robots, micro aerial vehicles (MAVs) and satellites.
    %, which potentially aids the future researches for aero-ground collaboration and heterogeneous image matching.
\end{itemize}
%only needs to learn matched feature grids instead of manually designed heuristic features.
%leave the network to learn features directly guided by the pose registration error back-propogated through the correspondence-free and global convergent solver. 

%\textbf{\textit{Contribution:}} To the best of our knowledge, we proposed the first general heterogeneous registration framework by leveraging the deep feature extractors and the differentiable correspondence-free global solver as in Fig.~\ref{fig:teaser}. We reformulate the heterogeneous registration problem as grid matching and training feature extracting functions that eliminate the modal differences between the inputs, and the output of which should be registered with the later on global optimal solver.
%\textbf{\textit{Contribution:}} 
%Our method is applicable to  both homogeneous and heterogeneous measurements.
%We reformulate the heterogeneous registration problem as grid matching and training feature extracting functions that eliminate the modal differences between the inputs, and the output of which should be registered with the later on global optimal solver.
%\textbf{Novelty with respect to \cite{chen2020deep}:} 
Our work is an extension of a conference paper \cite{chen2020deep},  where we introduce the Differentiable Phase Correlation Network (DPCN) that solves the pose registration problem between two gravity aligned BEV images. In this paper, we present a general framework, DPCN++, that solves homogeneous and heterogeneous pose registration in both 2D and 3D space, extending DPCN in terms of both method and experiments. Specifically, 1) we extend the registration dimension from 2D to 3D by leveraging spherical and $SO(3)$ Fourier transforms, leading to a framework capable of solving pose registration for both 2D BEV images pairs and 3D representation pairs. 2) We conduct extensive experiments for versatile 3D-3D pose registration tasks on both synthetic and real-world datasets. 
%In summary, DPCN++ includes major improvements both in terms of method and experiments.

\section{Related Works}
\label{sec:Relate Works}
 
%In this section, we review existing works on 
%Previous researches on measurement registration lay saperately on
%2D image matching and 3D representation matching, respectively.

\subsection{3D-3D Homogeneous Registration}
\label{subsec:3D-3D Registration}
3D-3D pose registration based on homogeneous measurements is one of the most studied scenarios in pose registration. We categorize previous works on this problem into direct regression, correspondence-based and correspondence-free methods.

\boldparagraph{Direct Regression Methods}To guide the feature learning by a differentiable pose regression solver in an end-to-end manner, some methods learn to directly regress the pose by a fully connected network. AlignNet \cite{gross2019alignnet} utilizes a MLP layer for alignment estimation from the global feature. PCRNet \cite{sarode2019pcrnet} introduces a canonical pose as an intermediate stage to reduce the space for direct pose regression. 3DRegNet \cite{pais20203dregnet} follows a similar way but utilizes correspondence as the input. These methods reveal the trend of end-to-end learning. One drawback is that the output pose is directly regressed by neural networks instead of being estimated by an explicit solver, so the lack of interpretability usually leads to weak generalization.

\boldparagraph{Correspondence-Based Methods}The correspondence-based registration method originates from the well-known iterative closest point (ICP) \cite{besl1992method}. Many follow-up variants are proposed \cite{rusinkiewicz2001efficient}, \cite{rusinkiewicz2019symmetric}, \cite{chetverikov2005robust}, \cite{granger2002multi}. The solver of the point-to-point ICP is global convergent. The main limitation is the nearest neighbor based correspondence strategy, which highly depends on the initialization. Therefore, learning is employed to improve the feature extraction and feature matching. DCP \cite{wang2019deep} incorporates DGCNN \cite{wang2019dynamic} for point cloud embedding and an attention-based module for feature matching, followed by a differentiable SVD solver for an end-to-end pose estimation architecture. DGR \cite{choy2020deep} uses fully convolutional geometric features \cite{choy2019fully} for feature extraction and applies 6-dimensional segmentation for correspondence prediction. DeepGMR \cite{yuan2020deepgmr} learns to find pose-invariant correspondences between Gaussian mixture models that approximate the shape, and compute the transformation based on the model parameters, so that the noise in the raw point cloud can be suppressed. The main limitation of correspondence-based learning methods is the robustness of the SVD solver. It enforces feature matching to be almost outlier free, which is unfortunately hard to achieve for the feature network. 

% For example, RPM-Net \cite{yew2020rpm} utilizes a feature extraction network to learn hybrid features containing spatial and local geometry information, and then leverages Sinkhorn layers and annealing to get point correspondences. 
Concerning the problem mentioned above, another line of efforts has been made for solver improvement. Go-ICP \cite{yang2015go} utilizes a Branch-and-Bound (BnB) method to search the whole SE(3) space to get a globally optimal solution. FGR \cite{zhou2016fast} uses second-order optimization and applies a Geman-McClure cost function to reach global registration of high accuracy. DGR \cite{choy2020deep} proposes an outlier robust registration method as post-processing to finetune the result yielded by the learning stage. TEASER \cite{yang2020teaser} is the first certifiable registration algorithm dealing with a large percentage of outliers, of which the main idea is the decoupling of scale, translation and rotation. However, these solvers are usually not differentiable, making it hard to guide the features and matching in an end-to-end manner.

\boldparagraph{Correspondence-Free Methods}The core idea of correspondence free methods is to estimate the pose based on the similarity of features between two measurements. According to the convergence of solver, we further categorize this line of methods into locally convergent and globally convergent ones.

The locally convergent correspondence-free pose registration is inspired by the optical flow in 2D image registration. A pioneering work, PointNetLK \cite{aoki2019pointnetlk}, extracts global features using PointNet \cite{qi2017pointnet} and extends Lucas-Kanade algorithm \cite{lucas1981iterative} to feature space for iterative estimation of the pose. Li et al. \cite{li2021pointnetlk} follow the idea of PointNetLK and propose an analytical form of Jacobian to improve the performance in mismatched conditions. Huang et al. \cite{huang2020feature} further add a feature-metric loss to the PointNetLK framework as a side task, which directly enforces the feature similarity between the aligned measurements. One of the limitations of this line of methods is the iterative solver, which is sensitive to the initialization, and the local minima may mislead the feature learning.

The globally convergent correspondence-free method mainly employs the idea of phase correlation. B{\"u}low et al. \cite{bulow2018scale} and PHASER \cite{bernreiter2021phaser} both utilize spherical and spatial Fourier transforms to estimate the relative pose using correlation \cite{wang2012local} in the spectrum. The global convergence lies in the correlation, which is an intrinsically exhaustive search, but can be evaluated efficiently via decoupling in the spectrum. Inspired by these methods, we introduce a differentiable version of phase correlation to enable end-to-end learning based on a globally convergent solver. In contrast to \cite{bulow2018scale,bernreiter2021phaser}, our framework achieves better registration performance and is applicable to versatile pose registration tasks by learning from data. Recently, with the progress of geometric deep learning, Zhu et al. \cite{zhu2022correspondence} apply SO(3)-equivariance embedding for feature learning. The rotation is globally estimated in the feature space, skipping the stage of making correspondence. However, this method may fail when the relative pose has both relative rotation and translation.

\subsection{3D-3D Heterogeneous Registration}
Multi-modal data may vary in the data structure, physical and anatomical principles, intensity and noise \cite{saiti2020application}, which makes the registration task much more challenging. Heterogeneous pose registration tasks mainly come from medical image analysis. In order to register heterogeneous measurements, one line of works \cite{chen2009hybrid}, \cite{chen2008multimodal}, \cite{schwab2015multimodal} employs handcrafted mutual information as a similarity measure \cite{wells1996multi} and utilizes bio-inspired optimization algorithm for minimization, like Particle Swarm Optimization (PSO). Kisaki et al. \cite{kisaki2014high} accelerate the registration by the Levenberg-Marquardt optimization algorithm, which, however, depends on the initial value. Several algorithms also use the Branch-and-Bound framework to overcome the randomness of the bio-inspired optimization algorithms \cite{enqvist2009optimal}, \cite{pan2019multi}. Yang et al. \cite{yang2021certifiable} propose a general framework for certifiable robust geometric perception, which exploits the registration between the 3D mesh model and point clouds. As the efforts for solver improvement in homogeneous pose registration, these solvers show good performance when the feature is correctly designed, but they are not differentiable, thus cannot be applied to guide the automatic feature learning.

Deep learning is also introduced to solve the heterogeneous registration, mainly for learning the similarity metric or estimating transformation directly. Lee et al. \cite{lee2009learning} reformulate the problem of learning similarity measures to a binary classification of aligned and misaligned patches, tested on CT-MRI and PET-MRI volumes. CNN/RNN based methods \cite{haskins2019learning}, \cite{sedghi2018semi}, \cite{wright2018lstm} are proposed to learn local similarity metrics for better registration. Reinforcement learning is utilized in \cite{liao2017artificial}, \cite{ma2017multimodal} to learn the pose regression directly, which can be trained in an end-to-end manner, but may have a weaker generalization due to the lack of the explicit solver. Cao et al. \cite{cao2018deep} employ the network to warp the image in a non-rigid way for medical image registration. These methods show the advantages of the learning-based methods, especially in learning the feature space across heterogeneous modals. One of their limitations is the dependency on the initialization, which constrains the convergence basin of the learned similarity and warping strategy.

\subsection{2D-2D Registration}
\label{subsec:2D-2D Registration}

The trend in 2D-2D registration is similar to 3D-3D registration. Early methods employ hand-crafted image feature point matching to generate correspondences, like SIFT \cite{lowe2004sift}, which is then fed to the solver for pose registration. To deal with large appearance change in two images, learning-based methods are applied for feature detection and matching, like SuperPoint \cite{superpoint}, R2D2 \cite{revaud2019r2d2} and D2Net \cite{dusmanu2019d2} which shows good performance in visual localization \cite{zhang2021reference}. To avoid the ground truth correspondences, direct regression is connected to the feature network in an end-to-end way in DAM \cite{Park2020}. However, as in 3D-3D registration, such solvers may have weaker generalization performance due to the lack of explicit constraints. Currently, a common practice in image registration is to employ a robust solver as a post-processing step \cite{jiao2021r2d2loc}. 

General image registration shown above deals with $6DoF$ pose. In this paper, we focus on the $4DoF$ registration problem for BEV images. It can be solved by the correspondence-based methods above, but also correspondence-free methods. To build robust similarity under appearance change for images, Kaslin et al. \cite{Kaslin2016} apply normalized cross-correlation (NCC) on geometric measurements, and WNCC \cite{xu2020collaborative} recursively rotates the source in a small range to match the rotation. Kummerle et al. \cite{Kummerle2010} and Ruchti et al. \cite{Ruchti2015} utilize hand-craft features to localize LiDAR against satellite maps. To learn the feature instead of hand-crafted ones, Kim et al. \cite{Kim2018} utilize feature maps trained from other tasks to measure the similarity. However, it is unclear how to design tasks to improve the feature learning. More recently, Lu et al. \cite{Lu2019} and Tang et al. \cite{Tang2020} propose to learn the embedding with a differentiable exhaustive search solver in the discretized solution space, and show good results. But due to the expensive search, the efficiency is low, and the discretization is only applied in the local range. To improve the search efficiency, Barnes et al. \cite{Barnes2019} propose to evaluate the similarity in the frequency spectrum. To further eliminate the constraint of local range, phase correlation (PC) \cite{SrinivasaReddy1996}, which searches the global optimal pose in the whole solution space by reformulating the problem into the spectrum, is investigated in the context of end-to-end deep learning for homogeneous \cite{weston2022fast} and heterogeneous pose registration \cite{chen2020deep}. These works show promises of introducing a global convergent and a differentiable solver, which guides the feature learning towards the best outcome. However, both methods can only be applied to 2D. In contrast, we leverage the classic phase correlation \cite{bulow2018scale} \cite{SrinivasaReddy1996}, combine them with deep learning, and propose a general framework for versatile pose registration that is applicable to both 3D-3D and 2D-2D pose registration.

% By revealing the problems encountered by prior works, one can think of an ideal matcher that can obtain the solution without exhaustive evaluation and have good interpretability and generalization. In our work, we set to propose such a learnable matcher, of which the essence is a differentiable phase correlation. Phase correlation \cite{SrinivasaReddy1996} is a similarity-based matcher that performs well for inputs with the same modality but only tolerate small high frequency noise. We modify the phase correlation into a differentiable manner and embed it into our end-to-end framework. This architecture allows our system to find optimal feature extractors with respect to the resultant pose of image matching. Specifically, we adopt the conventional phase correlation pipeline proposed by \cite{SrinivasaReddy1996} and explicitly endow the Discrete Fourier Transform (DFT) layer, log-polar transformation layer (LPT), and differentiable correlation layer (DC) with differentiability and thus make it trainable for our end-to-end matching network as shown in Figure \ref{fig:demo net}. Our experiments show the robustness and efficiency of the proposed method on matching heterogeneous sensor measurements.

\section{Overview}
\label{sec:overview}

Given two measurements with homogeneous or heterogeneous modality, \eg, point cloud and mesh, our goal is to estimate their relative pose up to $7DoF$, including rotation, scale and translation, without referring to an initial value. In this work, we first derive the method for $7DoF$ 3D-3D pose registration and consider the $4DoF$ orthogonal 2D-2D pose registration as its degenerated version. We now formally state the problem in sequel.

%can be utilized as a global registration method both in 3D grids and in 2D images. However, we wish it to register more, beyond grids. For instance, we wish it to register between partial point clouds and between heterogeneous measurements. To endow the method with more versatile capability, we introduce deep feature extractors to the phase correlation who eventually becomes deep phase correlation.

%Nevertheless, it is hard to call for a direct supervision for feature extraction in heterogeneous measurements and that we can not to supervise the features at the fingertips. As we introduce the deep features for a better pose registration, it is intuitive and will be effective if we can supervise the feature extraction indirectly from the relative poses between measurements. In order to achieve this, phase correlation in Section~\ref{sec:PCforReg} can be treated as block of solver who is able to properly back-propagate errors to train the feature extractors if the gradient is well designed.

\subsection{Problem Statement}

Let $v_{1}$ and $v_{2}$ denote two measurements linked by an unknown relative pose $\mathbf{T}=\{\boldsymbol{t}, \boldsymbol{r}, \mu\}$ with translation $\boldsymbol{t}\in \mathfrak{R}^{3}$, rotation $\boldsymbol{r}\in SO(3)$, and isotropic scale $\mu$. A general end-to-end learning-based pose registration problem can be stated as
\begin{equation}
    \min_{\theta_1,\theta_2} \| \mathbf{T}^*-\arg\max_{\mathbf{T}} \mathfrak{C}(Q_{\theta_1}(v_1), Q_{\theta_2}(\Omega_{\mathbf{T}}v_2)) \|
    \label{eq:problem_define_raw}
\end{equation}
where $Q_{\theta_1}$ and $Q_{\theta_2}$ are two deep networks extracting features from the inputs, such as point clouds, meshes and signed distant fields (SDF), $\Omega_{\mathbf{T}}$ is the pose transformation applied to the input with proper operation generated by $\mathbf{T}$, $\mathbf{T}^*$ is the ground truth, $\mathfrak{C}$ is a scoring function measure the similarity. The key challenge is the design of a differentiable and global convergent $\arg\max\mathfrak{C}(\cdot,\cdot)$, so that the gradient of $Q_{\theta_1}$ and $Q_{\theta_2}$ can be correctly derived for learning. 

\begin{figure}[t]
\centering
\includegraphics[width=\linewidth]{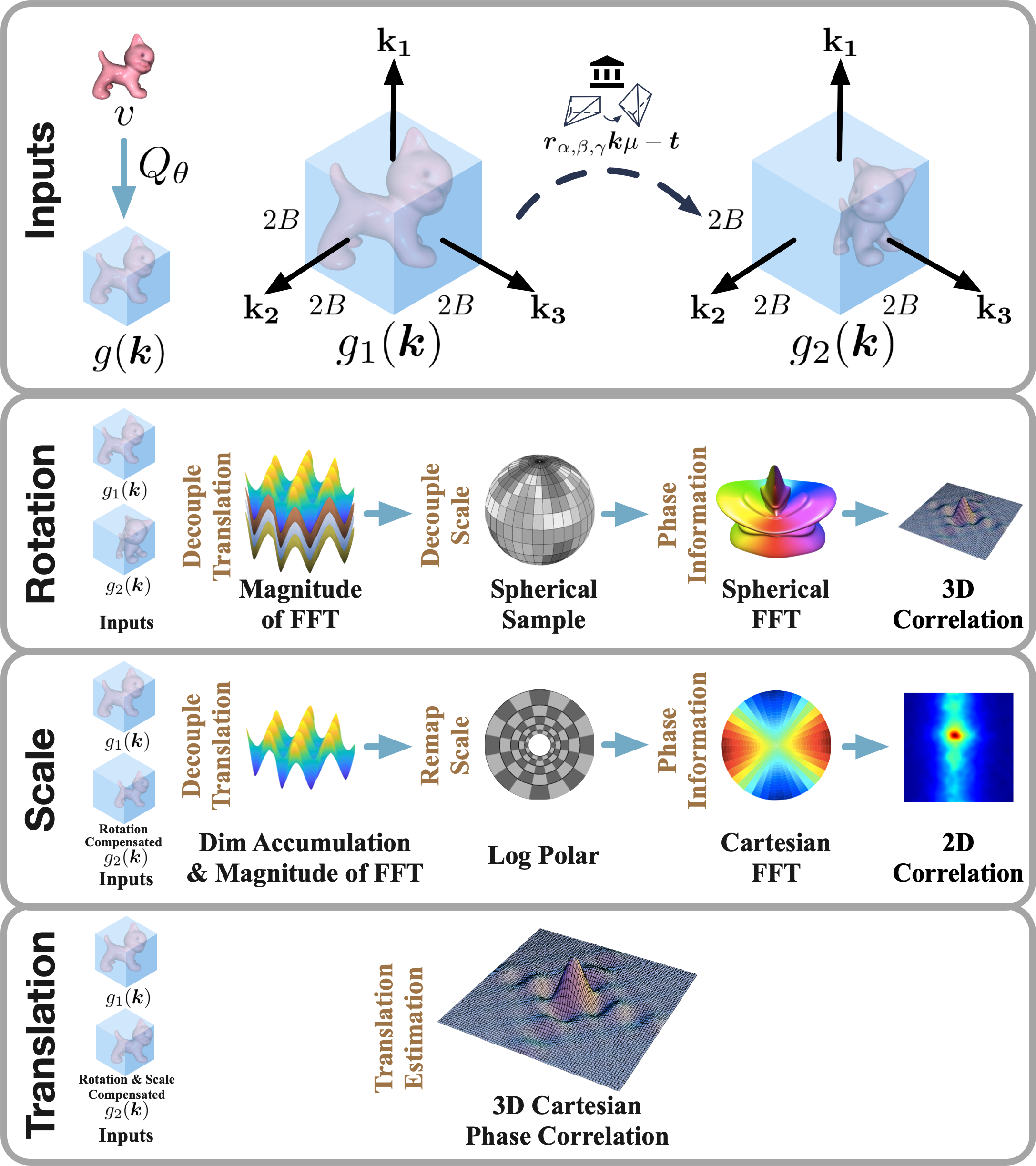}
\caption{The \textbf{overall pipeline} of the DPC on grids. \textbf{Rotation:} The translation and scale are decoupled from the rotation once we focus on the magnitudes of the corresponding Fourier transform and project them to the sphere. Then the $SO(3)$ rotation on the sphere is estimated with spherical phase correlation. \textbf{Scale:} The source is rotation-compensated with the target considering the estimated rotation. Then, both the source and the target are accumulated to 2D and log-polar transformed. With the log-polar transformation, the scale are remapped to the horizontal part of the accumulated images and is further estimated with the 2D Cartesian phase correlation. \textbf{Translation:} Once the source and the target are rotation and scale compensated, the corresponding translation is estimated with the 3D Cartesian phase correlation.}
\label{fig:coordinate define}
\end{figure}

In correspondence-based methods, $\arg\max\mathfrak{C}(\cdot,\cdot)$ is substituted with Euclidean distance between correspondences and closed-form SVD solver, thus differentiable and global convergent. But these methods assume that the correspondence yielded by $Q_{\theta_1}$ and $Q_{\theta_2}$ is perfect, which may be difficult in practice, especially for heterogeneous inputs. In correspondence-free methods, $\arg\max\mathfrak{C}(\cdot,\cdot)$ is approximated by a numerical iterative process, which is differentiable, but the gradient based iteration may mislead the learning of $Q_{\theta_1}$ and $Q_{\theta_2}$ due to the local minimums, thus highly depends on an initial value in both training and testing stage.

\subsection{Method Overview}

Our method, DPCN++, also consists of a trainable feature extraction network and a differentiable pose registration solver. To address the challenge, We follow the classic idea of correspondence-free phase correlation in \cite{bulow2018scale}, and present a differentiable phase correlation solver as $\arg\max\mathfrak{C}(\cdot,\cdot)$.

Specifically, as shown in Fig.~\ref{fig:coordinate define}, after applying feature extraction $Q_{\theta}$, we represent the inputs in feature grids with axes $\{\mathbf{k_{1}},\mathbf{k_{2}},\mathbf{k_{3}}\}$, and indexed by  $\boldsymbol{k}\in [-B,B-1]^{3}$, where $B$ is the bandwidth: 
\begin{gather}
    g_{1}(\boldsymbol{k}) = Q_{\theta_1}(v_{1}) \\
    g_{2}(\boldsymbol{k}) = Q_{\theta_2}(v_{2}),
    \label{eq:grid_transform}
\end{gather}
Then we feed the feature grids to the differentiable phase correlation solver, denoted as $Q_{DPC}$, to generate rotation, scale and translation, which can be supervised as 
\begin{equation}
    \min_{\theta_1,\theta_2} \| \mathbf{T}^*-Q_{DPC}(g_{1}(\boldsymbol{k}), g_{2}(\Omega_{\mathbf{T}}\boldsymbol{k})) \|.
    \label{eq:problem_define_grid}
\end{equation}
Thanks to the global convergence of differentiable phase correlation, the feature extraction network can be guided by the solver without ambiguity.

\section{Differentiable Phase Correlation}

%\subsection{Decouple of Pose Registration}
%\label{sec:PCforReg}

We begin the presentation of $Q_{DPC}$ from two feature grids $g_{1}$ and $g_{2}$ linked by a relative $7DoF$ pose, we have
\begin{equation}
    g_{1}(\boldsymbol{k}) =  g_{2}(\boldsymbol{r}_{\alpha,\beta,\gamma}\boldsymbol{k}\mu-\boldsymbol{t}),
    \label{eq:g1g2 relation}
\end{equation}
where we employ $Z-Y-Z$ Eular angle $\{\alpha,\beta,\gamma\}$ to parameterize the rotation $\boldsymbol{r}$. To solve the unknown pose, we sequentially decouple the translation, scale from the rotation, and estimate the rotation, scale and translation step-by-step inversely.
 
%The global registration problem can be refromulated as grid matching, where 2D and 3D counterparts are rasterized into 2D and 3D grids. For both dimensions, we address the grid matching by performing a correlation between source and target inputs in the Fourier domain. We seek to estimate the $7DoF$ poses  between 3D measurements, and $4DoF$ poses (translation $\boldsymbol{t}\in \mathbb{R}^{2}$, $\boldsymbol{r}\in (0,2\pi]$, and isotropic scale $\mu$) between 2D measurements. We will first dig into the registration pipeline in 3D cases followed by the elaboration in 2D since the 2D case can be treated as the deterioration in the translation and rotation from the 3D case. 

\boldparagraph{Translation Invariance}Following the common practice that lowercase letters indicate the time domain variables and the uppercases indicate the frequency domain, we have the corresponding 3D Discrete Fourier Transform of the grid $g(\boldsymbol{k})$ as $G(\boldsymbol{j})$:
\begin{equation}
    G(\boldsymbol{j}) = \frac{1}{(2B)^{3}}\sum_{\boldsymbol{k}}g(\boldsymbol{k})e^{-i2\pi (\boldsymbol{j}^{T}\boldsymbol{k})},
    \label{eq:Gg relation}
\end{equation}
where $\boldsymbol{j}$ is the sampled frequency with the axes of $\{\mathbf{j_{1}},\mathbf{j_{2}},\mathbf{j_{3}}\}$. With Eq.~\ref{eq:Gg relation}, the relationship between $g_{1}(\boldsymbol{k})$ and $g_{2}(\boldsymbol{k})$ as in Eq.~\ref{eq:g1g2 relation} can be given in the frequency domain as:
\begin{equation}
    G_{1}(\boldsymbol{j}) = {\mu}^{-3}G_{2}(\boldsymbol{r}_{\alpha,\beta,\gamma}\boldsymbol{j}{\mu}^{-1})e^{i2\pi(\boldsymbol{r}_{\alpha,\beta,\gamma}\boldsymbol{j}\mu^{-1})^{T}\boldsymbol{t}}.
    \label{eq:G1G2 relation}
\end{equation}
By taking the magnitude of the frequency spectrum, we have
\begin{equation}
    |G_{1}(\boldsymbol{j})| = {\mu}^{-3}|G_{2}(\boldsymbol{r}_{\alpha,\beta,\gamma}\boldsymbol{j}{\mu}^{-1})|,
    \label{eq:magG1G21}
\end{equation}
Note that the two magnitude spectrum is invariant to translation $\boldsymbol{t}$, and only depend on the relative rotation $\boldsymbol{r}_{\alpha,\beta,\gamma}$ and scale $\mu$. For brevity, we further denote the magnitude as $\hat{G}(\boldsymbol{j})$:
\begin{equation}
    \hat{G}(\boldsymbol{j}) = |G(\boldsymbol{j})| = |\mathfrak{F}(g(\boldsymbol{k}))|,
    \label{eq:trans_inv_g}
\end{equation}
where $\mathfrak{F}$ is the Fourier transform.

\boldparagraph{Scale Invariance}We further eliminate the scale dependency in $\hat{G}(\boldsymbol{j})$. Specifically, we formulate a spherical function $s(\boldsymbol{\lambda})$ defined on the unit sphere $\boldsymbol{\lambda} \in \varmathbb{S}^{2}$ by projecting each cell of $\hat{G}(\boldsymbol{j})$ to sphere coordinates following
%\begin{equation}
%    \boldsymbol{\lambda} =  [\arctan(\frac{j_{2}}{j_{1}}), \arccos(\frac{j_{3}}{\sqrt{j_{1}^{2}+j_{2}^{2}+j_{3}^{2}}})]
%    \label{eq:define spherical coord}
%\end{equation}
\begin{gather}
    s(\boldsymbol{\lambda})=\sum_{\boldsymbol{j}\in \Gamma_{\boldsymbol{\lambda}}} \hat{G}(\boldsymbol{j})\\
    \Gamma_{\boldsymbol{\lambda}} = \{\boldsymbol{j} | \boldsymbol{\lambda} = [\arctan(\frac{j_{2}}{j_{1}}), \arccos(\frac{j_{3}}{\sqrt{j_{1}^{2}+j_{2}^{2}+j_{3}^{2}}})]\}
    \label{eq:define_spherical_value}
\end{gather}
%\begin{multline}
%    \mathcal{C} : \{\theta = \arctan(\frac{j_{2}}{j_{1}}),\;\;\ \omega = %\arccos(\frac{j_{3}}{\sqrt{j_{1}^{2}+j_{2}^{2}+j_{3}^{2}}})\},
%    \label{eq:project grid to sphere}    
%\end{multline}
where $\boldsymbol{j} = [j_{1},j_{2},j_{3}]$. In spherical coordinates, $\boldsymbol{\boldsymbol{\lambda}}$ actually encodes a ray direction, so all cells passed by the ray are summed as the value of $s(\boldsymbol{\boldsymbol{\lambda}})$.

With radial aggregation, we note that $s(\boldsymbol{\boldsymbol{\lambda}})$ is invariant to scale change, and only depends on the rotation $\boldsymbol{r}_{\alpha,\beta,\gamma}$ as:
%\begin{equation}
%    s(\theta_{i},\omega_{i}) = \sum_{\boldsymbol{j}\models %\mathcal{C}}\hat{G}(\boldsymbol{j}),
%    \label{eq:interpolate sphere}
%\end{equation}
\begin{equation}
    s_{1}(\boldsymbol{\boldsymbol{\lambda}}) =  \mu^{-3}s_{2}(\boldsymbol{\boldsymbol{\lambda}}(\boldsymbol{r}_{\alpha,\beta,\gamma}\boldsymbol{j})).
    \label{eq:interpolate sphere}
\end{equation}

By further normalizing the grids, $\mu^{-3}$ in Eq.~\ref{eq:interpolate sphere} is eliminated with $s_{1}(\boldsymbol{\boldsymbol{\lambda}}) =s_{2}(\boldsymbol{\boldsymbol{\lambda}}(\boldsymbol{r}_{\alpha,\beta,\gamma}\boldsymbol{j}))$. The projected sphere is further resampled following the ``DH-Grid'' sampling theorem by Driscoll and Healy~\cite{healy2003ffts} and create a resampled grid on the sphere with the size of $2B\times2B$. Such sampling of $s(\boldsymbol{\boldsymbol{\lambda}})$ is implemented with interpolation, yielding $\check{s}(\check{\boldsymbol{\boldsymbol{\lambda}}})$ with the spherical coordinate $\check{\boldsymbol{\boldsymbol{\lambda}}}$ mapped from $\boldsymbol{\boldsymbol{\lambda}}$.

%With such projection, scaling is eliminated for all the points on the same spatial radial since they are projected to the same position on the sphere.

\subsection{Rotation Registration}

Based on the analysis above, we arrive at two spherical functions derived from the raw measurements, which is translation and scale invariant. Now we propose a global convergent and differentiable rotation registration solver to align $\check{s}_{1}(\check{\boldsymbol{\boldsymbol{\lambda}}})$ and $\check{s}_{2}(\check{\boldsymbol{\boldsymbol{\lambda}}})$.

\boldparagraph{Spherical Phase Correlation}Given two spherical functions with relative rotation $\boldsymbol{r}_{\alpha,\beta,\gamma}$, we define the correlation on unit sphere as:
 \begin{equation}
    f(\boldsymbol{r}_{\alpha,\beta,\gamma}) =  \int_{\check{\boldsymbol{\lambda}}\in\varmathbb{S}^{2}}\check{s}_{1}(\check{\boldsymbol{\lambda}})\check{s}_{2}(\boldsymbol{r}_{\alpha,\beta,\gamma}\check{\boldsymbol{\lambda}})d\check{\boldsymbol{\lambda}},
    \label{eq:define sphere corr}
\end{equation}
As shown in~\cite{makadia2006rotation}, the $SO(3)$ Fourier transform of $f(\boldsymbol{r}_{\alpha,\beta,\gamma})$, denoted as $F^{l}_{mn}$, can be efficiently evaluated by point-wise multiplication, $\odot$, of the spherical Fourier transform of two spheres as:
 \begin{equation}
    F^{l}_{mn} = \check{S}_{m,1}^{l}\odot\overline{\check{S}_{n,2}^{l}},
    \label{eq:SOFT_SFT}
\end{equation}
where $ \check{S}_{m,1}^{l}$ and $ \check{S}_{n,2}^{l}$ are the spherical Fourier transform of $\check{s}_{1}(\check{\boldsymbol{\lambda}})$ and $\check{s}_{2}(\check{\boldsymbol{\lambda}})$, and $\overline{\check{S}_{n,2}^{l}}$ is the conjugate. In our case, $m=n$. The spherical Fourier transform are defined as:
\begin{equation}
    \check{S}_{m}^{l} =  \frac{\sqrt{2\pi}}{2B}\sum_{\check{\boldsymbol{\lambda}}} a_{m} \check{s}(\check{\boldsymbol{\lambda}})Y^{l}_{m}(\check{\boldsymbol{\lambda}}),
    \label{eq:SFT}
\end{equation}
% \begin{equation}
%     \check{S}_{m}^{l} =  \frac{\sqrt{2\pi}}{2B}\sum_{q=0}^{2B-1}\sum_{p=0}^{2B-1}a_{p}\check{s}(\check{\boldsymbol{\lambda}}_{p,q})Y^{l}_{m}(\check{\boldsymbol{\lambda}}_{p,q}),
%     \label{eq:SFT}
% \end{equation}
%$\check{\boldsymbol{\lambda}}_{p,q}$ is the discrete spherical coordinate of the sphere
where $Y^{l}_{m}$ is the $(2l+1)$ spherical harmonics of degree $l$ and order $m$, $a_{m}$ is the weight to compensate for an oversampling at the pole, and empirically given by \cite{healy2003ffts}. The complete derivation of the spherical Fourier transform can be found in \cite{makadia2006rotation}. Note that in our case, $m=n$ and $l=B$ in Eq.~\ref{eq:SOFT_SFT}. Considering Eq.~\ref{eq:define sphere corr}, Eq.~\ref{eq:SOFT_SFT}, and Eq.~\ref{eq:SFT}, we have:
 \begin{equation}
    \mathfrak{F}_{SO(3)}(f(\boldsymbol{r}_{\alpha,\beta,\gamma}))= \mathfrak{F}_{S^2}(\check{s}_{1}(\check{\boldsymbol{\lambda}}))\odot\overline{\mathfrak{F}_{S^2}(\check{s}_{2}(\boldsymbol{r}_{\alpha,\beta,\gamma}\check{\boldsymbol{\lambda}}))},
    \label{eq:SOFT summary}
\end{equation}
where $\mathfrak{F}_{SO(3)}$ and $\mathfrak{F}_{S^{2}}$ denote for the $SO(3)$ Fourier transform and spherical Fourier transform respectively.

Therefore, by taking the inverse $SO(3)$ Fourier transform, denoted as $\mathfrak{i}\mathfrak{F}_{SO(3)}$ of Eq.~\ref{eq:SOFT summary}, we have the correlation map defined on the $Z-Y-Z$ Euler angle space. The rotation registration can then be solved by searching the index with the highest correlation:
\begin{equation}
 \begin{split}
    [\hat{\alpha},\hat{\beta},\hat{\gamma}] & = \arg\max f( \boldsymbol{r}_{\alpha,\beta,\gamma}) \\
        & = \arg\max\mathfrak{i}\mathfrak{F}_{SO(3)}(\mathfrak{F}_{SO(3)}(f(\boldsymbol{r}_{\alpha,\beta,\gamma})))\\
        %& = \arg\max\mathfrak{i}\mathfrak{F}_{SO(3)}(\check{S}_{m\,1}^{l} \odot \overline{\check{S}_{n\,2}^{l}})\\
        &=\arg\max\mathfrak{i}\mathfrak{F}_{SO(3)}(\mathfrak{F}_{S^2}(\check{s}_{1}(\check{\boldsymbol{\lambda}}))\\
        & \;\;\;\;\;\;\;\;\;\;\;\;\;\;\;\;\;\;\;\;\;\;\;\;\; \odot\overline{\mathfrak{F}_{S^2}(\check{s}_{2}(\boldsymbol{r}_{\alpha,\beta,\gamma}\check{\boldsymbol{\lambda}}))}).
 \end{split}
    \label{eq:phasecorr_G1G2_SO3_final}
\end{equation}
% that is:
%  \begin{equation}
% \begin{split}
%    [\alpha^{*},\beta^{*},\gamma^{*}] &=\arg\max(\mathfrak{i}\mathfrak{F}_{SO(3)}(\mathfrak{F}_{S^2}(\check{s}_{1}(\check{\boldsymbol{\lambda}}))\\
%        & \;\;\;\;\;\;\;\;\;\;\;\;\;\;\;\;\;\; \odot\overline{\mathfrak{F}_{S^2}(\check{s}_{2}(\boldsymbol{r}_{\alpha,\beta,\gamma}\check{\boldsymbol{\lambda}}))})).
% \end{split}
%    \label{eq:phasecorr_G1G2_SO3_final}
% \end{equation}

The rotation registration solver actually evaluates the correlation for all possible Euler angles with the discretization resolution of $\pm\frac{\pi}{(2B)}$ for $\alpha$ and $\gamma$, and $\pm\frac{\pi/2}{(2B)}$ for $\beta$ in an efficient way. Obviously, this solver is agnostic to the initial value and guarantees a global optimal rotation at the error level determined by the resolution.

%$\mathfrak{i}\mathfrak{F}_{SO(3)}$ generates the desired $f(\boldsymbol{r}_{\alpha,\beta,\gamma})$ with the size of $(2B+1)\times(2B+1)\times(2B+1)$. 

\boldparagraph{Probabilistic Approximation}To make the solver differentiable, we approximate $\arg\max$ by probabilistic modeling. We map the resultant correlation $f(\boldsymbol{r}_{\alpha,\beta,\gamma})$ in Eq.~\ref{eq:phasecorr_G1G2_SO3_final} to a discrete probability density function $p(\boldsymbol{r}_{\alpha,\beta,\gamma})$ by $\mathrm{softmax}$ function:
    \begin{equation}
      p(f(\boldsymbol{r}_{i}))=\frac{e^{\xi_{\boldsymbol{r}} f(\boldsymbol{r}_{i})}}{\sum_{j}e^{\xi_{\boldsymbol{r}} f(\boldsymbol{r}_{j})}},
      \label{eq:approxexp_rot}
    \end{equation}
where $\xi_{\boldsymbol{r}}$ is the solver temperature controlling the peak property of the density, and can be learned by data. With an abuse of notation, we use the subscript $i$ and $j$ to indicate the enumeration of the discretized 3D $Z-Y-Z$ Euler angle space. We take the expectation of the rotation as the estimation:
    \begin{equation}
      \hat{\boldsymbol{r}} = \sum_{\boldsymbol{r}}\boldsymbol{r}_{i}p(f(\boldsymbol{r}_{i})).
      \label{eq:approxexp_rot_l1}
    \end{equation}

We design two types of losses for supervision. The L1 loss between $\hat{\boldsymbol{r}}$ and the ground truth $\boldsymbol{r}^{*}$:
\begin{equation}
     \mathcal{L}_{\boldsymbol{r},l1} = \| \boldsymbol{r}^*- \hat{\boldsymbol{r}} \|_1,
      \label{eq:L1_rot}
\end{equation}
and the KL-Divergence between the rotation density and a gaussian $\delta_{\boldsymbol{r}^*}$ peaking at the ground truth $\boldsymbol{r}^{*}$:
\begin{equation}
 \mathcal{L}_{\boldsymbol{r},kld}=KLD(p(f(\boldsymbol{r})),\delta_{\boldsymbol{r}^*}),
\label{eq:kldr}
\end{equation}
where the standard deviation of the gaussian is a hand-crafted parameter. By probabilistic estimation, we approximate the global convergent solver Eq.~\ref{eq:phasecorr_G1G2_SO3_final} using a differentiable process. 

%Eq.~\ref{eq:problem_define_grid}, Eq.~\ref{eq:phasecorr_G1G2_trans_final} and Eq.~ \ref{eq:phasecorr_G1G2_SO3_final} with differentiable function. 
    
%For translation and scale, the same expectation estimator can be applied to approximate the non-differentiable $\arg\max$ based estimator. When learning, we assign a temperature coefficient $\xi$ to the $soft\max$ to tune the range of feature input, which accelerates the convergence, but does not make difference in theoretic derivation. 

\subsection{Scale Registration}

%If there are scale changes between the two inputs, we have to resolve the scale before estimating the translation as in section \ref{subsec:PC as Mul}(\textbf{Cartesian Phase Correlation when $\mu = 1$:}). Therefore we will elaborate upon such scale estimation in this section.
 
Recall the magnitude spectrum in Eq.~\ref{eq:magG1G21}, we can eliminate the effect of rotation by rotating $\hat{G}_{2}(\boldsymbol{j})$ with the estimated $\hat{\boldsymbol{r}}_{\hat{\alpha},\hat{\beta},\hat{\gamma}}$ in Eq.~\ref{eq:phasecorr_G1G2_SO3_final} and form $\hat{G}^{\hat{\boldsymbol{r}}}_{2}(\boldsymbol{j})$:
  \begin{equation}
   \hat{G}^{\hat{\boldsymbol{r}}}_{2}(\boldsymbol{j}) = \hat{G}_{2}(\hat{\boldsymbol{r}}_{\hat{\alpha},\hat{\beta},\hat{\gamma}}\boldsymbol{j}),
    \label{eq:rotate G2 for scale}
\end{equation}
which is differed to $\hat{G}_{1}(\boldsymbol{j})$ only by scale $\mu$: 
\begin{equation}
    \hat{G}_{1}(\boldsymbol{j}) = \mu^{-3}\hat{G}^{\hat{\boldsymbol{r}}}_{2}(\boldsymbol{j}\mu^{-1}).
    \label{eq:scale relation 3D}
\end{equation}

When dealing with such isotropic scale, we simplify the problem into 2D by accumulating the cubic grids along one axis (\eg $\mathbf{j}_{1}$) and form a 2D square grid:
 \begin{equation}
    \hat{G}_{\cdot_{(2d)}}(\boldsymbol{j}_{(2d)}) = \sum_{j_{1}=-B}^{B-1} \hat{G}_{\cdot}(\boldsymbol{j}),
    \label{eq:3D to 2D grid}
\end{equation}
where $\boldsymbol{j}_{(2d)} \in [-B,B-1]^{2}$ stands for axes in 2D grid $\hat{G}_{\cdot_{(2d)}}$ (\eg $\{\mathbf{j}_{2},\mathbf{j}_{3}\}$). Therefore, Eq.~\ref{eq:scale relation 3D} becomes:
\begin{equation}
    \hat{G}_{1_{(2d)}}(\boldsymbol{j}_{(2d)}) = \mu^{-3}\hat{G}^{\hat{\boldsymbol{r}}}_{2_{(2d)}}(\boldsymbol{j}_{(2d)}\mu^{-1}).
    \label{eq:scale relation 2D}
\end{equation}
By representing $\hat{G}_{\cdot_{(2d)}}(\boldsymbol{j}_{(2d)})$ in the log-polar coordinate, we have:
\begin{equation}
    \rho_.(\log|\boldsymbol{j}_{(2d)}|,\angle\boldsymbol{j}_{(2d)}) = \hat{G}_{\cdot_{(2d)}}(\boldsymbol{j}_{(2d)}).\label{eq:logpolar}
\end{equation}
%\begin{equation}
%    \hat{G}_{1_{(2d)}}(|\boldsymbol{j}_{(2d)}|,\angle\boldsymbol{j}_{(2d)}) = \hat{G}_{2_{(2d)}}(|\boldsymbol{j}_{(2d)}|\mu,\angle\boldsymbol{j}_{(2d)}).
%    \label{eq:polar transform}
%\end{equation}
We then aggregate the second dimension of the log-polar representation by summation, yielding:
\begin{equation}
    \hat{\rho}_.(\log|\boldsymbol{j}_{(2d)}|) = \sum_{\angle\boldsymbol{j}_{(2d)}} \rho_.(\log|\boldsymbol{j}_{(2d)}|,\angle\boldsymbol{j}_{(2d)}),
\end{equation}
which leads to the representation reflecting the scale as shift:
\begin{equation}
    \hat{\rho}_{1}(\log|\boldsymbol{j}_{(2d)}|) = \mu^{-3}\hat{\rho}_{2}(\log|\boldsymbol{j}_{(2d)}|+\log\mu^{-1}),
    \label{eq:logpolar transform}
\end{equation}
where $\log\mu^{-1}$ and $\log|\boldsymbol{j}_{(2d)}|$ are further denoted as $\underline{\mu}$ and $\underline{\boldsymbol{j}}$ for brevity.

\boldparagraph{Cartesian Phase Correlation}To solve the scale registration in Eq.~\ref{eq:logpolar transform}, we apply the similar phase correlation as Eq.~\ref{eq:define sphere corr} for rotation, but in Cartesian coordinates. Specifically, the correlation is defined as:
\begin{equation}
\begin{split}
    \Tilde{f}(\underline{\mu}) = \int \hat{\rho}_1(\underline{\boldsymbol{j}})\hat{\rho}_2(\underline{\boldsymbol{j}}-\underline{\mu})d\underline{\boldsymbol{j}},
\end{split}
    \label{eq:spatial correlation scale}
\end{equation}
We again apply Fourier transform to Eq.~\ref{eq:logpolar transform} to find the solution. Based on Eq.~\ref{eq:G1G2 relation}, we have:
  \begin{equation}
     P_{1}(\underline{\boldsymbol{J}}) = \mu^{-3}P_{2}(\underline{\boldsymbol{J}})e^{i2\pi\underline{\boldsymbol{J}}^{T}\underline{\mu}},
    \label{eq:scale only fourier transform G1G2}
\end{equation}
where $P_{1}$ and $P_{2}$ are the Fourier spectrum of $\hat{\rho}_{1}$ and $\hat{\rho}_{2}$, $\underline{\boldsymbol{J}}$ is the sampled frequency. To evaluate the correlation in frequency domain, we calculate the cross-power spectrum $F_{\mu}$ as:
 \begin{equation}
    \tilde{F}_{\underline{\mu}}(P_{1},P_{2}) = P_{1} \cdot \bar{P}_{2} = \mu^{-3}|P_{2}|^{2}e^{i2\pi\underline{\boldsymbol{J}}^{T}\underline{\mu}},
    \label{eq:phasecorr_G1G2 scale}
\end{equation}
Note that the phase of the cross-power spectrum is equivalent to the phase difference between grids. We then estimate the scale by taking the inverse Fourier transform ($\mathfrak{i}\mathfrak{F}$) of $\frac{\tilde{F}_{\underline{\mu}}}{|P_{2}|^{2}}$, yielding a normalized and impulsed correlation map $f(\underline{\mu})$, which ideally peaks at the real scale:
\begin{equation}
    f(\underline{\mu}) = \mathfrak{i}\mathfrak{F}(\frac{\tilde{F}_{\underline{\mu}}}{|P_{2}|^{2}}) = \mathfrak{i}\mathfrak{F}(\mu^{-3}e^{i2\pi\underline{\boldsymbol{J}}^{T}\underline{\mu}})
\end{equation}
As Eq.~\ref{eq:approxexp_rot_l1}, the estimation is built as:
\begin{equation}
\hat{\underline{\mu}} = \sum_{\underline{\mu}_i}\underline{\mu}_{i}p(f(\underline{\mu}_{i})),
    \label{eq:phasecorr_G1G2_scale_final}
\end{equation}
where $p(f(\underline{\mu}_{i}))$ is given by $\mathrm{softmax}$ function:
    \begin{equation}
      p(f(\underline{\mu}_{i}))=\frac{e^{\xi_{\mu} f(\underline{\mu}_{i})}}{\sum_{j}e^{\xi_{\mu} f(\underline{\mu}_{j})}},
      \label{eq:approxexp_scale}
    \end{equation}
following the probabilistic approximation in Eq.~\ref{eq:approxexp_rot}. After estimating $\hat{\underline{\mu}}$ and $p(f(\underline{\mu}_{i}))$, we can easily recover $\hat{\mu}$ and $p(f(\mu_{i}))$ with $\underline{\mu} = \log\mu^{-1}$. Then we define both L1 loss nad the KL-Divergence for scale registration:\
\begin{gather}
    \mathcal{L}_{\mu,l1} = \| \mu^*- \hat{\mu} \|_1 \\
    \mathcal{L}_{\mu,kld}=KLD(p(f(\mu)),\delta_{\mu^*}).
    \label{eq:loss scale}
\end{gather}

\subsection{Translation Registration}

With the estimated $\hat{\mu}$, we can further eliminate the effect of scale in frequency spectrum in Eq.~\ref{eq:g1g2 relation} as:
\begin{equation}
    g_{1}(\boldsymbol{k}) = g^{\hat{\boldsymbol{r}},\hat{\mu}}_{2}(\boldsymbol{k}-\boldsymbol{t}).
    \label{eq:define variable for dft g1g2}
\end{equation}
Note that the form is similar to Eq.~\ref{eq:logpolar transform} for the scale registration, but with the translation in 3D, so we apply 3D Cartesian phase correlation for the correlation of the translation as:
\begin{equation}
    f(\boldsymbol{t}) = \mathfrak{i}\mathfrak{F}(\frac{\tilde{F}}{|G_{2}|^{2}}) = \mathfrak{i}\mathfrak{F}(e^{i2\pi\boldsymbol{j}^{T}\boldsymbol{t}}),
    \label{tpc1}
\end{equation}
and build the expectation for translation registration:
\begin{equation}
    \hat{\boldsymbol{t}} = \sum_{\boldsymbol{t}}\boldsymbol{t}_{i}p(f(\boldsymbol{t}_{i})),
    \label{tpc2}
\end{equation}
where $p(f(\boldsymbol{t}_{i}))$ is derived by
\begin{equation}
    p(f(\boldsymbol{t}_{i}))=\frac{e^{\xi_{\boldsymbol{t}} f(\boldsymbol{t}_{i})}}{\sum_{j}e^{\xi_{\boldsymbol{t}} f(\boldsymbol{t}_{j})}}.
    \label{eq:phasecorr_G1G2_trans_final}
\end{equation}
We can also define both L1 loss and the KL-Divergence for translation registration:
\begin{gather}
    \mathcal{L}_{\boldsymbol{t},l1} = \| \boldsymbol{t}^*- \hat{\boldsymbol{t}} \|_1 \\
    \mathcal{L}_{\boldsymbol{t},kld}=KLD(p(f(\boldsymbol{t})),\delta_{\boldsymbol{t}^*}).
    \label{eq:loss t}
\end{gather}

Finally we finish the three-stage differentiable phase correlation solver, $Q_{DPC}$, for rotation, scale and translation registration, which is also globally convergent, and is able to back-propagate the error from losses to the input feature grids $g_1(\boldsymbol{k})$ and $g_2(\boldsymbol{k})$, as well as the three solver temperature parameters $\{\xi_{\boldsymbol{r}},\xi_{\mu},\xi_{\boldsymbol{t}}\}$.

%\subsection{Problem Formulation and System Defination}
%\label{subsec:Formulation and Defination}
%will be denoted as $\boldsymbol{k} \in [-B,B-1]^{3}$.

%\subsection{Decoupling of Translation, Scale and Rotation}
%\label{subsec:decouple of trans scale and rot}
%Since we are estimating rotations, scales, and translations step by step, we need to decouple these variables in the first place. 

%\subsection{Phase Correlation as Multiplication}
%\label{subsec:PC as Mul}
%Phase correlation is originally utilized as a mathmetical solution to estimate translational shifts along dimensions. However, with the expansion of Fourier transform from the Cartesian to the sphere ($\varmathbb{S}^{2}$) as well as the rotation group ($SO(3)$), 3D rotations between spheres can be further obtained. Here, following the registration pipeline, we will first introduce the spherical phase correlation for 3D rotation estimation before going to the Cartesian phase correlation for translation and scale retrieval.

 %\noindent\textbf{Cartesian Phase Correlation when $\mu = 1$:} 

 %\subsection{Scale Retrieval in Log-Polar Domain}
 %\label{subsec:scale retrieval}

\subsection{Extension to 2D}
\label{subsec:Downgrading to 2D}

In some applications, the pitch and roll, as well as the height can be aligned with onboard sensor \eg air-ground BEV image registration, leaving the $7DoF$ pose to $4DoF$. So we also extend the phase correlation to $4DoF$ pose registration with the input of 2D grids. 

Given two 2D grids $g_{1_{(2d)}}$ and $g_{2_{(2d)}}$ linked by the relative translation $\boldsymbol{t}$, rotation $\boldsymbol{r}$ and scale $\mu$:
\begin{equation}
   g_{1_{(2d)}}(\boldsymbol{k}_{(2d)}) =  g_{2_{(2d)}}(\boldsymbol{r}_{\alpha}\boldsymbol{k}_{(2d)}\mu-\boldsymbol{t}),
    \label{eq:2dg1g2}
\end{equation}
where $\boldsymbol{k}_{(2d)} \in [-B,B-1]^{2}$ denotes the index of 2D grid.

\boldparagraph{Decoupling of Translation, Scale and Rotation}We also follow the decoupling idea in $7DoF$ by first eliminating the translation. Replace Eq.~\ref{eq:magG1G21} and Eq.~\ref{eq:trans_inv_g} with 2D Fourier transform, we decouple the translation from pose registration:
\begin{equation}
     \hat{G}_{1_{(2d)}}(\boldsymbol{j}_{(2d)}) = \hat{G}_{2_{(2d)}}(\mu\boldsymbol{r}_{\alpha}\boldsymbol{j}_{(2d)}),
    \label{eq:pre_logpolar}
\end{equation}
where $\hat{G}_{\cdot_{(2d)}}$ is the magnitude of Fourier spectrum of $g_{\cdot_{(2d)}}$, and $\boldsymbol{j}_{(2d)}$ is the corresponding frequency coordinate in 2D. We apply Cartesian phase correlation for scale and rotation registration at the same time. 

We further represent $\hat{G}_{1_{(2d)}}$ and $\hat{G}_{2_{(2d)}}$ in the log-polar coordinate:
%\begin{equation}
%    p_{1}(|\boldsymbol{j}_{(2d)}|,\angle\boldsymbol{j}_{(2d)}) = p_{2}(|\boldsymbol{j}_{(2d)}|\mu,\angle\boldsymbol{j}_{(2d)}+\alpha).
%    \label{eq:polar transform 2d}
%\end{equation}
%By applying logarithm to $p_{1}$ and $p_{2}$ in the first dimension, we have $\hat{p}_{1}$ and $\hat{p}_{2}$:
\begin{equation}
    {\rho}_{1}(\log|\boldsymbol{j}_{(2d)}|,\angle\boldsymbol{j}_{(2d)}) = {\rho}_{2}(\log|\boldsymbol{j}_{(2d)}|+\log\mu,\angle\boldsymbol{j}_{(2d)}+\alpha).
    \label{eq:logpolar transform 2d}
\end{equation}
Note that for 2D grids, the rotation is only determined by one angle, so it is encoded linearly in one axis. 

\boldparagraph{Phase Correlation for Rotation, Scale and Translation}Note that Eq.~\ref{eq:logpolar transform 2d} has the similar form to Eq.~\ref{eq:define variable for dft g1g2}, where the rotation and scale are in the two axes, thus we similarly apply 2D Cartesian phase correlation like Eq.~\ref{tpc1}, Eq.~\ref{tpc2} and Eq.~\ref{eq:phasecorr_G1G2_trans_final} in 3D for estimation, and employ Eq.~\ref{eq:loss t} as losses. Once the rotation and scale are estimated, they are applied to compensate $g_{2_{(2d)}}$ to yield $g_{2_{(2d)}}^{\hat{\boldsymbol{r}}_{\alpha},\hat{\mu}}$, which leads to:
\begin{equation}
    g_{1_{(2d)}}(\boldsymbol{k}_{(2d)}) = g_{2_{(2d)}}^{\hat{\boldsymbol{r}}_{\alpha},\hat{\mu}}(\boldsymbol{k}_{(2d)}-\boldsymbol{t}).
    \label{eq:decoupled_trans_2d}
\end{equation}
To estimate $\hat{\boldsymbol{t}}$, we also apply 2D Cartesian phase correlation as that in rotation and scale stage, which finally completes the differentiable phase correlation solver for $4DoF$ case. 

%\subsection{Summary of the Registration on Grids}
%\label{subsec:PCsummary}

%We first decouple translation from rotation and scale by utilizing the magnitude of the Fourier transform. We further decouple the scale from rotation by spherical projection (for 3D cases) and log-polar projection (for 2D cases). Then the rotation is estimated through spherical phase correlation (3D) or Cartesian phase correlation (2D). After the rotation retrieval, the estimated rotation is applied to the source input and the output of which is rotational aligned with the template image. Then the scale is futher retrieved with Cartesian phase correlation. Finally, after template and source are scale and rotational aligned, we are able to solve the translation with Cartesian phase correlation.

 \begin{figure}[t]
    \centering
    \includegraphics[width=\linewidth]{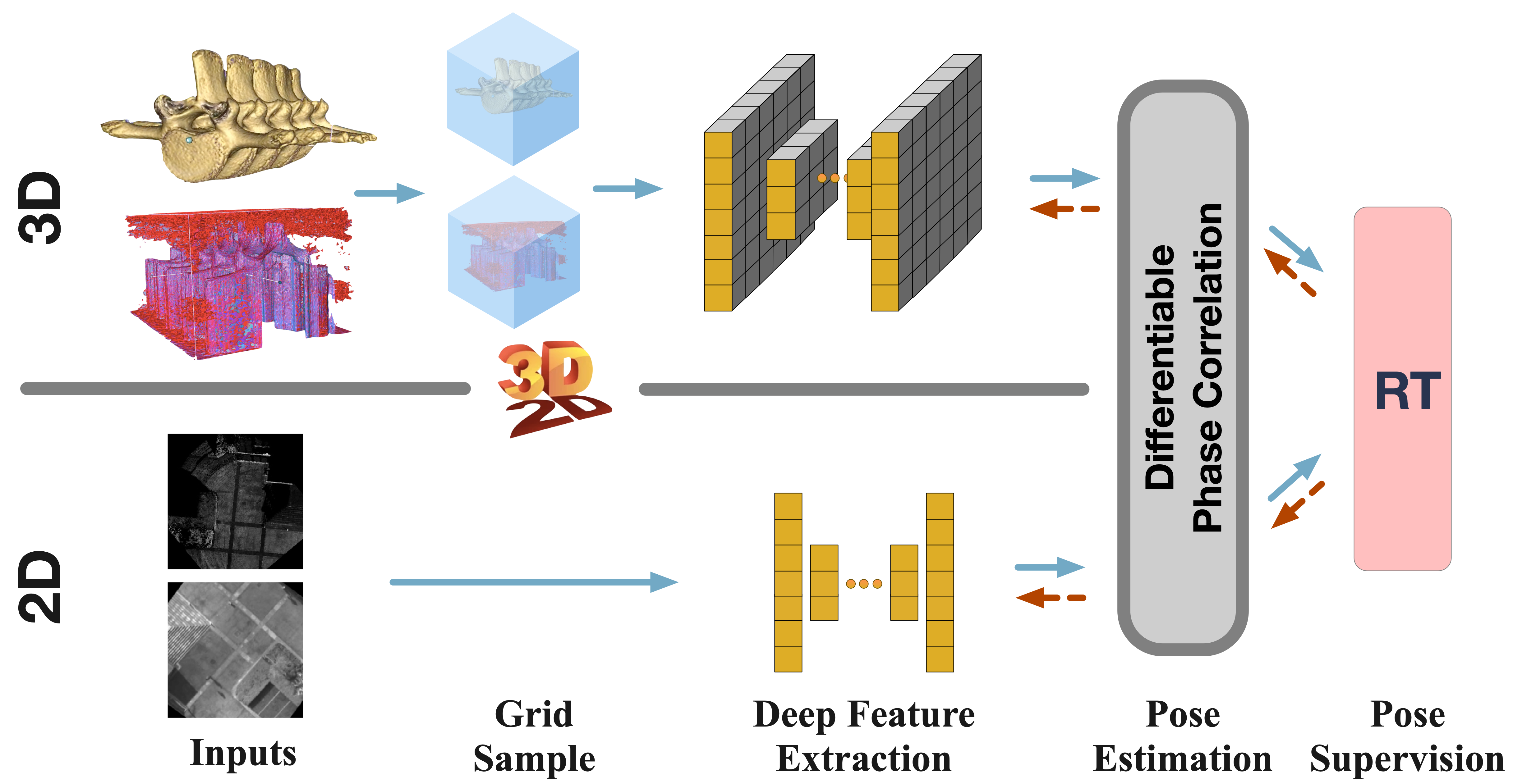}
    \caption{The overall pipeline of the rotation registration scheme of DPCN++. \textbf{3D:} The heterogeneous inputs are rasterized into $2B\times2B\times2B$ grid, where $B$ is the bandwidth. Then the two feature extractors retrieve unified features for the phase correlation solver. \textbf{2D:} The process is similar to the 3D version but different in the dimension of feature extraction.}
    \label{fig:net_structure}
\end{figure}

\section{Deep Phase Correlation Network}
\label{sec:DPCN++}

With the differentiable phase correlation $Q_{DPC}$, we have error backpropagated to the feature grids both in 3D and 2D. The remaining problem is the design of $Q_{\theta_1}$ and $Q_{\theta_2}$ for building the feature grid. The function consists of two parts, a voxelization part, which converts data in different modals to a unified grid based representation, and a deep network part, which extracts the dense feature from the grid.

\subsection{Data Voxelization}
%\label{subsec:GridSampling}

In the voxelization part, the process is simple. We first create a grid $V(\boldsymbol{k})$ with $\boldsymbol{k} \in [-B,B-1]^3$, and fill all voxels with zeros. Then, for non-grid data $v$ e.g. point cloud, we label the voxel with $1$ if there is a point occupying this voxel. Therefore, the resultant voxel is binary for pure point cloud, $V(\boldsymbol{k})\in \{0,1\}$. We can also assign voxels with the hand-crafted feature of points occupying it \eg density, resulting in a real valued grid, $V(\boldsymbol{k})\in \mathfrak{R}$. For data with additional features e.g. colored point cloud, a multi-channel grid can be built $V(\boldsymbol{k})\in \mathfrak{R}^C$, where $C$ is the size of the feature. In this paper, we focus on the first two types of grid. For grid data like SDF, we directly regard it as $V(\boldsymbol{k})$. For 2D cases, since the input images are already represented in the grid, we only need to reshape the images to the shape of $2B\times 2B$. 
% Detailed implementations of the grid voxelization are given in the Appendix~\ref{subsec:grid sampling append}.

\subsection{Deep Feature Extraction}
\label{subsec:Deep Feature Extractor}
In conventional phase correlation, low pass filter is applied to suppress high frequency noise in the two inputs, which can be seen as a special form of feature extractor \cite{SrinivasaReddy1996,bulow2018scale}. For more general form of inputs, the hand crafted low pass filter is far from sufficient. Considering that there is no common feature to directly supervise the feature extractor, an end-to-end learning of feature extractor leveraged by the differentiable phase correlation is able to address the problem without assuming feature types like points, lines or edges.

% For \textbf{homogeneous} data, we train a common U-Net for feature extraction of the two inputs. This is achieved by concatenating the source and the target along channels and feeding it to the same U-Net for each stage:
% \begin{gather}
%     g_{\boldsymbol{rs},1} \mathbin\Vert g_{\boldsymbol{rs},2} = Q_{\theta,\boldsymbol{rs}}(v_{1} \mathbin\Vert v_{2})\\
%     g_{\boldsymbol{t},1} \mathbin\Vert g_{\boldsymbol{t},2} = Q_{\theta,\boldsymbol{t}}(v_{1} \mathbin\Vert v_{2}),
% \label{eq:homo feature extraction}
% \end{gather}
% where $\mathbin\Vert$ is the concatenation while $Q_{\theta,\boldsymbol{rs}}$ and $Q_{\theta,\boldsymbol{t}}$ are the U-Nets trained for rotation(scale) stage and translation stage respectively. With $Q_{\theta,\boldsymbol{rs}}$ and $Q,\theta_{\boldsymbol{t}}$, we retrieve the feature grids for rotation(scale) estimation ($g_{\boldsymbol{rs},1}, g_{\boldsymbol{rs},2}$) and the feature grids for translation estimation ($g_{\boldsymbol{t},1}, g_{\boldsymbol{t},2}$).

Given a pair of inputs $v_1$ and $v_2$, after voxelization to data grids $V(\boldsymbol{k})$, features grids are then extracted using UNet $U(V(\boldsymbol{k}))$. We apply different feature extraction networks for rotation/scale registration, and translation registration:
\begin{gather}
    g_{\boldsymbol{r}\mu,1}(\boldsymbol{k}) = Q_{\theta_{1,\boldsymbol{r}\mu}}(v_{1}) = U_{\theta_{1,\boldsymbol{r}\mu}}(V_1(\boldsymbol{k}))\\
    g_{\boldsymbol{t},1}(\boldsymbol{k}) = Q_{\theta_{1,\boldsymbol{t}}}(v_{1}) =U_{\theta_{1,\boldsymbol{t}}}(V_1(\boldsymbol{k}))\\
    g_{\boldsymbol{r}\mu,2}(\boldsymbol{k}) = Q_{\theta_{2,\boldsymbol{r}\mu}}(v_{2}) = U_{\theta_{2,\boldsymbol{r}\mu}}(V_2(\boldsymbol{k}))\\
    g_{\boldsymbol{t},2}(\boldsymbol{k}) = Q_{\theta_{2,\boldsymbol{t}}}(v_{2}^{\boldsymbol{r},\mu}) = U_{\theta_{2,\boldsymbol{t}}}(V_2^{\boldsymbol{r},\mu}(\boldsymbol{k})),
\end{gather}
% \begin{gather}
%     g_{\boldsymbol{rs},1}(\boldsymbol{k}),g_{\boldsymbol{rs},2}(\boldsymbol{k}) = Q_{\theta_{1,\boldsymbol{rs}}}(v_{1}),Q_{\theta_{2,\boldsymbol{rs}}}(v_{2}) = U_{\theta_{2,\boldsymbol{rs}}}(V(\boldsymbol{k})),U_{\theta_{2,\boldsymbol{rs}}}(V(\boldsymbol{k}))\\
%     g_{\boldsymbol{t},1}(\boldsymbol{k}),g_{\boldsymbol{t},2}(\boldsymbol{k}) = Q_{\theta_{1,\boldsymbol{t}}}(v_{1}),Q_{\theta_{2,\boldsymbol{t}}}(v_{2}^{\boldsymbol{r},\mu})=U_{\theta_{2,\boldsymbol{rs}}}(V(\boldsymbol{k})),
% \label{eq:hetero feature extraction}
% \end{gather}
where $v_{2}^{\boldsymbol{r},\mu}$ and $V_2^{\boldsymbol{r},\mu}(\boldsymbol{k})$ are the rotation and scale compensated $v_{2}$ and the corresponding voxelized data grid respectively. Note that in the training stage, $\{\boldsymbol{r},\mu\}$ is given by the ground truth $\{\boldsymbol{r}^{*},\mu^{*}\}$ while in the inference stage, they are given by the estimated results $\{\hat{\boldsymbol{r}},\hat{\mu}\}$. The whole process of the DPCN++ forward inference is summarized in Algorithm~\ref{algo:DPCN++}.

\boldparagraph{Architecture in Detail}The UNet3D is adopted from \cite{3dunet}, with Each UNet constructed with 4 down-sampling encoder layers and 4 up-sampling decoder layers to extract features. We choose the LeakyReLU as activation \cite{xu2015empirical}. In the training stage, the parameters of the UNets as well as the temperature $\xi$ for $\mathrm{softmax}$ are tuned.

\begin{algorithm}[t]
	%\textsl{}\setstretch{1.8}
	\renewcommand{\algorithmicrequire}{\textbf{Input:}}
	\renewcommand{\algorithmicensure}{\textbf{Output:}}
	\caption{Deep Phase Correlation Network (DPCN++)}
	\begin{algorithmic}[1]
	\Require heterogeneous representation $\{v_{1},v_{2}\}$, bandwidth $B$, ground truth of relative pose $\{\boldsymbol{t}^{*},\boldsymbol{r}^{*}, \mu^{*}\}$
	\Ensure $\hat{\boldsymbol{t}},\hat{\boldsymbol{r}}, \hat{\mu}$
    \State $\triangleright$ \textbf{Feature Grids in Rotation(Scale) Stage with $Q_{\theta,\boldsymbol{r}\mu}$}
    \State $\{g_{\boldsymbol{r}\mu,1}, g_{\boldsymbol{r}\mu,2}\} \leftarrow \{Q_{\theta_{1},\boldsymbol{r}\mu}(v_{1}),Q_{\theta_{2},\boldsymbol{r}\mu}(v_{2})\}$
    \State $\triangleright$ \textbf{Decouple Translation}
    \State $\{\hat{G}_{1}, \hat{G}_{2}\} \leftarrow |FourierTransform(\{g_{\boldsymbol{r}\mu,1}, g_{\boldsymbol{r}\mu,2}\})|$
    \State $\triangleright$ \textbf{Decouple Scale}
    \If{dimension of $g_{\boldsymbol{r}\mu,1} = 2$}
        \State $\{s_{1}, s_{2}\} \leftarrow LogPolar(\{\hat{G}_{1}, \hat{G}_{2}\})$
    \ElsIf{dimension of $g_{\boldsymbol{r}\mu,1} = 3$}
        \State $\{s_{1}, s_{2}\} \leftarrow Spherical(\{\hat{G}_{1}, \hat{G}_{2}\})$ 
    \EndIf
    \State $\triangleright$ \textbf{Estimate} $\hat{\boldsymbol{r}}, \hat{\mu}$
    \If{dimension of $g_{\boldsymbol{r}\mu,1} = 2$}
        \State $\{\hat{\boldsymbol{r}},\hat{\mu}\} \leftarrow CartesianPhaseCorrelation(s_{1}, s_{2})$
    \ElsIf{dimension of $g_{\boldsymbol{r}\mu,1} = 3$}
        \State $\hat{\boldsymbol{r}} \leftarrow SphericalPhaseCorrelation(s_{1}, s_{2})$
        \State $\triangleright$ \textbf{Decouple Rotation}
        \State $g_{\boldsymbol{r}\mu,2}^{\hat{\boldsymbol{r}}} \leftarrow \mathbf{Transform}(g_{\boldsymbol{r}\mu,2},\hat{\boldsymbol{r}})$ 
        \State $\{g_{1,2D},g_{2,2D}^{\hat{\boldsymbol{r}}}\} \leftarrow \{\sum_{j_{1}}g_{\boldsymbol{r}\mu,1},\sum_{j_{1}}g_{\boldsymbol{r}\mu,2}^{\hat{\boldsymbol{r}}}\}$
        \State $\hat{\mu} \leftarrow CartesianPhaseCorrelation(LogPolar($\\
        \hspace{3cm}$|FourierTransform(\{g_{1,2D},g_{2,2D}^{\hat{\boldsymbol{r}}}\})|))$
    \EndIf
    \State $\triangleright$ \textbf{Estimate} $\hat{\boldsymbol{t}}$
    \State $v_{2}^{\hat{\boldsymbol{r}},\hat{\mu}} \leftarrow \mathbf{Transform}(v_{2},\{\hat{\boldsymbol{r}},\hat{\mu}\})$
    \State $\triangleright$ \textbf{Feature Grids in Translation Stage with $Q_{\theta,\boldsymbol{t}}$}
    \State $\{g_{\boldsymbol{t},1}, g_{\boldsymbol{t},2}\} \leftarrow \{Q_{\theta_{1},\boldsymbol{t}}(v_{1}),Q_{\theta_{2},\boldsymbol{t}}(v_{2}^{\hat{\boldsymbol{r}},\hat{\mu}})\}$
    \State $\hat{\boldsymbol{t}} \leftarrow CartesianPhaseCorrelation(g_{\boldsymbol{t},1}, g_{\boldsymbol{t},2}) $
    \State $\triangleright$ \textbf{Train All Feature Extractors}
    \State $\mathbf{BackPropagate}(L(\{\hat{\boldsymbol{t}},\hat{\boldsymbol{r}}, \hat{\mu}\},\{\boldsymbol{t}^{*},\boldsymbol{r}^{*}, \mu^{*}\}))$
    \State \textbf{return} $\hat{\boldsymbol{t}},\hat{\boldsymbol{r}}, \hat{\mu}$
	\end{algorithmic}  
	\label{algo:DPCN++}
\end{algorithm}

\subsection{The Chain Rule of DPCN++}
\label{subsec:Differentialization of the Phase Correlation}

We present the chain rule to derive the back-propagation evaluation to train DPCN++. We first show the gradient of $Q_{DPC}$, by which the gradient of loss with respect to the network parameter $\{\theta_{1,\boldsymbol{rs}}, \theta_{2,\boldsymbol{rs}}, \theta_{1,\boldsymbol{t}}, \theta_{2,\boldsymbol{t}}\}$ is derived. 

\boldparagraph{Gradient of DPC with respect to Input}We set rotation as an example to derive the gradient of the loss with respect to the first input $g_1(\boldsymbol{k})$ as: 
\begin{equation}
    \frac{\partial \mathcal{L}_r}{\partial g_1(\boldsymbol{k})}  =
    \frac{\partial \mathcal{L}_r}{\partial \hat{\boldsymbol{r}}}
    \sum_{\boldsymbol{r}_{i}} \frac{\partial \hat{\boldsymbol{r}}}{\partial f(\boldsymbol{r}_{i})}
    \frac{\partial f(\boldsymbol{r}_{i})}{\partial g_1(\boldsymbol{k})},
    % \frac{\partial F}{\partial \check{S}_{1}} \\
    % & \cdot\sum_{u=0}^{2B-1}\sum_{v=0}^{2B-1}  \frac{\partial \check{S}_{1}}{\partial s_{1}(\boldsymbol{\lambda}_{u,v})} 
    % \frac{\partial s_{1}(\boldsymbol{\lambda}_{u,v})}{\partial g_1(\boldsymbol{k})}
\label{eq:chain_rule_gk}
\end{equation}
where the gradient consists of three terms. The first term is common. The second term, according to Eq.~\ref{eq:approxexp_rot}, we have:
\begin{equation}
\frac{\partial \hat{\boldsymbol{r}}}{\partial f(\boldsymbol{r}_{i})}  = 
\frac{ \sum_{j}\xi_{\boldsymbol{r}}(\boldsymbol{r}_{i}-\boldsymbol{r}_{j})e^{{\xi_{\boldsymbol{r}}}(f(\boldsymbol{r}_{i})+f(\boldsymbol{r}_{j}))}}{(\sum_{k}e^{{\xi_{\boldsymbol{r}}}f(\boldsymbol{r}_{k})})^2}.
\end{equation}
For the third term, it is the gradient of the Fourier and spherical transforms. Note that these transforms are actually linear, so the derivation is not difficult. We leave the derivation in Appendix \ref{subsec:backward propogation}.

\boldparagraph{Gradient of DPC with respect to Temperature}For the temperature parameter $\xi_{\boldsymbol{r}}$, the gradient is: 
\begin{equation}
\frac{\partial \mathcal{L}_r}{\partial \xi_{\boldsymbol{r}}}=
\frac{\partial \mathcal{L}_r}{\partial \hat{\boldsymbol{r}}}
\frac{\partial  \hat{\boldsymbol{r}}}{\partial \xi_{\boldsymbol{r}}},
\label{eq:chain_rule_xi_r}
\end{equation}
where the first term is the same as that in Eq.~\ref{eq:chain_rule_gk}, and the second term is given as:
\begin{equation}
\frac{\partial  \hat{\boldsymbol{r}}}{\partial \xi_{\boldsymbol{r}}} = 
\frac{\sum_{i}\sum_{j}\boldsymbol{r}_{i}e^{{\xi_{\boldsymbol{r}}}(f(\boldsymbol{r}_{i})+f(\boldsymbol{r}_{j}))}(\boldsymbol{r}_{i}-\boldsymbol{r}_{j})}{(\sum_{k}e^{{\xi_{\boldsymbol{r}}}f(\boldsymbol{r}_{k})})^2}.
\label{eq:chain_rule_xi_r2}
\end{equation}

 \begin{figure*}[t]
    \centering
    \includegraphics[width=\textwidth]{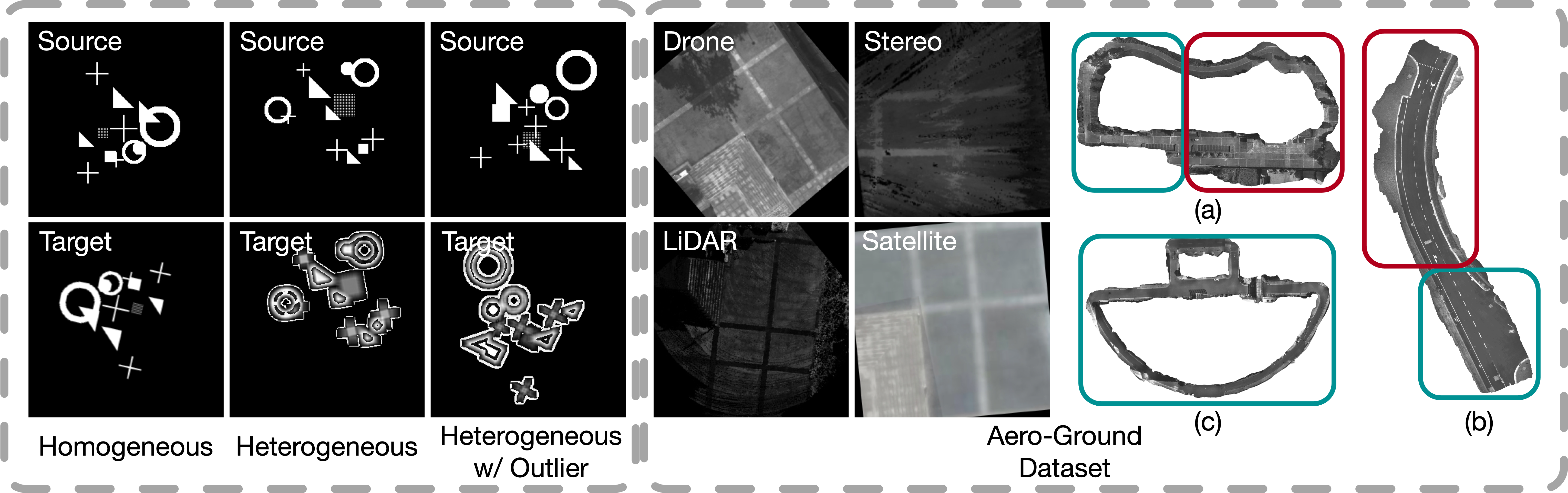}
    \caption{\textbf{Simulation dataset (left)} containing ``Homogeneous", ``Heterogeneous" and ``Heterogeneous w/ Outlier" sets. \textbf{Aero-Ground Dataset (right)} containing ``drone's view", ``LiDAR intensity", ``stereo" and ``satellite". The experiments are carried out on location (a) and (b) separately in which the model is trained on images pairs generated inside red areas and validated on images pairs generated inside blue area. The generalization is carried out with estimating poses of images inside location (c) with models trained on (a) and (b).}
    \label{fig:2d dataset}
\end{figure*}

\boldparagraph{Gradient of DPCN++}We finally show the full chain rule of the rotation supervised loss with respect to the feature extraction network parameters $\theta_{1,\boldsymbol{rs}}$:
% \begin{equation}
% \begin{split}
% \frac{\partial \mathcal{L}_r}{\partial \theta_1} &=
% \frac{\partial \mathcal{L}_r}{\partial g_1(\boldsymbol{k})}\frac{\partial g_1}{\partial \theta_1}\\
% &=\sum_{\boldsymbol{r}_{i}}  \mathfrak{i}\mathfrak{F}_{S^2} [\mathfrak{F}_{SO(3)}(\frac{\partial \mathcal{L}_r}{\partial \hat{\boldsymbol{r}}}
% \frac{\partial \hat{\boldsymbol{r}}}{\partial f(\boldsymbol{r}_{i})})
% \frac{\partial F}{\partial \check{S}_{1}}]
% \frac{\partial s_{1}}{\partial g_{1}}   
% \frac{\partial g_{1}}{\partial \theta_1}
% \end{split}
% \label{eq:chain_rule_appendix_2}
% \end{equation}

\begin{equation}
\frac{\partial \mathcal{L}_r}{\partial \theta_{1,\boldsymbol{rs}}} =\sum_{\boldsymbol{k}}
\frac{\partial \mathcal{L}_r}{\partial g_1(\boldsymbol{k})}\frac{\partial g_1(\boldsymbol{k})}{\partial \theta_{1,\boldsymbol{rs}}},
\label{eq:chain_rule_appendix_2}
\end{equation}
where the first term of the gradient is given in Eq.~\ref{eq:chain_rule_gk}, the second term is given by the gradient of the UNet, which we can calculate by auto-gradient in public available deep learning package.
% \begin{equation}
% \begin{split}
% \frac{\partial \mathcal{L}_r}{\partial \theta_1} &=
% \frac{\partial \mathcal{L}_r}{\partial g_1(\boldsymbol{k})}\frac{\partial g_1}{\partial \theta_1}\\
% &= \frac{\partial \mathcal{L}_r}{\partial \hat{\boldsymbol{r}}}
% \sum_{\boldsymbol{r}_{i}} \frac{\partial \hat{\boldsymbol{r}}}{\partial f(\boldsymbol{r}_{i})}
% \frac{\partial f(\boldsymbol{r}_{i})}{\partial F}
% \frac{\partial F}{\partial \check{S}_{1}} \\
% & \cdot\sum_{u=0}^{2B-1}\sum_{v=0}^{2B-1}  \frac{\partial \check{S}_{1}}{\partial s_{1}(\boldsymbol{\lambda}_{u,v})} 
% \frac{\partial s_{1}(\boldsymbol{\lambda}_{u,v})}{\partial g_1(\boldsymbol{k})}  
% \frac{\partial g_{1}}{\partial \theta_1}
% \end{split}
% \label{eq:chain_rule_appendix_2}
% \end{equation}

For the back-propagation pathway from loss to translation and scale, as well as the gradient with respect to the second input, a similar process can be followed. Note that the derivation above does not depend on any specific feature network, it is a general framework for implementing the back-propagation of versatile pose registration using DPCN++.

\subsection{Training Details}
\label{subsec:trainng details}

Following the forward and backward process introduced above, we can implement the whole DPCN++ with gradients except for several sampling processes including: spherical transform, DH-Grid transform and log-polar transform. We apply sampling by linear interpolation, through which the re-sampled pixels are a continuous function of pre-sampled pixels, keeping valid gradients.

\boldparagraph{Loss and Learning Setting} We conclude the DPCN++ by presenting the total loss of DPCN++, which is a combination of rotation loss $\mathcal{L}_{\boldsymbol{r}}$, translation loss $\mathcal{L}_{\boldsymbol{t}}$, and scale loss $\mathcal{L}_{\mu}$:
\begin{equation}
\begin{split}
 \mathcal{L}&=\mathcal{L}_{\boldsymbol{r}}+\mathcal{L}_{\boldsymbol{t}}+\mathcal{L}_{\mu}\\
 &=w_{a}\mathcal{L}_{\boldsymbol{r},kld}+w_{b}\mathcal{L}_{\boldsymbol{r},l1}+w_{c}\mathcal{L}_{\boldsymbol{t},kld}\\
 &+w_{d}\mathcal{L}_{\boldsymbol{t},l1}+w_{e}\mathcal{L}_{\mu,kld}+w_{f}\mathcal{L}_{\mu,l1},
 \end{split}
\label{eq:total_loss_concrete}
\end{equation}
where $w_{\cdot}$ are the weights for each loss. In the experiments, these weights are $\{1,3,3,1,1,3\}$. This total loss guides the learning of both feature extractor parameters and the optimizer temperature parameters.

The code is implemented under PyTorch, with the hardware settings of CPU i9 12900k, GPU RTX3090$\times 2$, and RAM 128GB. The learning rate is set to $3e^{-4}$ for heterogeneous and $5e^{-5}$ for homogeneous with epoch decay. Due to the memory consumption of 3D UNets, the training batch size is set to 1, which also means there is no batch level operation.

\begin{figure*}[t]
\centering
\includegraphics[width=\linewidth]{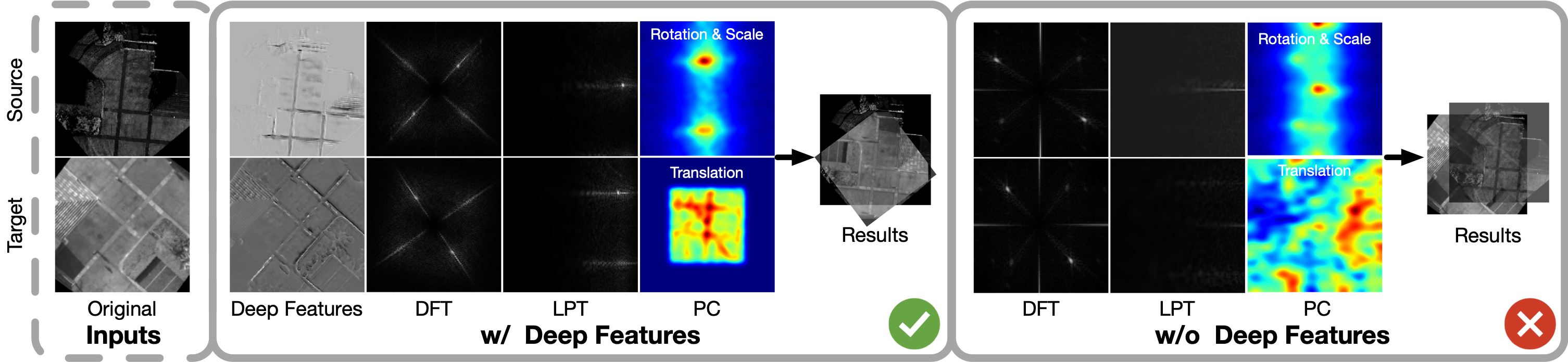}
\caption{\textbf{Qualitative Demonstration} of the case study for heterogeneous images. DFT, LPT and PC is abbreviated for discrete Fourier transform, Log-Polar transform and phase correlation, respectively.}
\label{fig:2d_result_case}
\vspace{-15pt}
\end{figure*}

\section{Experimental Results}
%In this section, we introduce the experimental setups for the evaluation as well as the results. We will exclusively introduce i)the dataset we evaluate each examples on, ii)  the baselines we compared to, iii) the metrics we adopt in evaluation, and iv) the results of each example. 

We first conduct experiments on 2D-2D BEV image registration, along with a case study to illustrate the intermediate steps of our method in 2D. Next, we show extensive comparisons of 3D-3D homogeneous and heterogeneous pose registration. Finally, we validate our design choices in the ablation study.

\subsection{2D-2D Pose Registration}

\boldparagraph{Datasets} We evaluate our approach on two 2D datasets. First, we construct a simulation dataset consisting of random 2D primitives, enabling a comparison of all methods using noise-free ground truth. As shown in Fig.~\ref{fig:2d dataset} (left), We further apply  Gaussian filtering to the target images to additionally simulate challenging heterogeneous image pairs. This dataset contains 2000 randomly generated images pairs for training and 1000 pairs for evaluation. 

\begin{table}[t]
% \captionsetup{justification=raggedright,singlelinecheck=false}
\vspace{-10pt}
\caption{Results of the heterogeneous 2D registration in simulation and in scene (a) and (b) of the AG dataset. Note: ``l2sat'', ``l2d'', ``s2sat'', ``s2d'' are the abbreviation for 
    ``LiDAR Local Map" to ``Satellite Map",
    ``LiDAR Local Map" to ``Drone's Birds-eye Camera",
    ``Stereo Local Map" to ``Satellite Map",
    ``Stereo Local Map" to ``Drone's Birds-eye Camera", respectively. The runtime are measured in milliseconds. The best numbers are in \textbf{bold} and the secondaries are  \underline{underlined}.} 
\label{tab: 2d results}
\centering
\resizebox{\linewidth}{!}{
\begin{tabular}{@{}lllllll@{}}
\toprule[1pt]
 Method & Exp. & $Acc_{x_{10}}$  & $Acc_{y_{10}}$  & $Acc_{r_{1}}$  & $Acc_{\mu_{0.2}}$ & Runtime \\ \midrule

 PC~\cite{SrinivasaReddy1996}  & sim   & 69.1 & 45.7 & 72.3 & \underline{97.6} & \underline{18.3}\\
                     R2D2~\cite{revaud2019r2d2}                   & sim    & 31.4 &  21.4 &  40.2 &  75.4 & 244.2 \\
                     DS~\cite{Barnes2019}             & sim    & 87.1 &  91.9 & 23.9 & $\setminus$ &  301.3\\
                     DAM~\cite{Park2020}       & sim   &  \underline{99.6} &  \underline{99.2} &  \underline{80.8} & $\setminus$ & 114.2 \\
                     RPR~\cite{Kendall2017}             & sim   & 62.1 &  49.1 &  78.3 &  96.7 & \textbf{6.47}\\
                     \textbf{DPCN++}             & sim   & \textbf{100}  & \textbf{100} &  \textbf{100}  & \textbf{100} & 22.1 \\ \midrule
                     
 R2D2~\cite{revaud2019r2d2}    & l2sat(a)  & 32.7 & 41.1 &  37.6  & 71.2 & 244.1 \\
                     & l2d(a)    & 39.1  & 45.7 &  30.1 & 73.5 & 244.1 \\
                     & s2sat(a)  & 32.9  & 40.6 &  27.6 &  69.5 & 244.3 \\
                     & s2d(a)    & 41.7  & 41.9 &  32.9 & 75.9 & 244.2 \\
                     DAM~\cite{Park2020}      & l2sat(a) &  \underline{55.4} & \underline{70.8} &  \underline{37.8} & $\setminus$ &  \underline{110.6} \\
                     & l2d(a)   &  \underline{39.4} &  \underline{66.8} &  \underline{22.5} &  $\setminus$ &  \underline{117.3} \\
                     & s2sat(a) &  \underline{35.2} &  \underline{33.6} &  \underline{24.1} &  $\setminus$ &  \underline{114.4} \\
                     & s2d(a)   &  \underline{51.5} &  \underline{43.9} &  \underline{33.9} &  $\setminus$ &  \underline{114.2} \\
                     \textbf{DPCN++}        & l2sat(a) &  \textbf{96.9} &  \textbf{98.0} &  \textbf{99.2} &  \textbf{95.5} & \textbf{24.75} \\
                     & l2d(a)   &  \textbf{98.2} &  \textbf{94.0} &  \textbf{99.2} &  \textbf{94.2} & \textbf{26.37} \\
                     & s2sat(a) &  \textbf{90.9} &  \textbf{97.8} &  \textbf{97.4} &  \textbf{93.7} & \textbf{23.61} \\
                     & s2d(a)   &  \textbf{91.3} &  \textbf{92.6} &  \textbf{99.3} &  \textbf{93.5} & \textbf{24.72} \\ \midrule

                      R2D2~\cite{revaud2019r2d2}                 & l2d(b)    & 26.7  & 39.4  & 22.4 &  64.3 & 244.2 \\
                     & s2d(b)   & 22.8  & 28.3 &  25.1 & 66.8 & 244.0 \\
                     DAM~\cite{Park2020}                   & l2d(b)   &  \underline{30.1} &  \underline{42.2} &  \underline{35.1} &  $\setminus$ & \underline{113.9} \\
                     & s2d(b)   &  \underline{40.9} &  \underline{49.6} &  \underline{27.4} &  $\setminus$ & \underline{116.5} \\
                     \textbf{DPCN++}        & l2d(b)   &  \textbf{96.2} &  \textbf{89.2} &  \textbf{99.7} &  \textbf{99.7} & \textbf{24.51}\\
                      & s2d(b)   &  \textbf{91.6} &  \textbf{90.6} &  \textbf{99.4} &  \textbf{95.0} & \textbf{25.63} \\
\bottomrule[1pt]
\end{tabular}
}
\vspace{-15pt}
\end{table}

To further verify our performance in real-world BEV image pose registration, we collect a multi-modal Aero-Ground (AG) Dataset. This dataset allows for evaluating cooperative localization between ground mobile robots, micro aerial vehicles (MAVs) and satellite. Specifically, it contains several different image pairs as follows:
\begin{itemize}
    \item ``LiDAR Local Map" to ``Drone's Birds-eye Camera";
    \item ``LiDAR Local Map" to ``Satellite Map";
    \item ``Stereo Local Map" to ``Drone's Birds-eye Camera";
    \item ``Stereo Local Map" to ``Satellite Map".
\end{itemize}
The $4DoF$ ground truth of the dataset is manually labeled. This dataset contains three scenes, where we split the training and testing regions without spatial overlapping, see Fig.~\ref{fig:2d dataset} (right).
%On the simulation dataset, we evaluate our work on homogeneous, heterogeneous, and heterogeneous with dynamic obstacles images pairs whereas on the Aero-Ground Dataset, 
%Our multi-modal dataset  evaluate our work on the application of cooperative SLAM system between ground mobile robots, the MAV and the Satellite. 
Both the simulation dataset and the AG dataset have image resolutions of $256 \times 256$. For each input pair, we constrain both horizontal ($x$) and vertical ($y$) translation changes in the range of $[-50, 50]$ pixels, together with rotation and scale change within $[0, \pi)$ and $[0.8, 1.2]$, respectively.

% \begin{figure}[t]
% \centering
% \includegraphics[width=\linewidth]{IEEEtran/Figure/2d_demo.png}
% \caption{The qualitative examples of 2D image registration with DPCN++.}
% \label{fig:2d__result_demo}
% \vspace{-10pt}
% \end{figure}

\begin{figure*}[t]
\centering
\includegraphics[width=\textwidth]{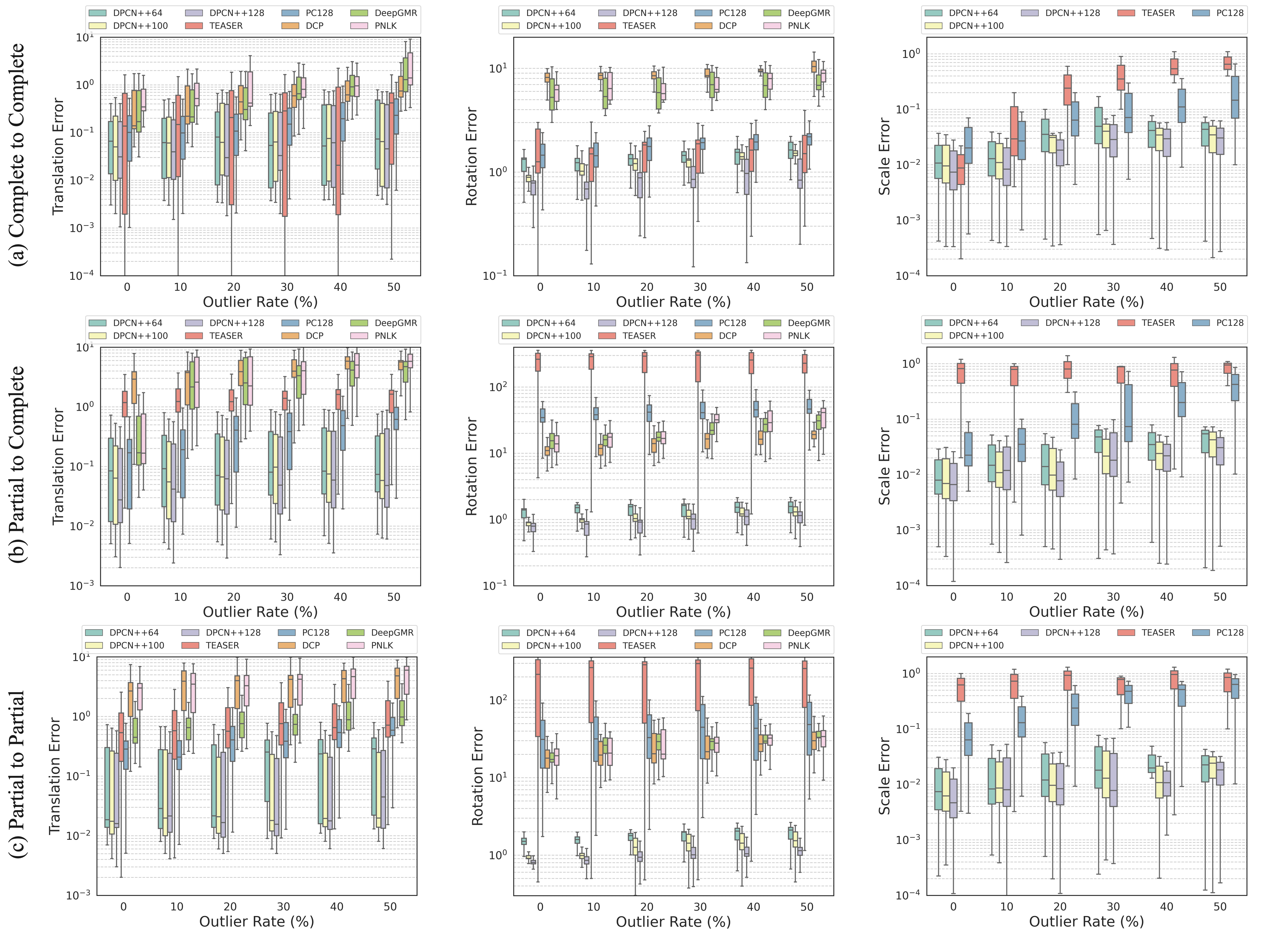}
\vspace{-10pt}
\caption{\textbf{Quantitative Comparison on Homogeneous Object-level Registration} in MVP dataset.}
\label{fig:pc pc quant}
\vspace{-15pt}
\end{figure*}

\begin{figure*}[t]
\centering
\includegraphics[width=0.85\textwidth]{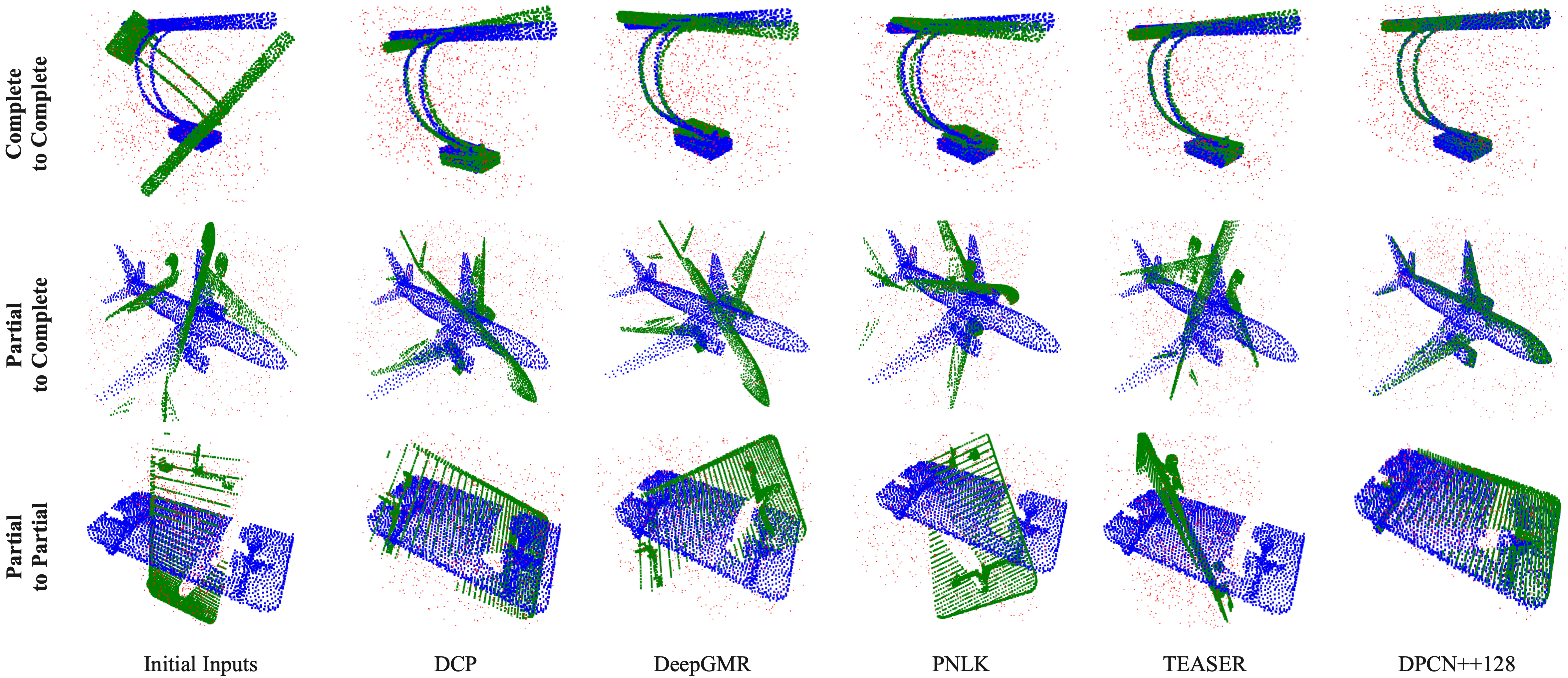}
\caption{\textbf{Qualitative Comparison on Homogeneous Object-level Registration} on the MVP dataset. We show point cloud-wise registration results with $50\%$ outliers. Note: Blue and green point the source and target and the red are the outliers.}
\label{fig:pc pc demo}
\vspace{-10pt}
\end{figure*}

\boldparagraph{Baselines} For 2D image $4DoF$ registration, we first compare the performance with Phase Correlation (PC)~\cite{SrinivasaReddy1996}, a learning-free version of our approach by eliminating the feature extraction network.  We further compare to learning-based methods, Relative Pose Regression (RPR) \cite{Kendall2017}, DAM~\cite{Park2020}, R2D2~\cite{revaud2019r2d2}, and Dense Search (DS) \cite{Barnes2019}.
 %\textsc{Phase Correlation} is the baseline for registering two homogeneous images and the pipeline of which is also partly adopted in our approach. We select it as a benchmark for evaluating the performance of our approach in matching homogeneous images. 
 %\textsc{QATM} is a representative work in image registration by learning features for matching, followed by non-maximum suppression for matching selection. It could handles translation displacement with high accuracy and therefore we select it as the benchmark for evaluating translation estimations in heterogeneous images. Unfortunately, the author of QATM only provided a pretrained model without a training script so that we could only evaluate its performance with the provided model. \yl{not a good argument, can we argue that it should generalize to other datasets? This can be omit if it is too detailed.}\jc{It is not fair to have QATM as a baseline, we should delete this..} 
RPR~\cite{Kendall2017} directly regresses the pose without integrating an explicit pose solver. DAM is similar to RPR but introduces feature correlations as an intermediate representation. As DAM does not estimate scale, we provide the ground truth scale to DAM in both training and testing.  We retrain both RPR and DAM on our datasets. R2D2 is a feature point based pose registration method, which, estimates a similarity transform using feature correspondences together with an SVD solver. This is recently popular for visual localization \cite{jiao2021r2d2loc} \cite{zhang2021reference}. Finally, DS is similar to our work but uses a different solver which exhaustively rotates the source image among a set of candidate angles to find the best match \wrt the target. Due to the exhaustive search, DS is computationally expensive, thus we also relax it to $3DoF$ with ground truth scale as DAM, and train it in a small range, $[0^{\circ},15^{\circ}]$.
 %It is selected as a benchmark to demonstrate the advantage of estimation continuity and efficiency of our approach.

% \boldparagraph{Setup}Few baseline support $4DoF$ (or more) pose estimations across sensors and therefore we relax the condition by providing rotation and scale ground truth as initials to several benchmark for comparison, including the DAM with scaling initialized. \yl{already mentioned in \textbf{baselines}? In addition, if rotation and scale are provided, are they only used as initialization or are they still optimized? Is there only DAM that needs the extra scale information? Not clear}

\boldparagraph{Evaluation Metrics} We evaluate the percentage of estimation with an error lower than a given threshold, Accuracy in Units (Acc):
\begin{gather}
     Acc_{\boldsymbol{r}_{1}} = \frac{1}{n}  \sum_{i=1}^{n}\#\{|\boldsymbol{r}^{*}_{i}-\hat{\boldsymbol{r}}_{i}| \leqslant 1^{\circ}\} \times 100\%  \\
     Acc_{\boldsymbol{t}_{10}} = \frac{1}{n}  \sum_{i=1}^{n}\#\{|\boldsymbol{t}^{*}_{i}-\hat{\boldsymbol{t}}_{i}| \leqslant 10pixels\} \times 100\%  \\
     Acc_{\mu_{0.2}} = \frac{1}{n} \sum_{i=1}^{n}\#\{|\mu^{*}_{i}-\hat{\mu}_{i}| \leqslant 0.2\} \times 100\% 
\label{eq: Err_2D}
\end{gather}
where $\#$ is the count of the set $\{\cdot\}$, and $n$ is the total amount of image pairs. Note that for the AG dataset, each ground image is generated at a scale of $0.1m$ per pixel. Thus, the threshold error of $10pixels$ for translation indicates for $1m$ in the real world.

\begin{figure*}[t]
\centering
\includegraphics[width=0.9\linewidth]{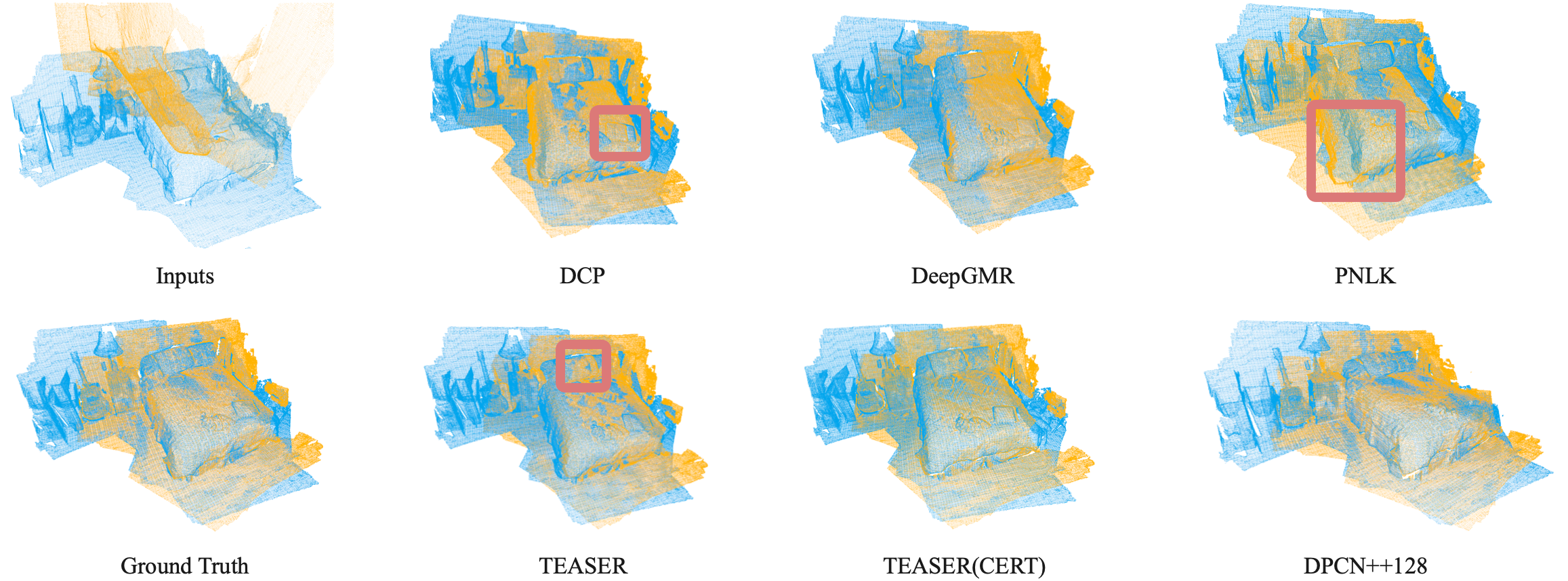}
\caption{\textbf{Qualitative Comparison on Homogeneous Scene-level Registration} on 3DMatch. We show one successful case for all baselines. The flaws of several baselines are captured in red boxes.}
\label{fig:3dmatch demo}
\vspace{-15pt}
\end{figure*}

\subsubsection{Case Study}
We first conduct a case study for the 2D registration by visualizing and analyzing intermediate results of our pipeline during the 2D heterogeneous registration on the AG dataset. Fig.~\ref{fig:2d_result_case} shows the intermediate result of discrete Fourier transform, and Log-Polar transform when estimating rotation and scale, as well as the phase correlation results for both stages (translation, rotation and scale). It reveals that, when dealing with heterogeneous inputs, the traditional PC fails due to the difference in appearance. In contrast, by training end to end, the feature extractors learn to generate features that can be registered by PC, thus capable of matching heterogeneous inputs. More cases can be found in Appendix~\ref{subsec:append_DPCN_2D_result}.

\subsubsection{Heterogeneous Registration}
We now compare with the baselines in the challenging heterogeneous registration. The comparisons given homogeneous inputs are shown in Appendix~\ref{subsec: append_Elaboration on 2D Sim}.
%, and elaborate more on the heterogeneous one in the manuscript. Therefore, in this section, we conduct experiments in simulation as well as on the Aero-Ground Dataset to match images across sensors.

\boldparagraph{Simulation Dataset} Experimental results on the simulation dataset are shown in Tab.~\ref{tab: 2d results} (sim). It can be seen that our approach reaches 100$\%$ accuracy rate on the simulation dataset, outperforming all baselines. Though PC and R2D2 perform well in the homogeneous case as shown in Appendix~\ref{subsec: append_Elaboration on 2D Sim}, they fail in heterogeneous cases, demonstrating the effectiveness of our combination of learned feature extractors and PC. Note that DPCN++ only requires ground truth pose for supervision, while R2D2 relies on supervision in the form of dense correspondences that are harder to obtain. The better performance of DAM against RPR validates the effectiveness of an explicit intermediate representation to bring inductive bias. But the direct regression of DAM causes the lower accuracy than that of DPCN++. For DS, it is mainly limited by the search range due to the expensive cost. It is worth noting that DPCN++ can also be considered as an exhaustive solver, but the solution space is significantly reduced by decoupling rotation, scale and translation. In terms of efficiency, our method is slower than and RPR, yet still allows for pose registration at 50 FPS. The performance is also evaluated by Mean Square Error (MSE) and more thresholds of $Acc$ in the Appendix~\ref{subsec: append_Elaboration on 2D Sim}.
%In real world experiments, 

\boldparagraph{AG Dataset} In this experiment, we select R2D2 and DAM as representative baselines for solver-based and solver-free methods, respectively. Tab.~\ref{tab: 2d results} shows evaluation results on scene (a) and (b) of the AG dataset. As can be seen, when estimating $4DoF$ poses across real-world sensors, our approach achieves accuracy rate above $89.2\%$ at a threshold of $1m$, typically sufficient for many applications. Though DAM is relaxed to $3DoF$ registration given ground truth scale, our approach still outperforms it by a large margin. The result of R2D2 indicates that, when the style of inputs changes drastically, matching local feature tends to be difficult. The Mean Square Error (MSE) and more thresholds for $Acc$ are evaluated in the Appendix~\ref{subsec: append_Elaboration on 2D Sim}~\ref{subsec: append_Elaboration on 2D AG}. The additional examples of the experiments comparing with the classical PC is shown in the Appendix \ref{subsec:append_DPCN_2D_result} and \ref{subsec:append_PC}.

To evaluate the generalization of DPCN++, we conduct experiments on scene (c) with DPCN++ trained on scene (a) and (b) and DAM model trained on partial data of scene (c). The results shown in the Appendix~\ref{subsec: append_Elaboration on 2D generalization} indicate that our approach is capable of estimating poses with a similar accuracy regardless of scene changes,
%as long as the sensor types are trained, 
and still outperforms DAM which is specifically trained on partial data of scene (c). Therefore, the generalization of employing an explicit interpretable solver is verified. 

\begin{figure*}[t]
\centering
\includegraphics[width=\textwidth]{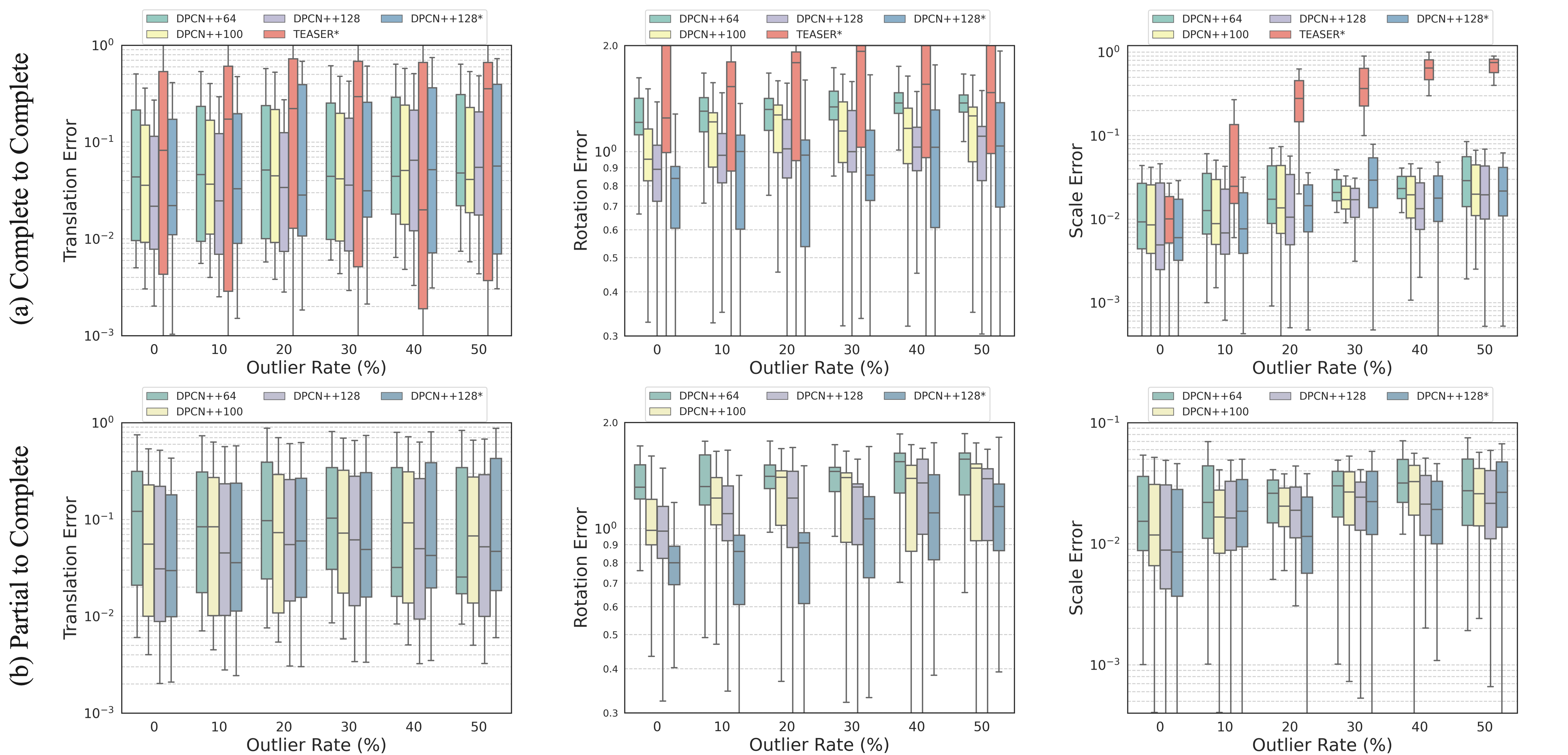}
\caption{\textbf{Quantitative Comparison on Heterogeneous Registration} of point cloud to mesh. The result of ``Partial to Partial'' is demonstrated in the Appendix \ref{subsec: append Quant on mvp}.}
\label{fig:pc mesh quant}
\vspace{-12pt}
\end{figure*}

\begin{figure*}[t]
\centering
\includegraphics[width=\textwidth]{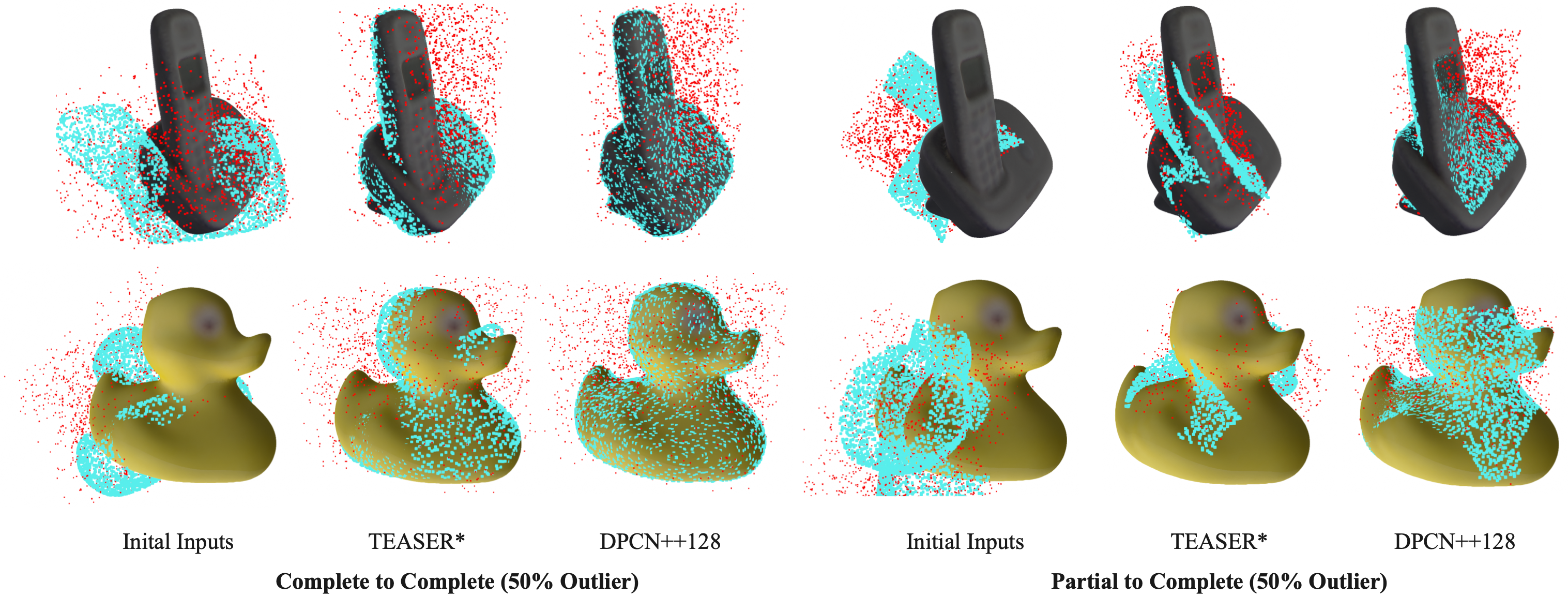}
\caption{\textbf{Qualitative Demonstration on Heterogeneous Registration} of point cloud to mesh. Blue points stands for the source point cloud and red are the outliers.}
\label{fig:pc mesh demo}
\vspace{-15pt}
\end{figure*}

\subsection{3D-3D Pose Registration}

\boldparagraph{Datasets} For homogeneous registration, we evaluate DPCN++ and the baselines on two different datasets, an object-level dataset MVP Benchmark~\cite{pan2021variational}, and a scene-level dataset 3DMatch~\cite{zeng20173dmatch}. The MVP Benchmark contains 6400 training pairs and 1200 testing pairs. The 3DMatch dataset consists of 54 scenes for training and 8 scenes for testing. For both datasets, we apply random $7DoF$ relative transformation between inputs pairs. Based on the completeness of measurement, three scenarios are tested including complete-to-complete, partial-to-complete, and partial-to-partial.

%We further evaluate  heterogeneous registration on two other datasets, ``Linemod''~\cite{hinterstoisser2012model} and ``RIRE''~\cite{MRICT_dataset_paper_RIRE}. 
For heterogeneous registration,  we first construct a heterogeneous dataset based on Linemod~\cite{hinterstoisser2012model} as a proof of concept. By extracting point cloud and building SDF from the mesh provided by Linemod, we evaluate versatile pose registration across these three modalities. We follow the settings of homogeneous registration in relative pose generation and scenarios. More detailed settings are elaborated in the Appendix \ref{append:3dhetero_setup}.

We further evaluate heterogeneous registration using medical imaging dataset, including CT-MRI pairs of RIRE~\cite{MRICT_dataset_paper_RIRE}, and CT-Ultrasound pairs on ``USCT''~\cite{3dusctDataset}. RIRE provides registered CT and MRI images of the human and ``USCT'' provides CT and Ultrasound images of humans spines, canine spines, and lamb spines. For all datasets, we randomly initialize relative transformations for each inputs pairs following the rules in Sec~\ref{subsubsec:3D Homogeneous Registration}. More details regarding the settings are provided in Appendix \ref{append:3dmedical_setup}.

\begin{figure*}[t]
\centering
\includegraphics[width=\textwidth]{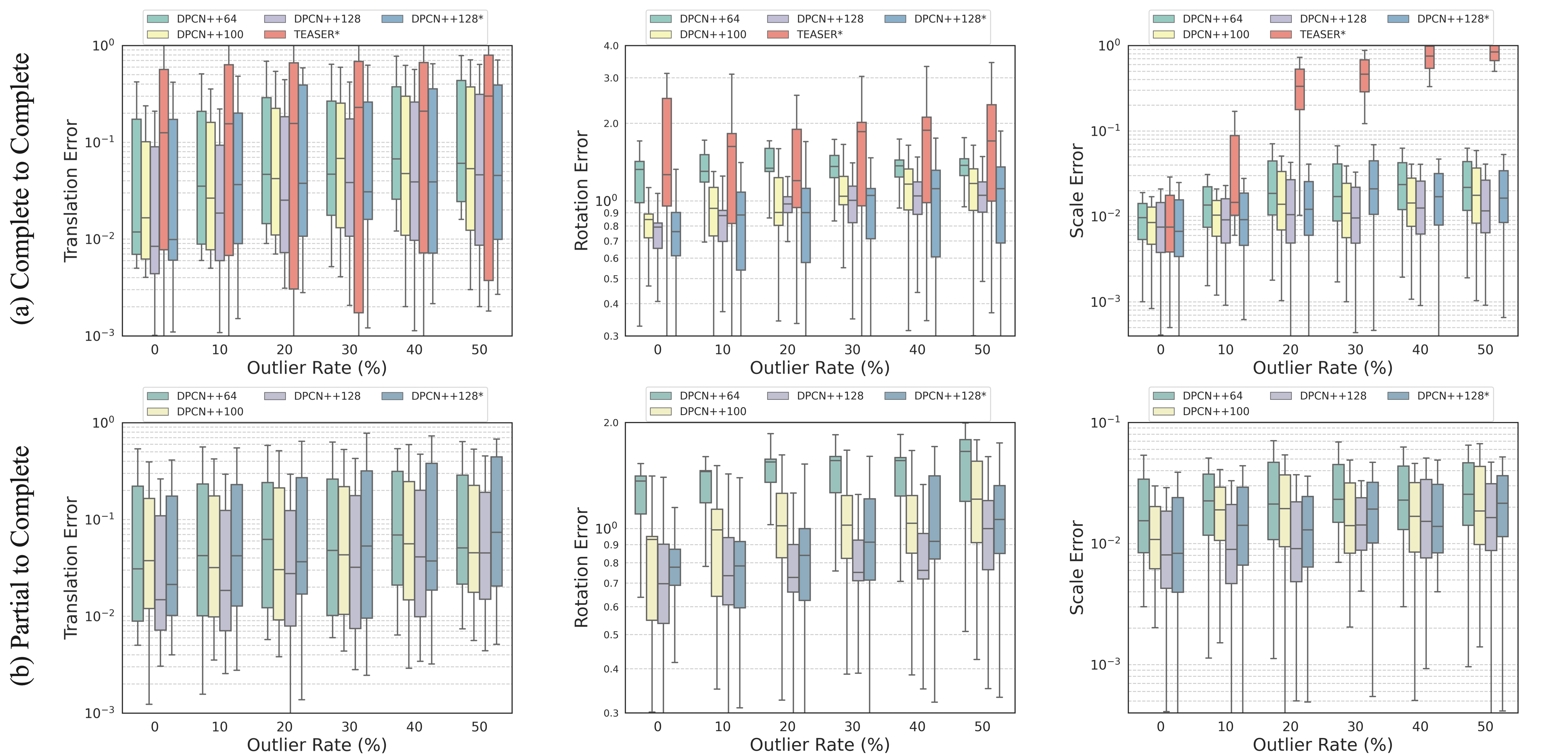}
\caption{\textbf{Quantitative Comparison on Heterogeneous Registration} of point cloud to SDF.}
\label{fig:pc sdf quant}
\vspace{-15pt}
\end{figure*}

\begin{figure}[t]
\centering
\includegraphics[width=\linewidth]{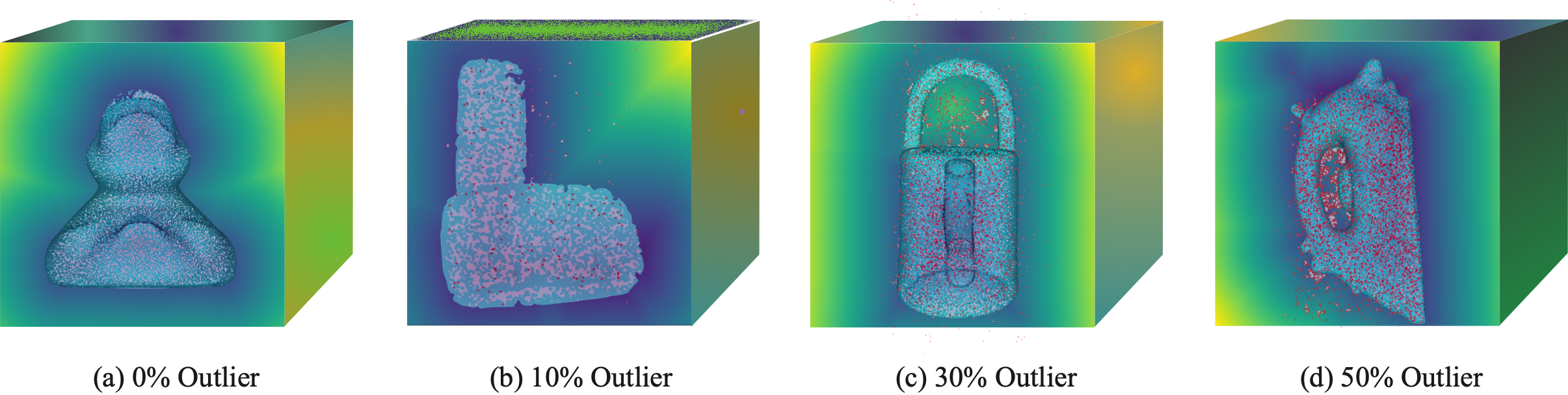}
\caption{\textbf{Qualitative Comparison on Heterogeneous Registration} of point cloud to SDF.}
\label{fig:pc sdf demo}
\vspace{-15pt}
\end{figure}

\begin{figure*}[t]
\centering
\includegraphics[width=\textwidth]{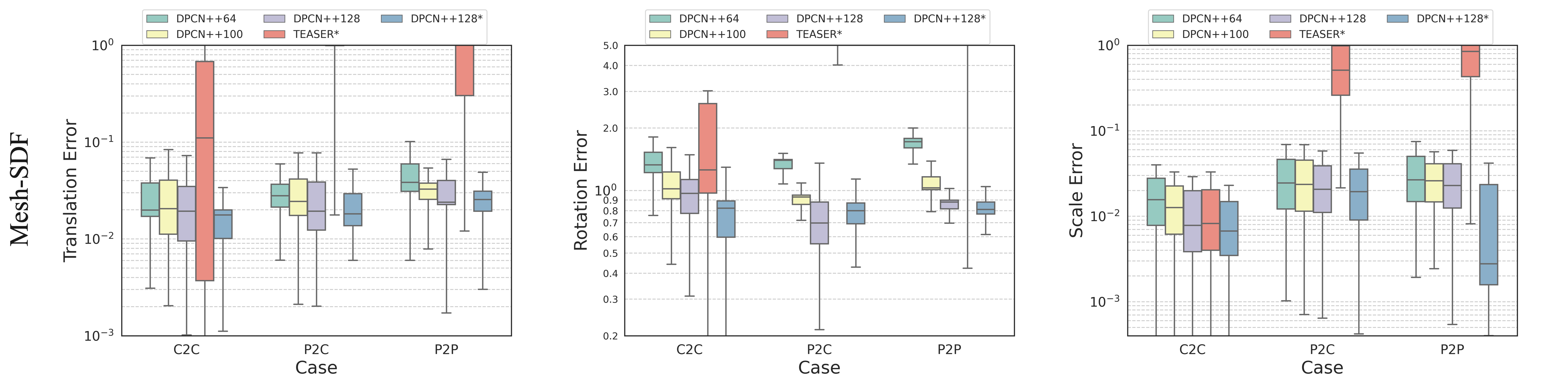}
\caption{\textbf{Quantitative Comparison on Heterogeneous Registration} of mesh to SDF.}
\label{fig:mesh sdf quant}
\vspace{-15pt}
\end{figure*}

\begin{figure}[t]
\centering
\includegraphics[width=\linewidth]{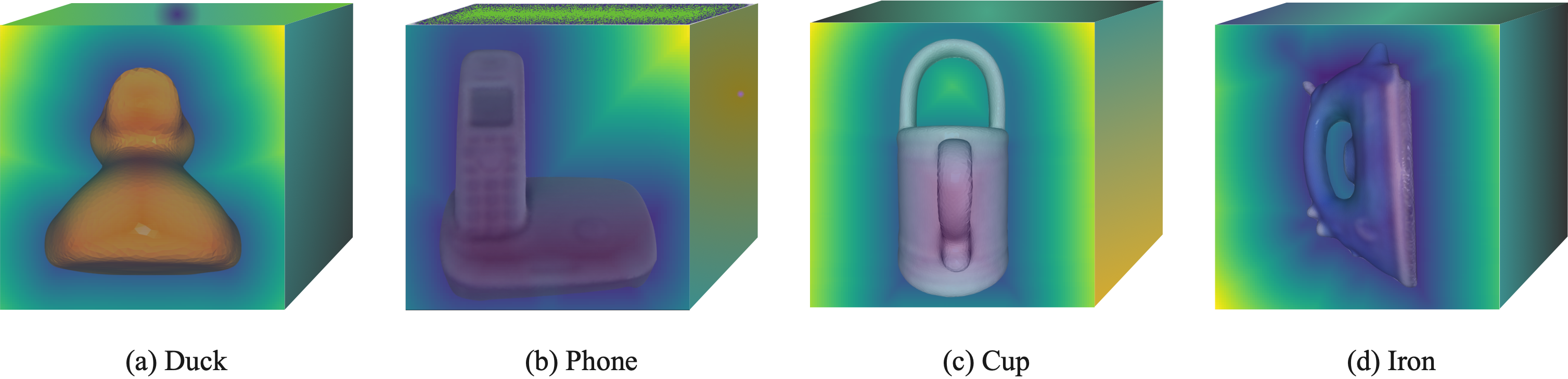}
\caption{\textbf{Qualitative Comparison on Heterogeneous Registration} of mesh to SDF.}
\label{fig:mesh sdf demo}
\vspace{-15pt}
\end{figure}

\boldparagraph{Baselines} We  compare homogeneous registration performance of DPCN++ to three lines of methods: 1) Learning-free methods: PC, the learning-free version of our approach by eliminating the feature extraction network but for $7DoF$ pose registration. TEASER~\cite{yang2020teaser}, a state-of-the-art $7DoF$ robust point cloud registration method with superior performance in handling large outlier rates. We also compare TEASER with certification (CERT)~\cite{yang2020teaser}, an improved version of TEASER at the cost of increasing the inference time. For correspondences of TEASER, we utilize the FPFH introduced in~\cite{yang2020teaser} following the original implementation.
2) Correspondence-based learning methods: DCP~\cite{wang2019deep} and DeepGMR~\cite{yuan2020deepgmr}, both methods overcome the local convergence of ICP by learning the feature correspondences. 3) Correspondence-free learning methods: PointNetLK~\cite{aoki2019pointnetlk}, a method employs a gradient based solver to minimize the similarity between learned features. Note that DCP, DeepGMR and PointNetLK are designed for $6DoF$ pose registration, thus we provide them the ground truth scale, and only evaluate rotation and translation. For all methods, we do not apply post-processing like ICP.

\begin{table}[t]
\centering
\caption{The evluation results on 3DMatch dataset.}
\label{tab:3dmatch_results}
\resizebox{\linewidth}{!}{%
\begin{tabular}{@{}llllllllll@{}}
\toprule[1pt]
\multicolumn{1}{c}{} & \multicolumn{8}{c}{Scenes} \\ \cmidrule(l){2-8} 
\multicolumn{1}{c}{Baselines} &
  \multicolumn{1}{c}{\rotatebox[origin=c]{90}{Home 1}} &
  \multicolumn{1}{c}{\rotatebox[origin=c]{90}{Home 2}} &
  \multicolumn{1}{c}{\rotatebox[origin=c]{90}{Hotel 1}} &
  \multicolumn{1}{c}{\rotatebox[origin=c]{90}{Hotel 2}} &
  \multicolumn{1}{c}{\rotatebox[origin=c]{90}{Hotel 3}} &
  \multicolumn{1}{c}{\rotatebox[origin=c]{90}{Study}} &
  \multicolumn{1}{c}{\rotatebox[origin=c]{90}{MIT Lab}} &
  \multicolumn{1}{c}{\rotatebox[origin=c]{90}{$\;$Time (ms)}} \\ \midrule
TEASER~\cite{yang2020teaser}               & 92.9  & 86.5 & \underline{97.8} & 89.4 & 94.4 & 91.1 & 83.1 & 39 \\
TEASER (CERT)       & 94.1  & 88.7 & \textbf{98.2} & 91.9 & 94.4 & \underline{94.3} & 88.6 & $\gg 1000$ \\
DCP~\cite{wang2019deep}                    & 85.1  & 84.9 & 86.2 & 89.3 & 80.5 & 79.6 & 85.4 & 73 \\
DeepGMR~\cite{yuan2020deepgmr}        & 83.5  & 88.1 & 82.2 & 85.8 & 83.1 & 81.3 & 81.7 & 63 \\
PNLK~\cite{aoki2019pointnetlk}                  & 81.2 & 82.9 & 81.5 & 86.3 & 79.1 & 77.4 & 80.5 & $> 200$\\
DPCN++64              & 93.5  & 91.2 & 94.7 & 93.1 & 95.7 & 92.3 & 92.1 &  \textbf{17}\\
DPCN++100             & \underline{94.4}  & \underline{92.2} & 95.6 & \underline{95.3} & \underline{96.2} & 94.1 & \underline{92.6} &  \underline{23}\\
DPCN++128             & \textbf{96.2}  & \textbf{94.1} & 96.1 & \textbf{95.9} & \textbf{97.3} & \textbf{94.7} & \textbf{93.5} &  37\\ \bottomrule[1pt]
\end{tabular}%
}
\vspace{-15pt}
\end{table}

\boldparagraph{Evaluation Metrics} We follow the metrics for evaluating 3D poses in \cite{wang2019deep}, \cite{yang2020teaser}, \cite{yuan2020deepgmr}, with the error for translation $E_{\boldsymbol{t}}$, rotation $E_{\boldsymbol{r}}$ and scale $E_{\mu}$ defined as:
\begin{gather}
    E_{\boldsymbol{t}} = \|\hat{\boldsymbol{t}}-\boldsymbol{t}^{*}\| \\
    E_{\boldsymbol{r}} = | \arccos{\frac{tr(\hat{\boldsymbol{r}}^{T}\boldsymbol{r}^{*})-1}{2}} | \\
    E_{\mu} = \|\hat{\mu}-\mu^{*}\|,
\label{eq: Err_3D}
\end{gather}
where $\hat{\cdot}$ and $\cdot^{*}$ are the estimated result and the ground truth of $\cdot$, respectively. 

\subsubsection{Homogeneous Registration}

\label{subsubsec:3D Homogeneous Registration}
\boldparagraph{Object-Level Registration} 
%For point cloud registration, we test the performance of both object level registration on the ``MVP Benchmark'' and scene level registration on ``3DMatch''.
%
%\noindent\textsc{\textbf{Setup:}} 
The MVP Benchmark contains complete point clouds and the corresponding partial version, we thus consider three scenarios for object-level point cloud registration: i) complete-to-complete, ii) partial-to-complete, and iii) partial-to-partial. The target relative pose is generated by randomly sampling from $\|\boldsymbol{t}\|\leq1m$, $\boldsymbol{r}\in SO(3)$, and $\mu \in [0.8,1.2]$. To evaluate the robustness against noisy input, we also add random outliers to one of the point clouds. The outlier rate ranges from $0\%$ to $50\%$ with an increase of $10\%$ at a time.

%Denote the aligned complete and two partial point clouds as $v_{c}$, $v_{p1}$, $v_{p2}$. We randomly generate their relative transformation $\boldsymbol{T}_{\boldsymbol{t},\boldsymbol{r},\mu}$ where $\|\boldsymbol{t}\|\leq1m, \boldsymbol{r}\in SO(3), \mu \in [0.8,1.2]$, such that in three cases $\{v_{c1}=\boldsymbol{T}_{\boldsymbol{t},\boldsymbol{r},\mu}v_{c2}\:|\:v_{p1}=\boldsymbol{T}_{\boldsymbol{t},\boldsymbol{r},\mu}v_{c}\:|\:v_{p2}=\boldsymbol{T}_{\boldsymbol{t},\boldsymbol{r},\mu}v_{p1}\}$. 

%\noindent\textsc{\textbf{Result:}} 
Fig.~\ref{fig:pc pc quant} shows the translation, rotation and scale error \wrt varying outlier rate for three scenarios. DPCN++ is evaluated at three different bandwidths $\{64,100,128\}$, named DPCN++64, DPCN++100 and DPCN++128.
%, which leads to different resolution and precision of the results. 
Fig.~\ref{fig:pc pc quant} (a) shows that when registering two complete point clouds, DPCN++ is competitive with the state-of-the-art method TEASER given various outlier rates. The learning-free PC at bandwidth 128 (PC128) also shows similar performance throughout different rates of outliers. These results demonstrate the advantage of the global convergent solver in the initialization-free setting when the data is completed.

However, when the source or the target is partially observed as shown in Fig.~\ref{fig:pc pc quant} (b) and Fig.~\ref{fig:pc pc quant} (c), the result is different. Both TEASER and PC have degenerated performance, indicating their sensitivity to completeness. The missing part of the source makes the corresponding part of the target become significant outliers. In contrast, DPCN++ retains a similar performance compared to the complete-to-complete setting and achieves the best performance, validating the effectiveness of the deep feature extraction network. By training end-to-end guided by the differentiable solver, the network is capable of filtering the outliers and enhancing crucial features for registration. For other methods with deep feature extractor including DCP, DeepGMR and PNKL, their performances also degenerate less than TEASER and PC when the point cloud is not complete. However, these methods struggle to achieve low error based on their own solvers, reflecting the advantage of the global convergent and correspondence-free phase correlation solver used in DCPN++. 
The qualitative results in Fig.~\ref{fig:pc pc demo} lead to consistent conclusions.

%The processing time of each baselines in the Appendix~\ref{subsec: append Quant on 3D Cases} (Fig.~\ref{fig:time_consume}) showing that the DPCN++ is able to run in real-time even when the bandwidth is $128$.

%We provide each case with one example of $0\%$ outlier and one example of $50\%$ outliers. It can be seen that the randomly generated noise will effect the quality of the registration but in a small extend.

\boldparagraph{Scene-Level Registration} 
We follow the same way as in object-level registration to generate the target relative pose between a pair of point clouds, which is also the same settings in \cite{yang2020teaser}. We do not add additional random outliers as the point clouds are collected in the real world and contains sensor noises. We employ 500 pairs in the Kitchen scene for training and leave other scenes for testing. We follow the same metrics as in \cite{yang2020teaser} that the registration is success when 1) rotation error smaller that $10^{\circ}$, and 2) translation error less than $30cm$.

Tab.~\ref{tab:3dmatch_results} shows the success rate of all methods, upon which the results of TEASER and TEASER (CERT) are taken from \cite{yang2020teaser}. It can be seen that our method performs the best on most scenes. TEASER is also highly competitive on this dataset as the overlap between partial observations is larger compared to the object-level registration, thus TEASER is less affected by the completeness of the point cloud.  TEASER (CERT) raises a higher accuracy with certification, but at the cost of much more computation time. Compared with other learning-based methods, our method still achieves higher accuracy owing to the solver. Fig.~\ref{fig:3dmatch demo} shows the cases for qualitative evaluation. Considering that only one threshold may not fully reflect the registration error, we additionally report results evaluated with varying thresholds in Appendix~\ref{subsec: append Quant on 3DMatch}.

% Apart from the standard thresholds we adopted that defines ``success'', we also \jc{elaborate upon the success rate of more thresholds} in Appendix~\ref{subsec: append Quant on 3D Cases}. One can see that DPCN++ has high success rate even when the error threshold is low, showing consistent result with that in object-level registration. \yl{Underlying logic not clear. Say that the thresholds are chosen following Teaser. As shown in the qualitative results, the thresholds are relatively large to reflect subtle registration error. We therefore additionally report results evaluated with lower thresholds in the Appendix.}

\subsubsection{Heterogeneous Registration}

\boldparagraph{Point Cloud-Mesh-SDF} To verify the heterogeneous $7DoF$ registration, we carry out experiments across 3D representations including i) point cloud, ii) mesh and iii) Signed Distance Field (SDF). The relative pose between the two measurements is generated following the settings in homogeneous registration.
We take the best performing baseline among all sub-second methods, TEASER, for comparison. As TEASER is not directly applicable for heterogeneous registration, we convert mesh and SDF to point cloud to evaluate it in the homogeneous setting. We denote this baseline as TEASER*. Moreover, we also consider a baseline DPCN++* which also takes a pair of point clouds as input. Note that both TEASER* and DPCN++* assume the conversion between different types of representations are known, while the conversion is non-trivial in many applications, \eg, medical imaging.

\begin{figure*}[t]
\centering
\includegraphics[width=\textwidth]{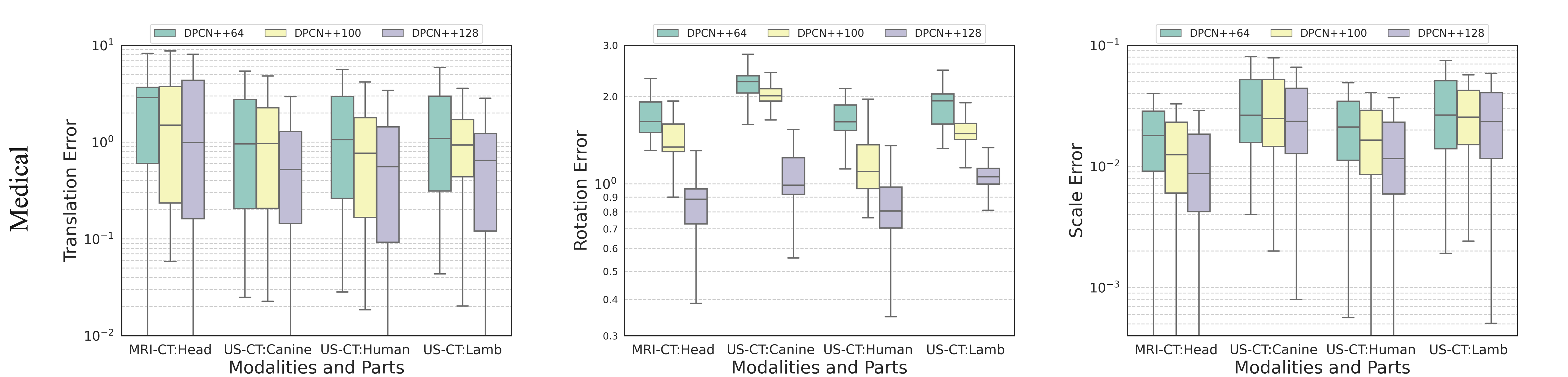}
\caption{\textbf{Quantitative Comparison on Heterogeneous Registration} of medical images.}
\label{fig:medical quant}
\end{figure*}

\begin{figure*}[t]
\centering
\includegraphics[width=\linewidth]{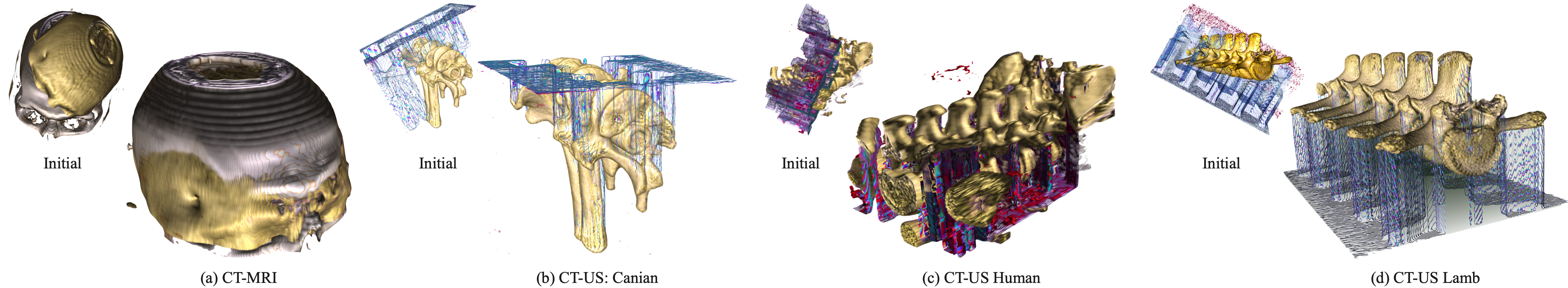}
\caption{\textbf{Qualitative Comparison on Heterogeneous Registration} of medical images.}
\label{fig:medical demo}
\end{figure*}

Fig.~\ref{fig:pc mesh quant}, Fig.~\ref{fig:pc sdf quant}, and Fig.~\ref{fig:mesh sdf quant} show the translation, rotation and scale error cloud to mesh, point cloud to SDF, and mesh to SDF registration respectively. Note that in all scenarios, DPCN++ achieves similar performance to that of DPCN++* without applying the conversion in prior. Instead, the common features from heterogeneous input can be built by the extraction network. This result verifies the usefulness of heterogeneous registration in practice. In addition, as the result of homogeneous registration, DPCN++ in complete-to-complete scenario has smaller errors than those of the partial-to-complete scenario by a small margin, while TEASER* has a large degeneration. The qualitative cases of the versatile registration are shown in Fig.~\ref{fig:pc mesh demo}, Fig.~\ref{fig:pc sdf demo} and Fig.~\ref{fig:mesh sdf demo}.

%\textit{Point cloud to mesh.}   Together with the quantitative results, we make the following observations. i) The average performance of registration will gradually deteriorate (slightly) as the rate of outliers rises. However, this trend is not significant in the translation stage, showing that the discrete Fourier transform and correlation is not as sensitive to such noise as the spherical Fourier transform and correlation. ii) The deterioration is obvious among $0\%$ to $10\%$ outliers. \textcolor{red}{not yet make the statement about the comparison to the TEASER* and DPCN*.}

%\textit{Point cloud to SDF.}  We visualize them by transforming the point clouds with the estimated pose and make it transparent with an alpha value of $0.5$. It can be seen that the registration is quite successful with lower oultier rate. However, in the demonstration of the $50\% Outlier$ the rotation estimation is slightly deviated from the ground truth.

%\textit{Mesh to SDF.}  Similar to the case above, we visualize this by transforming the meshes. The results follows the rule: the bigger the bandwidth, the better the accuracy.

\boldparagraph{CT-MRI-3D Ultrasound} We further evaluate the heterogeneous 3D registration with real medical images captured by CT, MRI and 3D ultrasound. In this task, the different modality captures the different part of the body, \eg, CT images usually reflects the structure of the bone while MRI, muscle and tissue. As the conversion between different modalities is unknown, we only evaluate DPCN++ in this problem. We focus on the rigid part, \ie bone, registration for medical imaging. Fig.~\ref{fig:medical quant} and Fig~\ref{fig:medical demo} show the quantitative and qualitative results. Note that even with large noise in 3D ultrasound for the human spine, a successful registration between 3D ultrasound and CT is achieved. These results again show that by training DPCN++ end-to-end, the feature extractors are able to learn common features from data to enable challenging heterogeneous registration, demonstrating the value of DPCN++ in real applications.
%which assures that the capability of the heterogeneous registration is not trivial for real applications. 

%\textit{CT to MRI.} The value range of pixels in CT images usually reflects the structure of the bone while the range in MRI reflects the muscle and tissue. The concept of registering CT and MRI is applicable since most muscles and tissues are built around the bone. Therefore, even different in the range of values, the correspondences between CT and MRI should still be found with trained feature extractors.  

%\textit{CT to 3D ultrasound.} Unlike the MRI, 3D ultrasound reflects the 3D shapes of the interested surface inside the patients' body. The data of 3D ultrasound contains much noises than CT, making it even more difficult to be pose aligned with CT. Such phenomenon can be interpreted by the evaluation metrics in Fig~\ref{fig:medical quant}, where the error for ``CT to 3D ultrasound'' registration is a bit higher than that of the ``CT to MRI''. Also, it can be seen that, among all the ``CT to 3D ultrasound'' cases, the human spine marks the better result due to the remarkable structure. Still, even facing much more noise as in CT-US registration, DPCN++ can still handle the problem successfully with low error.

\subsection{Ablation Study}

\boldparagraph{Bandwidth}An important factor to tune DPCN++ is the bandwidth. We compare DPCN++ with the bandwidth of $\{64,100,128\}$ in both homogeneous and heterogeneous pose registration tasks above. All results show that a higher bandwidth brings a better accuracy. This result is intuitive since a higher bandwidth gives a finer resolution of pose estimation.

%\boldparagraph{Feature extraction}Another factor in DPCN++ is the necessity of deep feature extraction. An ablated version of DPCN++ without network, PC128, is also compared in all tasks above. The result is that the feature extraction is able to significantly improves the accuracy when the input is not completed. For the heterogeneous, the necessity of deep feature extraction is obvious.

%We conduct ablation studies to verify the functions of specific modules in DPCN++. First, we verify the usage of U-Net in the point cloud settings by comparing DPCN++128 and PC128. Specifically, we train DPCN++128 in two ways: i) with one UNet (DPCN++128(1U)) in which case the source and target are fed into a shared U-Net, and ii) with two independent U-Nets for the source and target (DPCN++128 (2U)). 

%The results in Fig.~\ref{fig:ablation}(a) shows that when not providing any training, the traditional phase correlation is hindered in the settings of partial point cloud registration, while the trainable version gives satisfying results. Moreover, when focusing on the results of DPCN++128 (1U) and DPCN++128 (2U), we find that DPCN++128 (1U) is relatively more precise, which can be explained that when the inputs are of the same modality, the shared feature extractor is more likely to learn the common feature.

\boldparagraph{Solver Architecture} We further focus on the solver architecture by evaluating the necessity of decoupling translation and scale from the rotation. We compare DPCN++ with and without the step of building translation invariance. Theoretically, DPCN++ without translation invariance (DPCN w/o t-) fails when initial translation error is large. As a reference, we also add the result of PC128. Both versions of DPCN++ are trained on the dataset with $0cm$ relative translation, but evaluated on dataset with various relative translation. 

\begin{figure}[t]
\centering
\includegraphics[width=\linewidth]{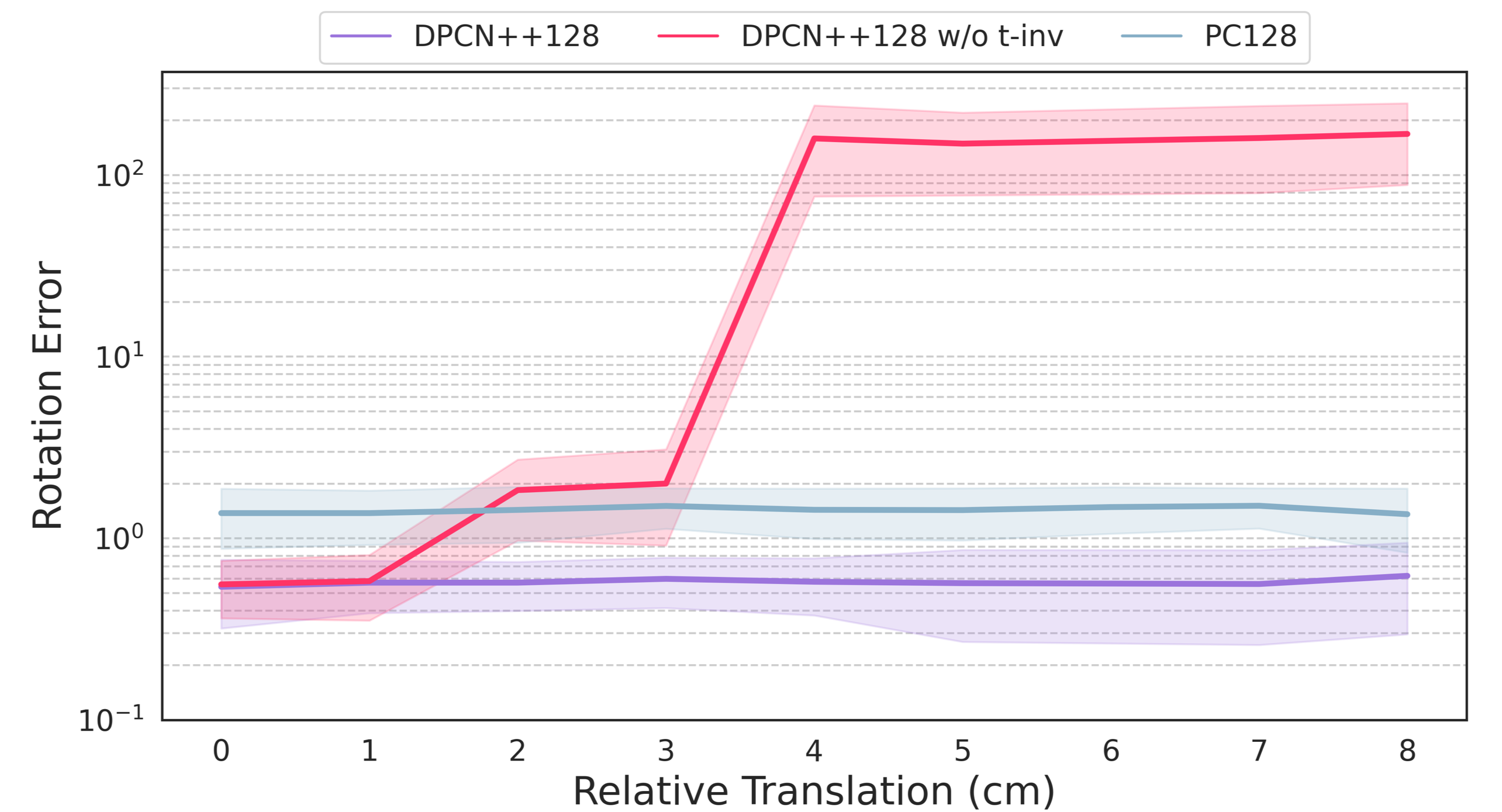}
\caption{\textbf{Quantitative Comparison for Ablation Study.} The translation in the horizontal axis stands for the initial translation between the source and the target.}
\label{fig:ablation}
\vspace{-15pt}
\end{figure}

In Fig.~\ref{fig:ablation}, while DPCN++ gives similar performance with unseen growing translation, DPCN++ without translation invariance highly depends on the relative translation. Despite achieving the same performance given small relative translation, it fails soon when the translation grows. On the other hand, PC128 shows similar performance with growing translation but has almost consistently larger error comparing to DPCN++ for all relative translation. This result demonstrates the superiority of DPCN++ by incorporating strong inductive bias with the carefully designed solver architecture, where this inductive bias is hard to be learned purely from data.

%We provide PHASER with the bandwidth up to 128 while testing the performance of DPCN++ in three different bandwidths: $\{64,100,128\}$.

\subsection{Limitation}
\label{subsec:limitation}
Though DPCN++ is shown to be versatile, robust, and effective in the experiments, the main bottleneck of the method is the intensive memory consumption. In this paper, we choose the UNet as the deep feature extractor, which limits the bandwidth of the DPCN++ up to $128$ while taking two RTX3090 for training. We believe that with more light-weight feature extractors to be proposed, and with the advancement of the hardware, the limitation can be relieved in the future. We also report a representative failure case in Appendix~\ref{subsec: append Visual on 3D Cases}, which reveals the fact DPCN++ might be degraded to the inputs with relatively small overlaps. The authors believe that with attention involves, the degeneration brought by small overlaps can be further reduced.

\section{Conclusion}
\label{subsec:Conclusion}

We present a general end-to-end feature grid registration framework, DPCN++, for versatile pose registration problems. In DPCN++, the core component is a differentiable, global convergent, correspondence-free phase correlation solver that enables back-propagating the pose error to the learnable feature extractors, which addresses the outlier sensitivity of correspondence-based methods, and the local convergence of previous correspondence-free methods. With the DPC solver, it is unnecessary to define the specific form of feature, which is learned fully driven by data. We evaluate DPCN++ on both 2D and 3D registration tasks including BEV images, object and scene level 3D measurements, as well as medical images. All results show that our method is able to register both homogeneous and heterogeneous sensor measurements with competitive performance, verifying the strong inductive bias brought by the DPC solver.

In the future, we would like to investigate the shape deformation in the pose registration tasks to address the problem when there is no model exactly the same to the given sensor measurement.

% % use section* for acknowledgment
%\section{Acknowledgment}
%\label{subsec:Acknowledgment}

%\textcolor{red}{The authors would like to thank...}

\ifCLASSOPTIONcaptionsoff
  \newpage
\fi

%\printbibliography



%\bibliographystyle{IEEEtran}

{
\bibliographystyle{IEEEtranS}
\bibliography{IEEEtran/bibliography_short,IEEEtran/DPCN++}

% Generated by IEEEtranS.bst, version: 1.14 (2015/08/26)
\begin{thebibliography}{10}
\providecommand{\url}[1]{#1}
\csname url@samestyle\endcsname
\providecommand{\newblock}{\relax}
\providecommand{\bibinfo}[2]{#2}
\providecommand{\BIBentrySTDinterwordspacing}{\spaceskip=0pt\relax}
\providecommand{\BIBentryALTinterwordstretchfactor}{4}
\providecommand{\BIBentryALTinterwordspacing}{\spaceskip=\fontdimen2\font plus
\BIBentryALTinterwordstretchfactor\fontdimen3\font minus
  \fontdimen4\font\relax}
\providecommand{\BIBforeignlanguage}[2]{{%
\expandafter\ifx\csname l@#1\endcsname\relax
\typeout{** WARNING: IEEEtranS.bst: No hyphenation pattern has been}%
\typeout{** loaded for the language `#1'. Using the pattern for}%
\typeout{** the default language instead.}%
\else
\language=\csname l@#1\endcsname
\fi
#2}}
\providecommand{\BIBdecl}{\relax}
\BIBdecl

\bibitem{aoki2019pointnetlk}
Y.~Aoki, H.~Goforth, R.~A. Srivatsan, and S.~Lucey, ``Pointnetlk: Robust \&
  efficient point cloud registration using pointnet,'' in \emph{CVPR}, 2019.

\bibitem{Barnes2019}
D.~Barnes, R.~Weston, and I.~Posner, ``{Masking by Moving: Learning
  Distraction-Free Radar Odometry from Pose Information},'' in \emph{CoRL},
  2019.

\bibitem{bernreiter2021phaser}
L.~Bernreiter, L.~Ott, J.~Nieto, R.~Siegwart, and C.~Cadena, ``Phaser: A robust
  and correspondence-free global pointcloud registration,'' \emph{IEEE Robotics
  and Automation Letters}, vol.~6, no.~2, pp. 855--862, 2021.

\bibitem{besl1992method}
P.~J. Besl and N.~D. McKay, ``Method for registration of 3-d shapes,'' in
  \emph{Sensor fusion IV: control paradigms and data structures}, 1992.

\bibitem{bulow2018scale}
H.~B{\"u}low and A.~Birk, ``Scale-free registrations in 3d: 7 degrees of
  freedom with fourier mellin soft transforms,'' \emph{IJCV}, vol. 126, no.~7,
  pp. 731--750, 2018.

\bibitem{3dslicer}
.~BWH and D.~Slice. (2020) 3d-slicer.

\bibitem{cao2018deep}
X.~Cao, J.~Yang, L.~Wang, Z.~Xue, Q.~Wang, and D.~Shen, ``Deep learning based
  inter-modality image registration supervised by intra-modality similarity,''
  in \emph{MLMI Workshop}, 2018.

\bibitem{chen2008multimodal}
Y.-W. Chen, C.-L. Lin, and A.~Mimori, ``Multimodal medical image registration
  using particle swarm optimization,'' in \emph{ISDA}, 2008.

\bibitem{chen2009hybrid}
Y.-W. Chen, A.~Mimori, and C.-L. Lin, ``Hybrid particle swarm optimization for
  3-d image registration,'' in \emph{ICIP}, 2009.

\bibitem{chen2020deep}
Z.~Chen, X.~Xu, Y.~Wang, and R.~Xiong, ``Deep phase correlation for end-to-end
  heterogeneous sensor measurements matching,'' in \emph{CoRL}, 2020.

\bibitem{chetverikov2005robust}
D.~Chetverikov, D.~Stepanov, and P.~Krsek, ``Robust euclidean alignment of 3d
  point sets: the trimmed iterative closest point algorithm,'' \emph{IVC},
  vol.~23, no.~3, pp. 299--309, 2005.

\bibitem{choy2020deep}
C.~Choy, W.~Dong, and V.~Koltun, ``Deep global registration,'' in \emph{CVPR},
  2020.

\bibitem{choy2019fully}
C.~Choy, J.~Park, and V.~Koltun, ``Fully convolutional geometric features,'' in
  \emph{Proceedings of the IEEE/CVF International Conference on Computer
  Vision}, 2019, pp. 8958--8966.

\bibitem{cohen2018spherical}
T.~S. Cohen, M.~Geiger, J.~K{\"o}hler, and M.~Welling, ``Spherical cnns,''
  \emph{arXiv preprint arXiv:1801.10130}, 2018.

\bibitem{superpoint}
D.~DeTone, T.~Malisiewicz, and A.~Rabinovich, ``Superpoint: Self-supervised
  interest point detection and description,'' in \emph{CVPRW}, 2018.

\bibitem{dusmanu2019d2}
M.~Dusmanu, I.~Rocco, T.~Pajdla, M.~Pollefeys, J.~Sivic, A.~Torii, and
  T.~Sattler, ``D2-net: A trainable cnn for joint detection and description of
  local features,'' \emph{arXiv preprint arXiv:1905.03561}, 2019.

\bibitem{enqvist2009optimal}
O.~Enqvist, K.~Josephson, and F.~Kahl, ``Optimal correspondences from pairwise
  constraints,'' in \emph{ICCV}, 2009.

\bibitem{granger2002multi}
S.~Granger and X.~Pennec, ``Multi-scale em-icp: A fast and robust approach for
  surface registration,'' in \emph{ECCV}, 2002.

\bibitem{gross2019alignnet}
J.~Gro{\ss}, A.~O{\v{s}}ep, and B.~Leibe, ``Alignnet-3d: Fast point cloud
  registration of partially observed objects,'' in \emph{3DV}, 2019.

\bibitem{haskins2019learning}
G.~Haskins, J.~Kruecker, U.~Kruger, S.~Xu, P.~A. Pinto, B.~J. Wood, and P.~Yan,
  ``Learning deep similarity metric for 3d mr--trus image registration,''
  \emph{IJCARS}, vol.~14, no.~3, pp. 417--425, 2019.

\bibitem{healy2003ffts}
D.~M. Healy, D.~N. Rockmore, P.~J. Kostelec, and S.~Moore, ``Ffts for the
  2-sphere-improvements and variations,'' \emph{Journal of Fourier analysis and
  applications}, vol.~9, no.~4, pp. 341--385, 2003.

\bibitem{hinterstoisser2012model}
S.~Hinterstoisser, V.~Lepetit, S.~Ilic, S.~Holzer, G.~Bradski, K.~Konolige, and
  N.~Navab, ``Model based training, detection and pose estimation of
  texture-less 3d objects in heavily cluttered scenes,'' in \emph{ACCV}, 2012.

\bibitem{huang2020feature}
X.~Huang, G.~Mei, and J.~Zhang, ``Feature-metric registration: A fast
  semi-supervised approach for robust point cloud registration without
  correspondences,'' in \emph{CVPR}, 2020.

\bibitem{jiao2021r2d2loc}
Y.~Jiao, Y.~Wang, X.~Ding, B.~Fu, S.~Huang, and R.~Xiong, ``2-entity random
  sample consensus for robust visual localization: Framework, methods, and
  verifications,'' \emph{IEEE Transactions on Industrial Electronics}, vol.~68,
  no.~5, pp. 4519--4528, 2021.

\bibitem{Kaslin2016}
R.~Kaslin, P.~Fankhauser, E.~Stumm, Z.~Taylor, E.~Mueggler, J.~Delmerico,
  D.~Scaramuzza, R.~Siegwart, and M.~Hutter, ``{Collaborative localization of
  aerial and ground robots through elevation maps},'' \emph{SSRR 2016}, pp.
  284--290, 2016.

\bibitem{Kendall2017}
A.~Kendall and R.~Cipolla, ``{Geometric loss functions for camera pose
  regression with deep learning},'' in \emph{CVPR}, 2017.

\bibitem{Kim2018}
J.~Kim, J.~Kim, S.~Choi, M.~A. Hasan, and C.~Kim, ``{Robust template matching
  using scale-adaptive deep convolutional features},'' \emph{APSIPA ASC}, 2018.

\bibitem{kisaki2014high}
M.~Kisaki, Y.~Yamamura, H.~Kim, J.~K. Tan, S.~Ishikawa, and A.~Yamamoto, ``High
  speed image registration of head ct and mr images based on
  levenberg-marquardt algorithms,'' in \emph{SCIS and ISIS}, 2014.

\bibitem{Kummerle2010}
R.~K{\"u}mmerle, B.~Steder, C.~Dornhege, A.~Kleiner, G.~Grisetti, and
  W.~Burgard, ``Large scale graph-based slam using aerial images as prior
  information,'' \emph{AR}, vol.~30, no.~1, pp. 25--39, 2011.

\bibitem{lee2009learning}
D.~Lee, M.~Hofmann, F.~Steinke, Y.~Altun, N.~D. Cahill, and B.~Scholkopf,
  ``Learning similarity measure for multi-modal 3d image registration,'' in
  \emph{CVPR}, 2009.

\bibitem{li2021pointnetlk}
X.~Li, J.~K. Pontes, and S.~Lucey, ``Pointnetlk revisited,'' in
  \emph{Proceedings of the IEEE/CVF Conference on Computer Vision and Pattern
  Recognition}, 2021, pp. 12\,763--12\,772.

\bibitem{liao2017artificial}
R.~Liao, S.~Miao, P.~de~Tournemire, S.~Grbic, A.~Kamen, T.~Mansi, and
  D.~Comaniciu, ``An artificial agent for robust image registration,'' in
  \emph{AAAI}, 2017.

\bibitem{lowe2004sift}
D.~G. Lowe, ``Distinctive image features from scale-invariant keypoints,''
  \emph{IJCV}, vol.~60, no.~2, pp. 91--110, 2004.

\bibitem{Lu2019}
W.~Lu, Y.~Zhou, G.~Wan, S.~Hou, and S.~Song, ``{L3-net: Towards learning based
  lidar localization for autonomous driving},'' in \emph{CVPR}, 2019.

\bibitem{lucas1981iterative}
B.~D. Lucas, T.~Kanade \emph{et~al.}, ``An iterative image registration
  technique with an application to stereo vision,'' 1981.

\bibitem{ma2017multimodal}
K.~Ma, J.~Wang, V.~Singh, B.~Tamersoy, Y.-J. Chang, A.~Wimmer, and T.~Chen,
  ``Multimodal image registration with deep context reinforcement learning,''
  in \emph{MICCAI}, 2017.

\bibitem{makadia2006rotation}
A.~Makadia and K.~Daniilidis, ``Rotation recovery from spherical images without
  correspondences,'' \emph{PAMI}, vol.~28, no.~7, pp. 1170--1175, 2006.

\bibitem{3dusctDataset}
N.~Masoumi, C.~J. Belasso, M.~O. Ahmad, H.~Benali, Y.~Xiao, and H.~Rivaz,
  ``Multimodal 3d ultrasound and ct in image-guided spinal surgery: public
  database and new registration algorithms,'' \emph{IJCARS}, pp. 1--11, 2021.

\bibitem{regression_tmi}
S.~Miao, Z.~J. Wang, and R.~Liao, ``A cnn regression approach for real-time
  2d/3d registration,'' \emph{T-MI}, vol.~35, no.~5, pp. 1352--1363, 2016.

\bibitem{pais20203dregnet}
G.~D. Pais, S.~Ramalingam, V.~M. Govindu, J.~C. Nascimento, R.~Chellappa, and
  P.~Miraldo, ``3dregnet: A deep neural network for 3d point registration,'' in
  \emph{CVPR}, 2020.

\bibitem{pan2019multi}
J.~Pan, Z.~Min, A.~Zhang, H.~Ma, and M.~Q.-H. Meng, ``Multi-view global 2d-3d
  registration based on branch and bound algorithm,'' in \emph{ROBIO}, 2019.

\bibitem{pan2021variational}
L.~Pan, X.~Chen, Z.~Cai, J.~Zhang, H.~Zhao, S.~Yi, and Z.~Liu, ``Variational
  relational point completion network,'' in \emph{CVPR}, 2021.

\bibitem{pan2020gem}
Y.~Pan, X.~Xu, X.~Ding, S.~Huang, Y.~Wang, and R.~Xiong, ``Gem: online globally
  consistent dense elevation mapping for unstructured terrain,'' \emph{IEEE
  Transactions on Instrumentation and Measurement}, vol.~70, pp. 1--13, 2020.

\bibitem{Pan2019}
Y.~Pan, X.~Xu, Y.~Wang, X.~Ding, and R.~Xiong, ``{GPU accelerated real-time
  traversability mapping},'' in \emph{ROBIO}, 2019.

\bibitem{Park2020}
J.~H. Park, W.~J. Nam, and S.~W. Lee, ``{A two-stream symmetric network with
  bidirectional ensemble for aerial image matching},'' \emph{Remote Sensing},
  2020.

\bibitem{qi2017pointnet}
C.~R. Qi, H.~Su, K.~Mo, and L.~J. Guibas, ``Pointnet: Deep learning on point
  sets for 3d classification and segmentation,'' in \emph{CVPR}, 2017.

\bibitem{revaud2019r2d2}
J.~Revaud, C.~De~Souza, M.~Humenberger, and P.~Weinzaepfel, ``R2d2: Reliable
  and repeatable detector and descriptor,'' \emph{NIPS}, 2019.

\bibitem{Ruchti2015}
P.~Ruchti, B.~Steder, M.~Ruhnke, and W.~Burgard, ``{Localization on
  OpenStreetMap data using a 3D laser scanner},'' in \emph{ICRA}, 2015.

\bibitem{rusinkiewicz2019symmetric}
S.~Rusinkiewicz, ``A symmetric objective function for icp,'' \emph{ACM
  Transactions on Graphics (TOG)}, vol.~38, no.~4, pp. 1--7, 2019.

\bibitem{rusinkiewicz2001efficient}
S.~Rusinkiewicz and M.~Levoy, ``Efficient variants of the icp algorithm,'' in
  \emph{Proceedings third international conference on 3-D digital imaging and
  modeling}.\hskip 1em plus 0.5em minus 0.4em\relax IEEE, 2001, pp. 145--152.

\bibitem{saiti2020application}
E.~Saiti and T.~Theoharis, ``An application independent review of multimodal 3d
  registration methods,'' \emph{Computers \& Graphics}, vol.~91, pp. 153--178,
  2020.

\bibitem{sarode2019pcrnet}
V.~Sarode, X.~Li, H.~Goforth, Y.~Aoki, R.~A. Srivatsan, S.~Lucey, and
  H.~Choset, ``Pcrnet: Point cloud registration network using pointnet
  encoding,'' \emph{arXiv preprint arXiv:1908.07906}, 2019.

\bibitem{schwab2015multimodal}
L.~Schwab, M.~Schmitt, and R.~Wanka, ``Multimodal medical image registration
  using particle swarm optimization with influence of the data's initial
  orientation,'' in \emph{CIBCB}, 2015.

\bibitem{sedghi2018semi}
A.~Sedghi, J.~Luo, A.~Mehrtash, S.~Pieper, C.~M. Tempany, T.~Kapur, P.~Mousavi,
  and W.~M. Wells~III, ``Semi-supervised deep metrics for image registration,''
  \emph{arXiv preprint arXiv:1804.01565}, 2018.

\bibitem{SrinivasaReddy1996}
B.~{Srinivasa Reddy} and B.~N. Chatterji, ``{An FFT-based technique for
  translation, rotation, and scale-invariant image registration},'' \emph{TIP},
  1996.

\bibitem{Tang2020}
T.~Y. Tang, D.~{De Martini}, D.~Barnes, and P.~Newman, ``{RSL-Net: Localising
  in Satellite Images from a Radar on the Ground},'' \emph{RA-L}, 2020.

\bibitem{wang2012local}
C.~Wang, X.~Jing, and C.~Zhao, ``Local upsampling fourier transform for
  accurate 2d/3d image registration,'' \emph{Computers \& Electrical
  Engineering}, vol.~38, no.~5, pp. 1346--1357, 2012.

\bibitem{wang2019densefusion}
C.~Wang, D.~Xu, Y.~Zhu, R.~Mart{\'\i}n-Mart{\'\i}n, C.~Lu, L.~Fei-Fei, and
  S.~Savarese, ``Densefusion: 6d object pose estimation by iterative dense
  fusion,'' in \emph{CVPR}, 2019.

\bibitem{wang2019deep}
Y.~Wang and J.~M. Solomon, ``Deep closest point: Learning representations for
  point cloud registration,'' in \emph{ICCV}, 2019.

\bibitem{wang2019dynamic}
Y.~Wang, Y.~Sun, Z.~Liu, S.~E. Sarma, M.~M. Bronstein, and J.~M. Solomon,
  ``Dynamic graph cnn for learning on point clouds,'' \emph{ACM Trans. on
  Graphics}, vol.~38, no.~5, pp. 1--12, 2019.

\bibitem{wells1996multi}
W.~M. Wells~III, P.~Viola, H.~Atsumi, S.~Nakajima, and R.~Kikinis,
  ``Multi-modal volume registration by maximization of mutual information,''
  \emph{Medical Image Analysis}, vol.~1, no.~1, pp. 35--51, 1996.

\bibitem{MRICT_dataset_paper_RIRE}
J.~West, J.~M. Fitzpatrick, M.~Y. Wang, B.~M. Dawant, C.~R. Maurer~Jr, R.~M.
  Kessler, R.~J. Maciunas, C.~Barillot, D.~Lemoine, A.~Collignon \emph{et~al.},
  ``Comparison and evaluation of retrospective intermodality brain image
  registration techniques,'' \emph{Journal of computer assisted tomography},
  vol.~21, no.~4, pp. 554--568, 1997.

\bibitem{weston2022fast}
R.~Weston, M.~Gadd, D.~De~Martini, P.~Newman, and I.~Posner, ``Fast-mbym:
  Leveraging translational invariance of the fourier transform for efficient
  and accurate radar odometry,'' \emph{arXiv preprint arXiv:2203.00459}, 2022.

\bibitem{3dunet}
A.~Wolny, L.~Cerrone, A.~Vijayan, R.~Tofanelli, A.~V. Barro, M.~Louveaux,
  C.~Wenzl, S.~Strauss, D.~Wilson-Sánchez, R.~Lymbouridou, S.~S. Steigleder,
  C.~Pape, A.~Bailoni, S.~Duran-Nebreda, G.~W. Bassel, J.~U. Lohmann,
  M.~Tsiantis, F.~A. Hamprecht, K.~Schneitz, A.~Maizel, and A.~Kreshuk,
  ``Accurate and versatile 3d segmentation of plant tissues at cellular
  resolution,'' \emph{eLife}, vol.~9, p. e57613, jul 2020.

\bibitem{wright2018lstm}
R.~Wright, B.~Khanal, A.~Gomez, E.~Skelton, J.~Matthew, J.~V. Hajnal,
  D.~Rueckert, and J.~A. Schnabel, ``Lstm spatial co-transformer networks for
  registration of 3d fetal us and mr brain images,'' in \emph{Data Driven
  Treatment Response Assessment and Preterm, Perinatal, and Paediatric Image
  Analysis}.\hskip 1em plus 0.5em minus 0.4em\relax Springer, 2018, pp.
  149--159.

\bibitem{xu2015empirical}
B.~Xu, N.~Wang, T.~Chen, and M.~Li, ``Empirical evaluation of rectified
  activations in convolutional network,'' \emph{arXiv preprint
  arXiv:1505.00853}, 2015.

\bibitem{xu2020collaborative}
X.~Xu, Z.~Chen, J.~Guo, Y.~Wang, Y.~Wang, and R.~Xiong, ``Collaborative
  localization of aerial and ground mobile robots through orthomosaic map,'' in
  \emph{2020 IEEE International Conference on Real-time Computing and Robotics
  (RCAR)}.\hskip 1em plus 0.5em minus 0.4em\relax IEEE, 2020, pp. 122--127.

\bibitem{yang2021certifiable}
H.~Yang and L.~Carlone, ``Certifiable outlier-robust geometric perception:
  Exact semidefinite relaxations and scalable global optimization,''
  \emph{arXiv preprint arXiv:2109.03349}, 2021.

\bibitem{yang2020teaser}
H.~Yang, J.~Shi, and L.~Carlone, ``Teaser: Fast and certifiable point cloud
  registration,'' \emph{IEEE Transactions on Robotics}, vol.~37, no.~2, pp.
  314--333, 2020.

\bibitem{yang2015go}
J.~Yang, H.~Li, D.~Campbell, and Y.~Jia, ``Go-icp: A globally optimal solution
  to 3d icp point-set registration,'' \emph{PAMI}, vol.~38, no.~11, pp.
  2241--2254, 2015.

\bibitem{yuan2020deepgmr}
W.~Yuan, B.~Eckart, K.~Kim, V.~Jampani, D.~Fox, and J.~Kautz, ``Deepgmr:
  Learning latent gaussian mixture models for registration,'' in \emph{ECCV},
  2020.

\bibitem{zeng20173dmatch}
A.~Zeng, S.~Song, M.~Nie{\ss}ner, M.~Fisher, J.~Xiao, and T.~Funkhouser,
  ``3dmatch: Learning local geometric descriptors from rgb-d reconstructions,''
  in \emph{CVPR}, 2017.

\bibitem{zhang2021reference}
Z.~Zhang, T.~Sattler, and D.~Scaramuzza, ``Reference pose generation for
  long-term visual localization via learned features and view synthesis,''
  \emph{IJCV}, vol. 129, no.~4, pp. 821--844, 2021.

\bibitem{zhong2015adaptive}
H.~Zhong, N.~Wen, J.~J. Gordon, M.~A. Elshaikh, B.~Movsas, and I.~J. Chetty,
  ``An adaptive mr-ct registration method for mri-guided prostate cancer
  radiotherapy,'' \emph{Physics in Medicine \& Biology}, vol.~60, no.~7, p.
  2837, 2015.

\bibitem{zhou2016fast}
Q.-Y. Zhou, J.~Park, and V.~Koltun, ``Fast global registration,'' in
  \emph{European conference on computer vision}.\hskip 1em plus 0.5em minus
  0.4em\relax Springer, 2016, pp. 766--782.

\bibitem{Zhou2018}
------, ``{Open3D}: {A} modern library for {3D} data processing,''
  \emph{arXiv:1801.09847}, 2018.

\bibitem{zhu2022correspondence}
M.~Zhu, M.~Ghaffari, and H.~Peng, ``Correspondence-free point cloud
  registration with so (3)-equivariant implicit shape representations,'' in
  \emph{Conference on Robot Learning}.\hskip 1em plus 0.5em minus 0.4em\relax
  PMLR, 2022, pp. 1412--1422.

\end{thebibliography}
}


\begin{IEEEbiography}[{\includegraphics[width=1in,height=1.25in,clip,keepaspectratio]{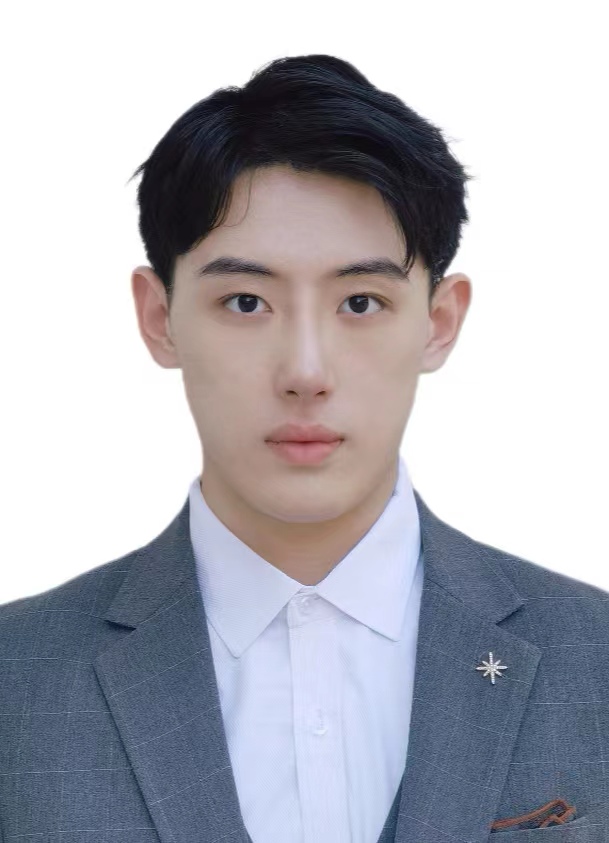}}]{Zexi Chen}
received his B.Eng. degree in Mechatronical Engineering from Zhejiang University in 2019. He is currently a Ph.D. candidate in the State Key Laboratory of Industrial Control and Technology, and Institute of Cyber-Systems and Control, Zhejiang University. His research interests include representation registration, 3D vision and scene understanding.
\end{IEEEbiography}

\begin{IEEEbiography}[{\includegraphics[width=1in,height=1.25in,clip,keepaspectratio]{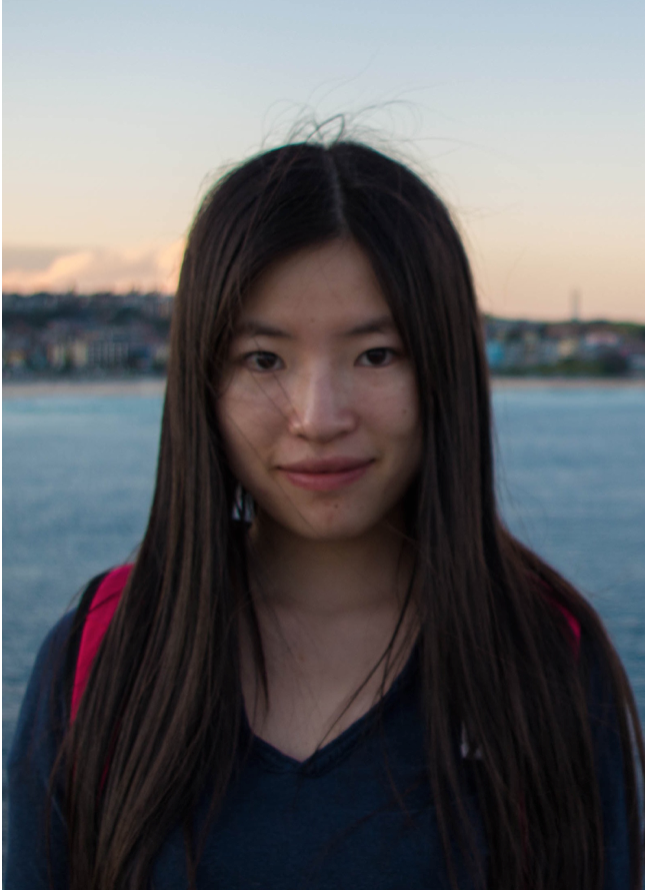}}]{Yiyi Liao}
received her Ph.D. degree from the Department of Control Science and Engineering, Zhejiang University, China in 2018. From 2018 to 2021, she was a postdoctoral researcher at the Autonomous Vision Group, University of Tübingen and Max Planck Institute for Intelligent Systems, Germany. She is currently an Assistant Professor at Zhejiang University. Her research interests include 3D vision and scene understanding.

\end{IEEEbiography}

\begin{IEEEbiography}[{\includegraphics[width=1in,height=1.25in,clip,keepaspectratio]{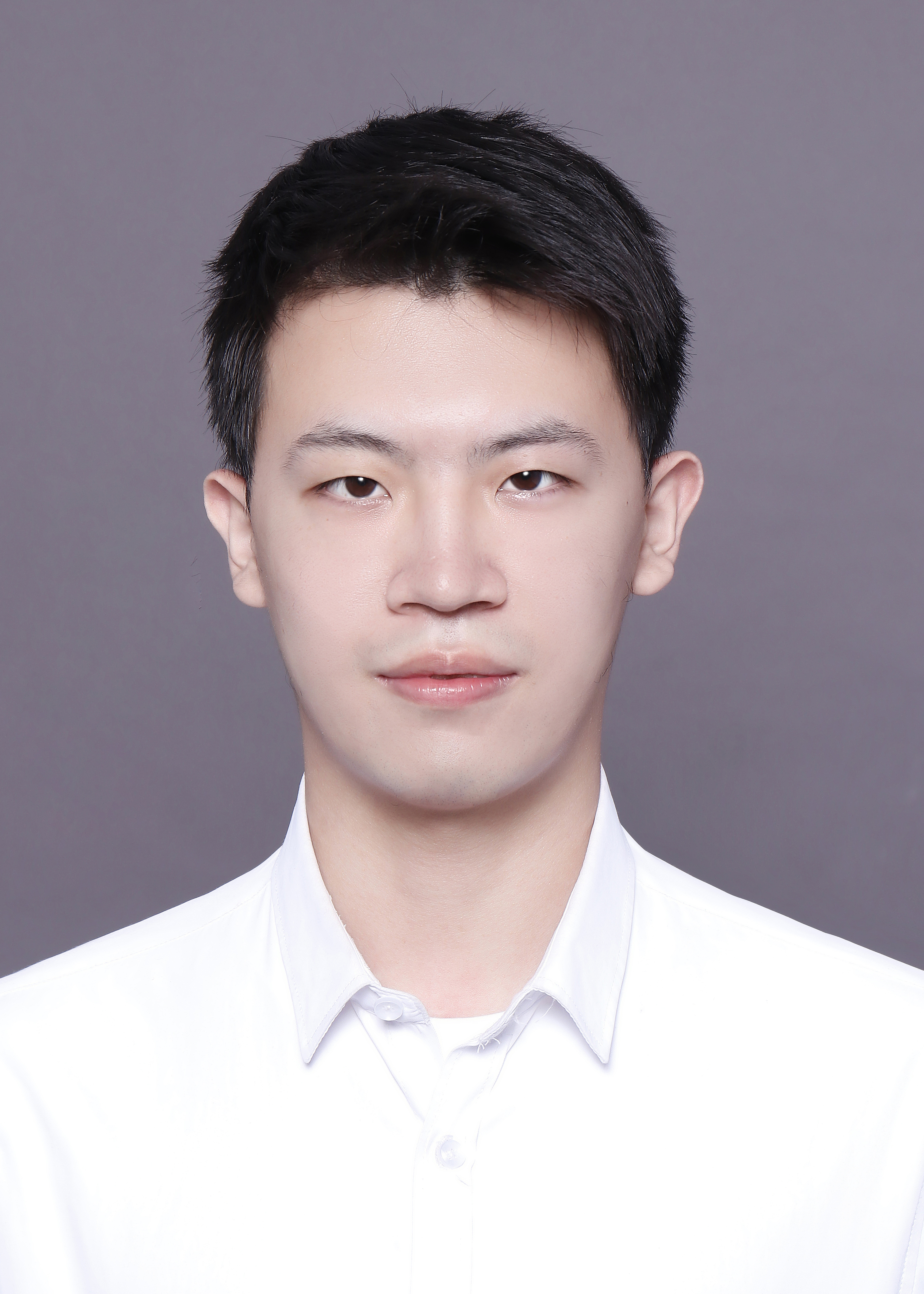}}]{Haozhe Du}
received his B.Eng. degree in Control Science and Engineering and a minor in Mechatronical Engineering from Zhejiang University in 2022, where he is currently pursuing the M.S. degree. His latest research interests include representation registration and 3D vision.
\end{IEEEbiography}

\begin{IEEEbiography}[{\includegraphics[width=1in,height=1.25in,clip,keepaspectratio]{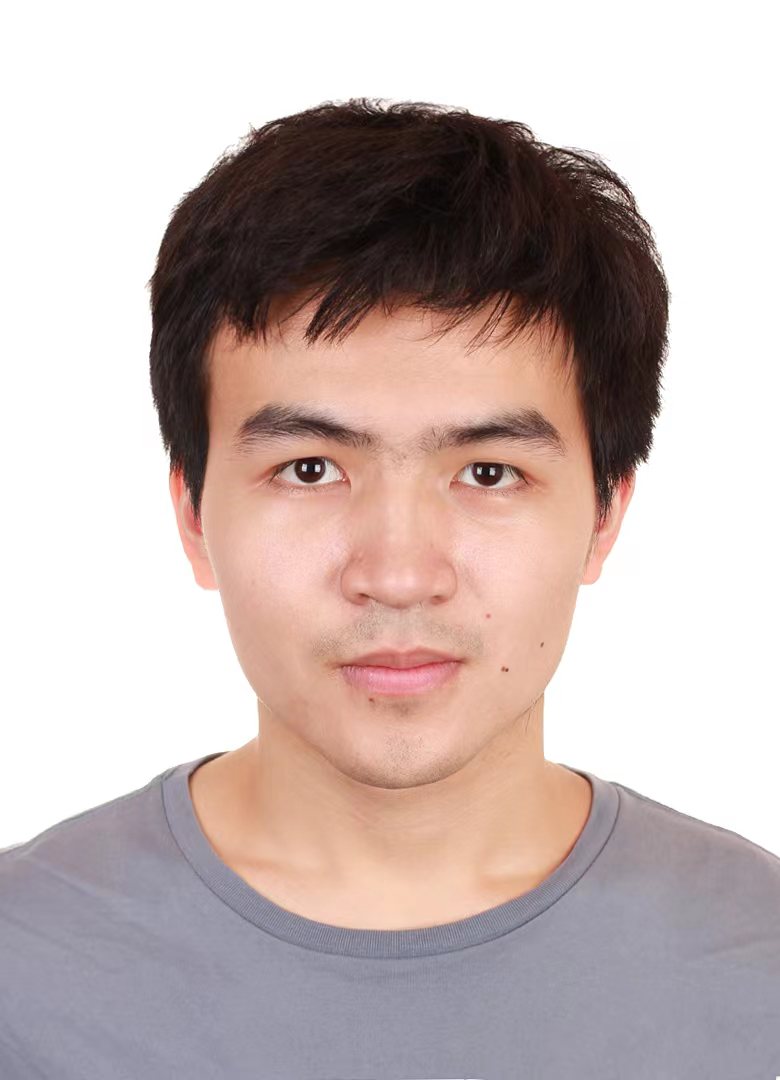}}]{Haodong Zhang}
Haodong Zhang received the B.S degree from the Department of Control Science and Engineering, Zhejiang University, Hangzhou, China, in 2020, where he is currently pursuing the M.S. degree. His latest research interests include imitation learning and transfer learning.
\end{IEEEbiography}

\begin{IEEEbiography}[{\includegraphics[width=1in,height=1.25in,clip,keepaspectratio]{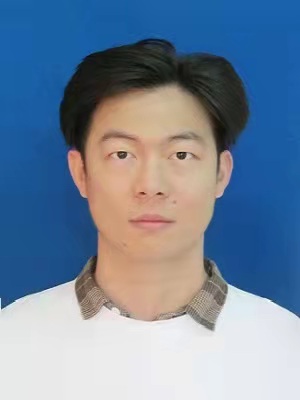}}]{Xuecheng Xu}
Xuecheng Xu received the B.S degree from the Department of Control Science and Engineering, Zhejiang University, Hangzhou, China, in 2019, where he is currently pursuing the PhD. degree. His latest research interests include SLAM and multi-robot systems.
\end{IEEEbiography}

\begin{IEEEbiography}[{\includegraphics[width=1in,height=1.25in,clip,keepaspectratio]{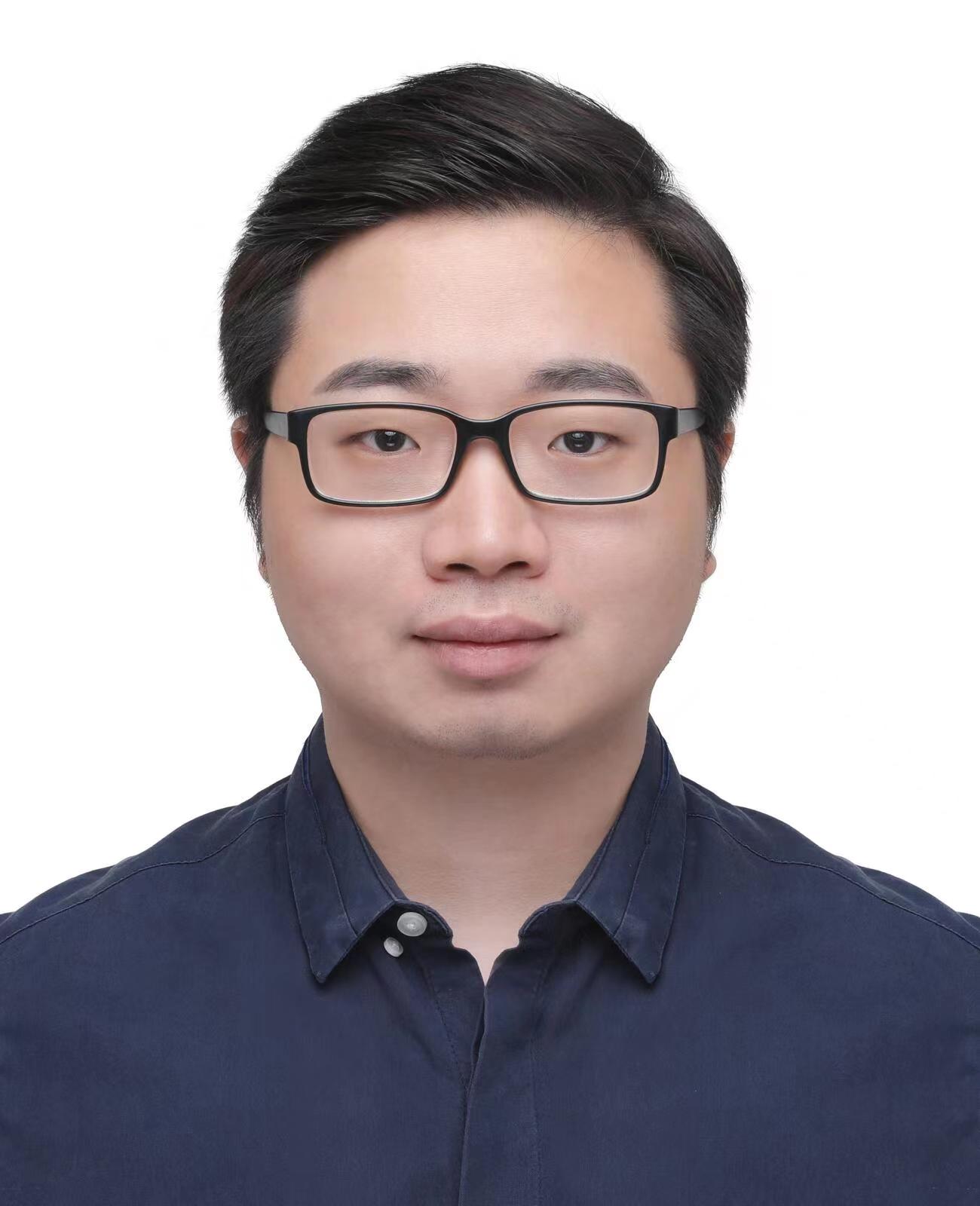}}]{Haojian Lu}
(Member, IEEE) received B.Eng. degree in Mechatronical Engineering from Beijing Institute of Technology in 2015, and he received Ph.D. degree in Robotics from City University of Hong Kong in 2019. He was a Research Assistant at City University of Hong Kong, from 2019 to 2020. He is currently a professor in the State Key Laboratory of Industrial Control and Technology, and Institute of Cyber-Systems and Control, Zhejiang University. His research interests include micro/nanorobotics, bioinspired robotics, medical robotics, micro aerial vehicle and soft robotics.
\end{IEEEbiography}

\begin{IEEEbiography}[{\includegraphics[width=1in,height=1.25in,clip,keepaspectratio]{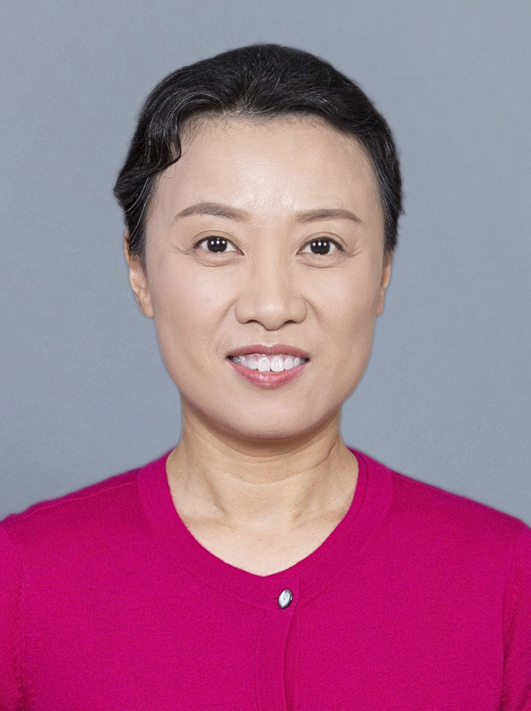}}]{Rong Xiong}
received her PhD in Control Science and Engineering from the Department of Control Science and Engineering, Zhejiang University, Hangzhou, P.R. China in 2009. She is currently a Professor in the Department of Control Science and Engineering, Zhejiang University, Hangzhou, P.R. China. Her latest research interests include motion planning and SLAM.
\end{IEEEbiography}

\begin{IEEEbiography}[{\includegraphics[width=1in,height=1.25in,clip,keepaspectratio]{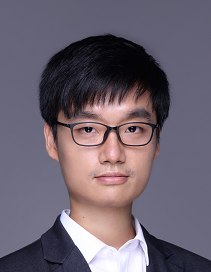}}]{Yue Wang}
received his Ph.D. degree in Control Science and Engineering from Department of Control Science and Engineering, Zhejiang University, Hangzhou, P.R. China in 2016. He is currently an Associate Professor in the Department of Control Science and Engineering, Zhejiang University, Hangzhou, P.R. China. His latest research interests include mobile robotics and robot perception.
\end{IEEEbiography}


%\newpage


%\vfill



\clearpage

\onecolumn

\appendices

\section{Experimental Setups}

\subsection{Point-Mesh-SDF Registration}
\label{append:3dhetero_setup}
\noindent\textsc{\textbf{Setup:}} To design such experiments, we choose the ``Linemod'' dataset which provides meshes of different objects. We generate point clouds and SDFs out of such meshes with the python library ``Open3D''\cite{Zhou2018}, and transform point cloud $v_{pc}$ or SDF $v_{sdf}$ with the randomely generated transformation $\boldsymbol{T}_{\boldsymbol{t},\boldsymbol{r},\mu}$. Note that we sample $6000$ points for each $v_{pc}$ from the mesh $v_{m}$, and the size of the $v_{sdf}$ is $2B\times2B\times2B$, which is dependent of the bandwidth $B$. We train the DPCN++ with $2000$ samples of $\{v_{pc}, v_{m}, v_{sdf}\}$ generated from $\{cat, holepuncher, can, bench, driller\}$ and test the performance with $500$ samples from $\{duck, lamp, cup, ape, phone, iron\}$ Note that for experiments that contain point clouds we also generate outliers in the same way as \ref{subsubsec:3D Homogeneous Registration}. To test the robustness of the framework, we also conduct experiments on three cases, i) complete to complete, ii) partial to complete, and iii) partial to partial. Unlike using the MVP dataset, we have to make our own partial-complete data. For point clouds and meshes, we obtain such partial data by directly cutting  points or meshes from a random region which makes up $20\%$ of the whole area of the object. For the SDF, we randomly select a region ($20\%$ of the whole volume of the input) and paint the pixels in it as $1e^{999}$. For comparison, we extract point clouds from the inputs of all the cases and test the corresponding performances of TEASER and DPCN++128 on such point clouds, denoted as TEASER* and DPCN++128*.

\subsection{Medical Images Registration}
\label{append:3dmedical_setup}
\noindent\textsc{\textbf{Setup:}} We conduct experiments between CT, MRI, and 3D Ultrasounds. For these medical images, the data comes with an perfect format of grids, but with different range of values for difference usages. Therefore, it is possible to align these 3D images, which will eventually help in practical applications. We choose ``RIRE'' which contains one labeled pair of CT and MRI of human head for CT-MRI rigid registration, and\cite{3dusctDataset} which contains several paires of labeled CT to Ultrasound images (one pair of canine spine, three pairs of human spine and one pair of lamp spine). Since there is not much of trainable data with ground truth of relative poses, we randomly crop every trainable samples from each datasets into $1000$ pieces for training. We also label some non-labeled data by experts with the software ``3D Slicer''~\cite{3dslicer} for testing. Unlike what we do in the point-mesh-sdf case, we can not extract point cloud from either CT, MRI, or 3D ultrasound since the points are of no sense, and therefore are not able to compare to TEASER* or DPCN++128*.

\section{Elaboration on Algorithm:}
\label{subsec: Elaboration on FFTs}

\subsection{Detailed Backward Propagation Formula}
\label{subsec:backward propogation}

Based on Eq.\ref{eq:chain_rule_gk} and following the notations in Eq.\ref{eq:Gg relation} and Eq.\ref{eq:phasecorr_G1G2_SO3_final}, we can further elaborate on the backward propagation formula of DPC due to the chain rule of partial derivative. 

\begin{equation}
\frac{\partial \mathcal{L}_r}{\partial g_1(\boldsymbol{k})}=
\frac{\partial \mathcal{L}_r}{\partial \hat{\boldsymbol{r}}}
\sum_{\boldsymbol{r}_{i}} \frac{\partial \hat{\boldsymbol{r}}}{\partial f(\boldsymbol{r}_{i})}
\frac{\partial f(\boldsymbol{r}_{i})}{\partial F}
\frac{\partial F}{\partial \check{S}_{1}}
\sum_{\boldsymbol{\check{\lambda}}}\sum_{\boldsymbol{j}}
\frac{\partial \check{S}_{1}}{\partial \check{s}_{1}(\boldsymbol{\check{\lambda}})}   
\frac{\partial \check{s}_{1}(\boldsymbol{\check{\lambda}})}{\partial \hat{G}_{1}(\boldsymbol{j})}
\frac{\partial \hat{G}_{1}(\boldsymbol{j})}{\partial g_1(\boldsymbol{k})}  
\label{eq:app_chain_rule_detailed}
\end{equation}

From Eq.\ref{eq:Gg relation}, and Eq.\ref{eq:phasecorr_G1G2_SO3_final}, the forward process mainly consists of Fourier transform, spherical Fourier transform, Hadamard product and inverse SO(3) Fourier transform. Since the Hadamard product is quite simple, we omit this derivation and display the detailed backward propagation derivation of the three Fourier transforms in the following sections. 

\subsubsection{Fourier Transform}

From Eq.\ref{eq:Gg relation} we can know that:

\begin{equation}
    G_{1}(\boldsymbol{j}) = \frac{1}{(2B)^{3}}\sum_{\boldsymbol{k}}g(\boldsymbol{k})e^{-i2\pi (\boldsymbol{j}^{T}\boldsymbol{k})},
    \label{eq:append_Gj}
\end{equation}

where $\boldsymbol{k}=[x\;y\;z]^T$ and this formula shows the form of 3D discrete Fourier transform.  $\hat{G}_{1}(\boldsymbol{j})$ is the magnitude of the frequency spectrum $G_{1}(\boldsymbol{j})$. Based on the definitions, we have:

\begin{equation}
    \hat{G}_{1}(\boldsymbol{j}) = G_{1}(\boldsymbol{j}) \bar{G}_{1}(\boldsymbol{j})
\end{equation}

Then following the chain rule, the derivative of the 3D discrete Fourier transform can be calculated below:

\begin{equation}
    \frac{\partial \hat{G}_{1}(\boldsymbol{j})}{\partial g_1(\boldsymbol{k})} = 
    \frac{\partial \hat{G}_{1}(\boldsymbol{j})}{\partial G_{1}(\boldsymbol{j})}
    \frac{\partial G_{1}(\boldsymbol{j})}{\partial g_1(\boldsymbol{k})} + 
    \frac{\partial \hat{G}_{1}(\boldsymbol{j})}{\partial \bar{G}_{1}(\boldsymbol{j})}
    \frac{\partial \bar{G}_{1}(\boldsymbol{j})}{\partial g_1(\boldsymbol{k})}
\end{equation}

Therefore taking $ \frac{\partial G_{1}(\boldsymbol{j})}{\partial g_1(\boldsymbol{k})}$ as an example, it can be derived as follows:

\begin{equation}
    \frac{\partial G_{1}(\boldsymbol{j})}{\partial g_1(\boldsymbol{k})} = 
    e^{-i2\pi (\boldsymbol{j}^{T}\boldsymbol{k})}
\end{equation}

\subsubsection{Spherical Fourier Transform}

Here we derive the partial derivative of spherical Fourier transform. For each sampled grid described as $\check{s}(\boldsymbol{\check{\lambda}})$ in spherical coordinates, the partial derivative can be represented as:

% From Eq.\ref{eq:define_spherical_value} we know that: $\boldsymbol{\lambda} = [\arctan(\frac{j_{2}}{j_{1}}), \arccos(\frac{j_{3}}{\sqrt{j_{1}^{2}+j_{2}^{2}+j_{3}^{2}}})]$. 

\begin{equation}
    \frac{\partial \check{S}_{1}}{\partial \check{s}_{1}(\boldsymbol{\check{\lambda}})} = \begin{bmatrix}
    \frac{\partial \check{S}_{11}}{\partial \check{s}_{1}(\boldsymbol{\check{\lambda}})} & \frac{\partial \check{S}_{12}}{\partial \check{s}_{1}(\boldsymbol{\check{\lambda}})} & \cdots & \frac{\partial \check{S}_{1B^2}}{\partial \check{s}_{1}(\boldsymbol{\check{\lambda}})}
    \end{bmatrix}^T
\end{equation}

where $\check{S}_{1p} (p=1,2,..,B^2)$ is for the $p$ th element of $\check{S_{1}}$. According to \cite{healy2003ffts}, the spherical Fourier transform can be represented more specifically in our case:

\begin{equation}
   \check{S}_{1p} =  \frac{\sqrt{2\pi}}{2B}\sum_{\boldsymbol{\check{\lambda}}}a_{u}\check{s}_{1}(\boldsymbol{\check{\lambda}})Y_p(\boldsymbol{\check{\lambda}}),
    \label{eq:append_SFT}
\end{equation}

where $a_u$ is a compensating weight and $Y_p$ is the spherical harmonics. More details about the spherical Fourier transform can be found in \cite{healy2003ffts}. We can calculate $\frac{\partial \check{S}_{1p}}{\partial \check{s}_{1}(\boldsymbol{\check{\lambda}})}$ based on the spherical Fourier transform formula:

\begin{equation}
    \frac{\partial \check{S}_{1p}}{\partial \check{s}_{1}(\boldsymbol{\check{\lambda}})} = \frac{\sqrt{2\pi}}{2B} a_u Y_p(\boldsymbol{\check{\lambda}}) \;\;\; (p=1,2,...,B^2)
\end{equation}

\subsubsection{Inverse SO(3) Fourier Transform}

According to \cite{cohen2018spherical}, the $SO(3)$ Fourier transformation $\mathfrak{F}_{SO(3)}$  as well as its inverse transformation $\mathfrak{i}\mathfrak{F}_{SO(3)}$, can ultimately be regarded as linear transformation. Therefore, from Eq.\ref{eq:phasecorr_G1G2_SO3_final} we know that the inverse $SO(3)$ Fourier transformation of $F$ is $f \in \mathbb{R}^{2B\times2B\times2B}$. One element of $f$ can be represented as follows:

\begin{equation}
    f(\boldsymbol{r}_{i}) = W_i F
\end{equation}

where $W_i$ represents the inverse $SO(3)$ Fourier transform with respect to $ f(\boldsymbol{r}_{i})$ and we use the symbolic expression to avoid tedious expressions and highlight the essence. More details are illustrated in \cite{cohen2018spherical}. Then we can derive the following partial derivative and $W_i^T$ can be regarded as $SO(3)$ Fourier transform, the inverse transform of inverse $SO(3)$ Fourier transform.

\begin{equation}
    \frac{\partial f(\boldsymbol{r}_{i})}{\partial F} = W_i^T
\end{equation}

\section{Additional Demonstration for 2D Registration}
\label{subsec: append_Additional Demonstration 2d}

\subsection{DPCN++ Visual Results on 2D Images}
\label{subsec:append_DPCN_2D_result}
\begin{figure*}[ht]
\centering
\includegraphics[width=\linewidth]{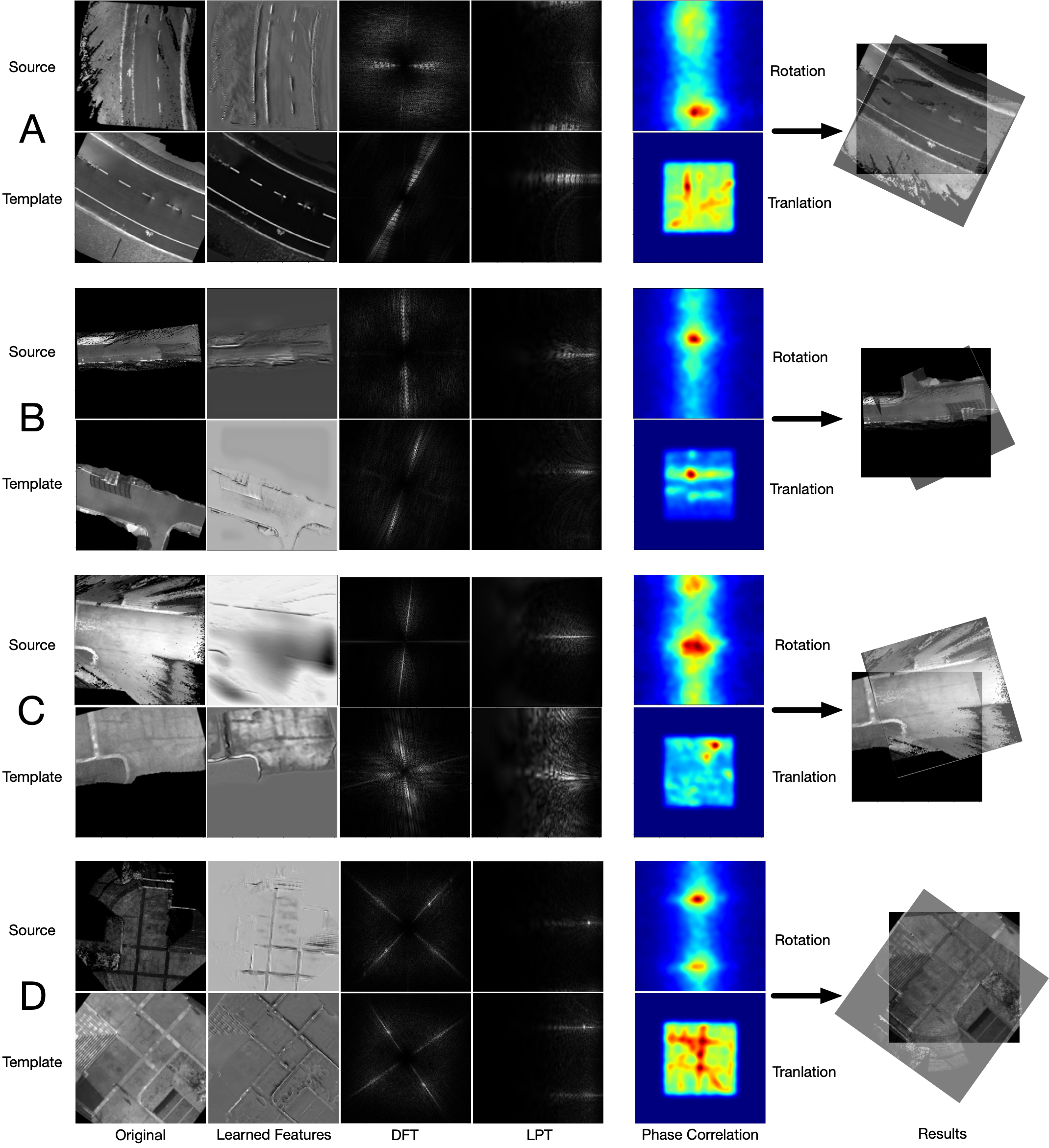}
\caption{Additional four demonstrations matching heterogeneous images pairs from Aero-Ground Dataset. The respective comparisons on classical phase correlation are shown in subsection \ref{subsec:append_PC}.}
\label{fig: additional demo}
\end{figure*}

\clearpage

\subsection{Conventional Phase Correlation Results on 2D Images}
\label{subsec:append_PC}
\begin{figure*}[ht]
\centering
\includegraphics[width=0.85\linewidth]{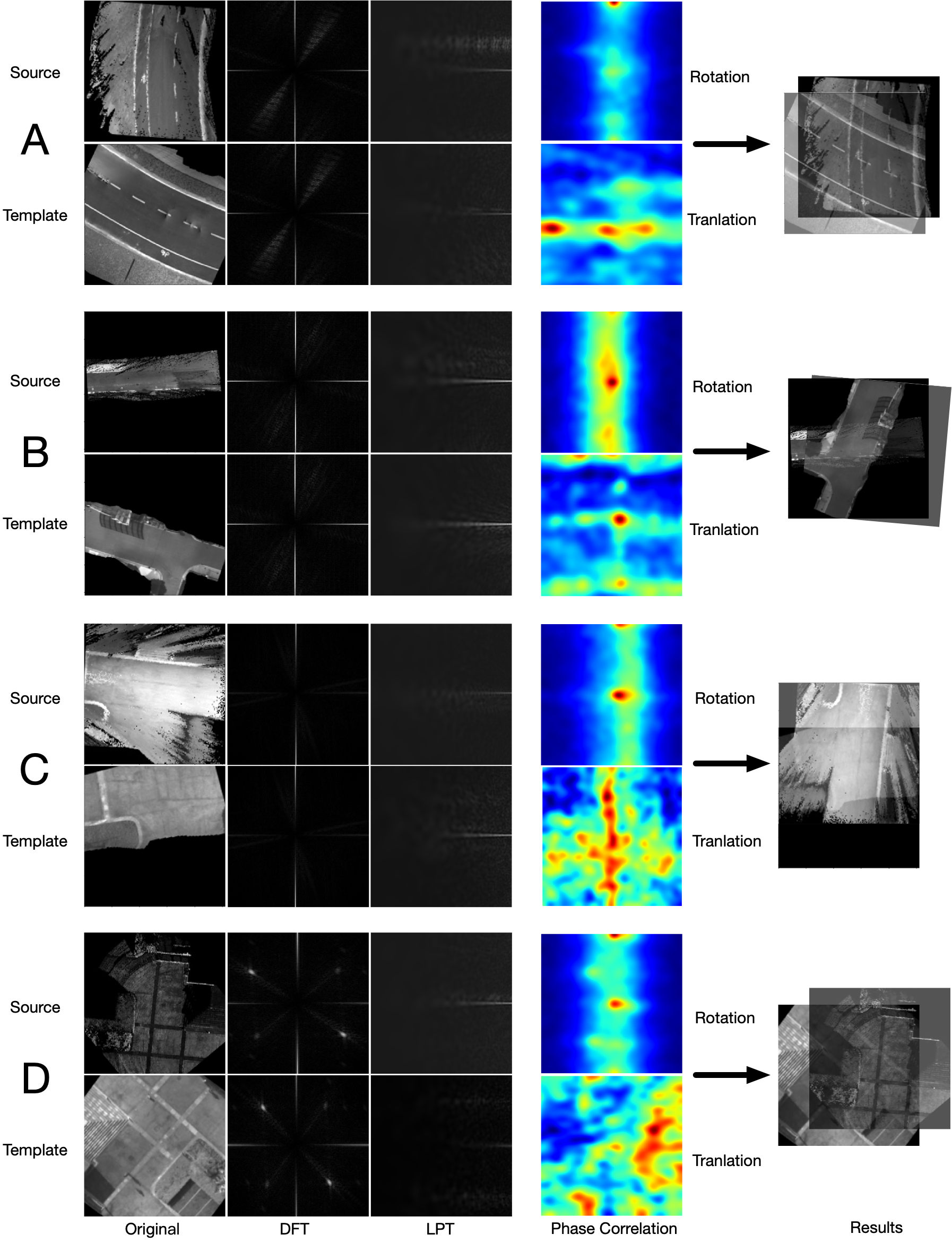}
\caption{Comparison using classical phase correlation to match heterogeneous images pairs from subsection \ref{subsec:append_DPCN_2D_result}.}
\label{fig: additional compare}
\end{figure*}

\clearpage

\subsection{Elaboration on 2D Registration: Simulation}
\label{subsec: append_Elaboration on 2D Sim}

In this subsection, we will give the results on 2D registration in the simulation dataset, shown in Tab.~\ref{tab: 2d_simulation}. In additional to the $Acc$, we also bring the Mean Square Error (MSE) to the evaluation:
\begin{gather*}
    E_{\boldsymbol{r}} = \frac{1}{n} \cdot \sum_{i=1}^{N}(\boldsymbol{r}^{*}_{i}-\hat{\boldsymbol{r}_{i}})^2,\\
    E_{\boldsymbol{t}} = \frac{1}{n} \cdot \sum_{i=1}^{N}(\boldsymbol{t}^{*}_{i}-\hat{\boldsymbol{t}_{i}})^2,\\
    E_{\mu} = \frac{1}{n} \cdot \sum_{i=1}^{N}(\mu^{*}_{i}-\hat{\mu_{i}})^2.
    \label{eq:append_mse}
\end{gather*}

\begin{table*}[h]
\renewcommand{\arraystretch}{1.2}
\captionsetup{justification=raggedright, singlelinecheck=true}
\caption{Results of simulation dataset. We choose the threshold error of $5\:pixels$ for translation, $1^{\circ}$ for rotation and $0.2\times$ for scale. \textbf{Note}:  $Exp.1$, $Exp.2$ and $Exp.3$ is conducted on ``Homogeneous", ``Heterogeneous w/ Outlier" and ``Heterogeneous'' sets respectively. \textbf{Bold} is the best performance and \underline{underline} is the secondary.}
\centering
\resizebox{0.9\textwidth}{!}{
\begin{tabular}{@{}cllllllllll@{}}
\toprule[1pt]
\multicolumn{1}{l}{\textbf{Baselines}} &
  Exp. &
  $E_x$ &
  $Acc_{x_5}(\%)$ &
  $E_y$ &
  $Acc_{y_5}(\%)$ &
  $E_r$ &
  $Acc_{r_1}(\%)$ &
  $E_{\mu}$ &
  $Acc_{\mu_{0.2}}(\%)$ &
  Runtime(ms) \\ \midrule
% \begin{tabular}[c]{@{}c@{}}
% \textsc{\underline{Image}}\\ 
% \textsc{\underline{Registration}}
% \end{tabular} 

PC & 1 & \underline{0.6635} & \underline{99.2} & \underline{0.923} & \underline{99.5} & 0.066 & 99.7 & 0.071 & 98.9 & 141.4 \\
 & 2 & 2319.553 & 49.2 & 2945.301 & 42.6 & 121.502 & 67.9 & \underline{0.121} & \underline{96.7} & 137.0 \\
 & 3   & 1774.159 & 69.1 & 3233.813 & 45.7 & 145.856 & 72.3 & 0.199 & \underline{97.6} & \underline{18.8}\\
\textsc{R2D2}        & 1 & 1.031 & 99.1 & 0.995 & 98.6 & \underline{0.055} & \underline{99.8} & \underline{0.069} & \underline{99.2} & 244.1 \\
                     & 2 & 458.837 & 32.1 & 328.077 & 28.6 & 291.781 & 41.4 & 0.221 & 71.3 & 243.9 \\
                     & 3   & 475.639 & 31.4 & 271.551 & 21.4 & 328.524 & 40.2 & 0.174 & 75.4 & 244.2 \\
\textsc{DAM}        & 1 & 53.759 & 90.6 & 28.682 & 95.9 & 19.224 & 81.7 & $\setminus$ & $\setminus$ & 111.7 \\
                     & 2 & 46.1165 & 71.3 & 89.6835 & 68.2 & \underline{36.5608} & \underline{77.8} & $\setminus$ & $\setminus$ & 110.4 \\
                      & 3   & \underline{2.581} & \underline{99.6} & \underline{4.623} & \underline{99.2} & \underline{28.934} & \underline{80.8} & $\setminus$ & $\setminus$ & 114.2 \\
\textsc{RPR}         & 1 & 8.675 & 96.9 & 2.220 & 97.3 & 16.272 & 90.2 & 0.080 & 95.2 & \textbf{6.1} \\
                     & 2 & 14.6842 & 51.1 & 11.1322 & 56.8 & 101.3329 & 76.8 & 0.1846 & 96.1 & \textbf{6.5} \\
                      & 3   & 22.363 & 62.1 & 32.833 & 49.1 & 97.851 & 78.3 & \underline{0.136} & 96.7 & \textbf{6.47}\\
\textsc{DS}          & 1 & 7.409 & 85.1 & 11.110 & 79.3 & 26.265 & 33.4 & $\setminus$ & $\setminus$ & 304.5 \\
                     & 2 & \underline{6.285} & \underline{83.9} & \underline{7.817} & \underline{79.6} & 25.664 & 27.1 & $\setminus$ & $\setminus$ & 301.4 \\
                     & 3   & 5.497 & 87.1 & 7.333 & 91.9 & 31.389 & 23.9 & $\setminus$ & $\setminus$ & 301.3\\
\textbf{\textsc{DPCN++}} & 1 & \textbf{0.103} & \textbf{100} & \textbf{0.216} & \textbf{100} & \textbf{0.052} & \textbf{100} & \textbf{0.052} & \textbf{100} & \underline{20.5}\\
  & 2 & \textbf{0.076} & \textbf{100} & \textbf{0.467} & \textbf{100} & \textbf{0.091} & \textbf{100} & \textbf{0.001} & \textbf{100} & \underline{22.3}\\
   & 3   & \textbf{0.007} & \textbf{100} & \textbf{0.017} & \textbf{100} & \textbf{0.039} & \textbf{100} & \textbf{0.064} & \textbf{100} & 22.1 \\
\bottomrule[1pt]
\end{tabular}
}
\label{tab: 2d_simulation}
\end{table*}

% \multirow{6}{*}{(Sim)}
%                      & PC~\cite{SrinivasaReddy1996}  & sim   & 1774.159 & 69.1 & 3233.813 & 45.7 & 145.856 & 72.3 & 0.199 & \underline{97.6} & \underline{18.8}\\
%                      & R2D2~\cite{revaud2019r2d2}                   & sim   & 475.639 & 31.4 & 271.551 & 21.4 & 328.524 & 40.2 & 0.174 & 75.4 & 244.2 \\
%                      & DS~\cite{Barnes2019}             & sim   & 5.497 & 87.1 & 7.333 & 91.9 & 31.389 & 23.9 & $\setminus$ & $\setminus$ & 301.3\\
%                      & DAM~\cite{Park2020}       & sim   & \underline{2.581} & \underline{99.6} & \underline{4.623} & \underline{99.2} & \underline{28.934} & \underline{80.8} & $\setminus$ & $\setminus$ & 114.2 \\
%                      & RPR~\cite{Kendall2017}             & sim   & 22.363 & 62.1 & 32.833 & 49.1 & 97.851 & 78.3 & \underline{0.136} & 96.7 & \textbf{6.47}\\

%                      & \textbf{DPCN++}             & sim   & \textbf{0.007} & \textbf{100} & \textbf{0.017} & \textbf{100} & \textbf{0.039} & \textbf{100} & \textbf{0.064} & \textbf{100} & 22.1 \\ \midrule

\begin{figure*}[h]
    \captionsetup[subfigure]{justification=centering}
    \centering
  \subfloat[Estimation of $x$ in Homogeneous dataset\label{Homo_x}]{%
      \includegraphics[width=0.25\linewidth]{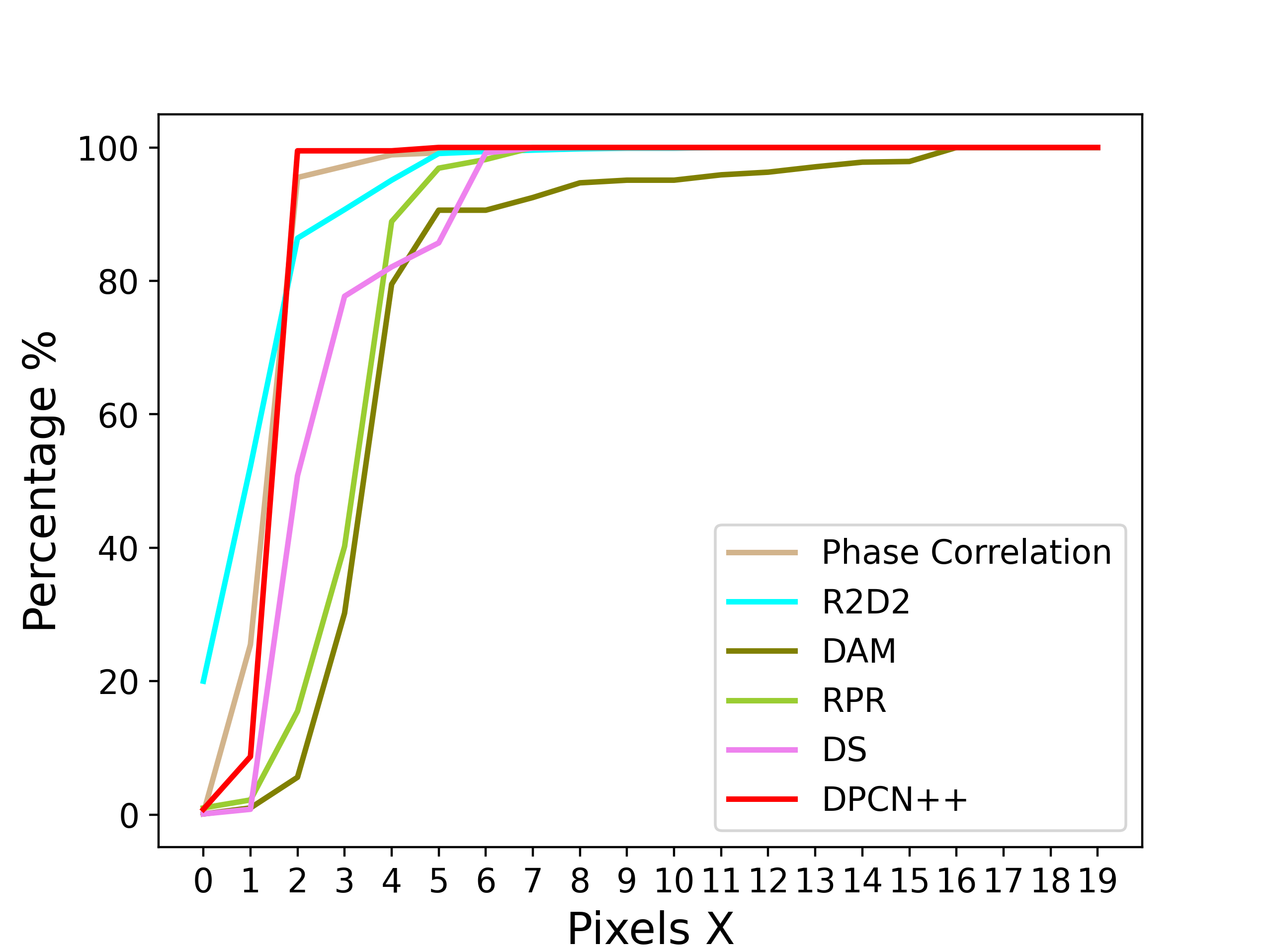}}
  \subfloat[Estimation of $y$ in Homogeneous dataset\label{Homo_y}]{%
        \includegraphics[width=0.25\linewidth]{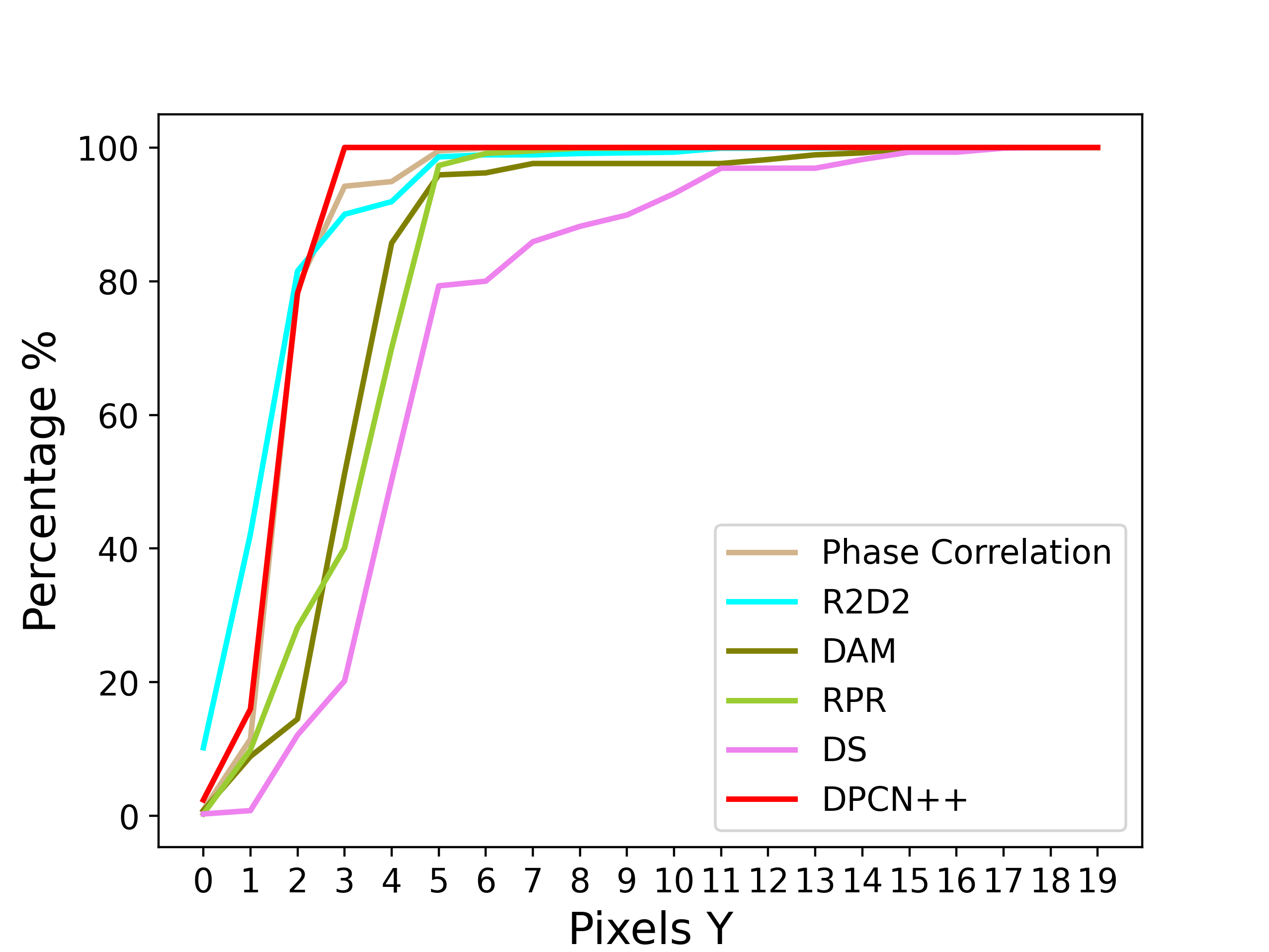}}
  \subfloat[Estimation of $x$ in Heterogeneous dataset\label{Hetero_x}]{%
        \includegraphics[width=0.25\linewidth]{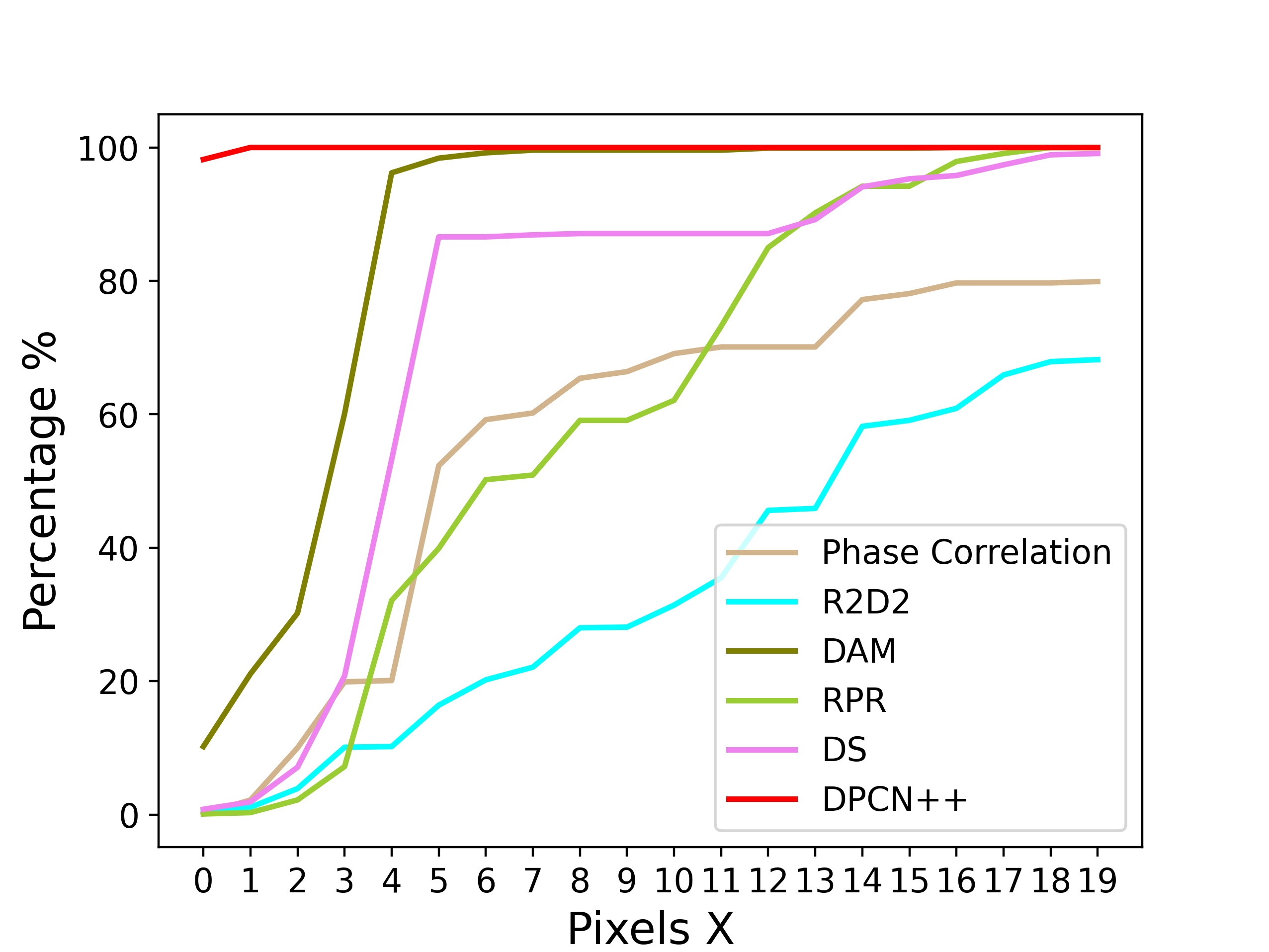}}
  \subfloat[Estimation of $y$ in Heterogeneous dataset\label{Hetero_y}]{%
        \includegraphics[width=0.25\linewidth]{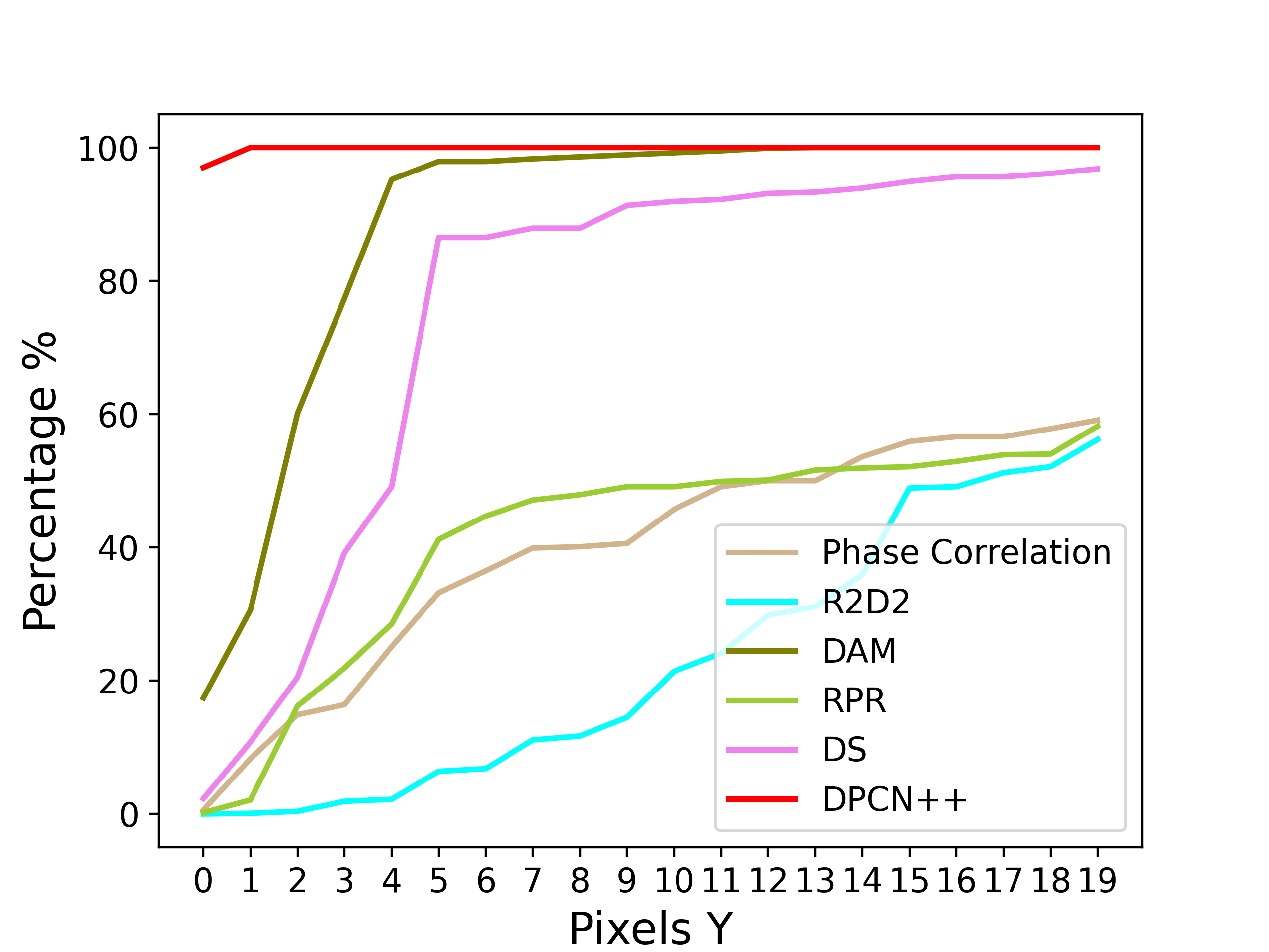}}
        \\
  \subfloat[Estimation of $x$ in Heterogeneous w/ Outlier dataset\label{Dynamic_x}]{%
        \includegraphics[width=0.25\linewidth]{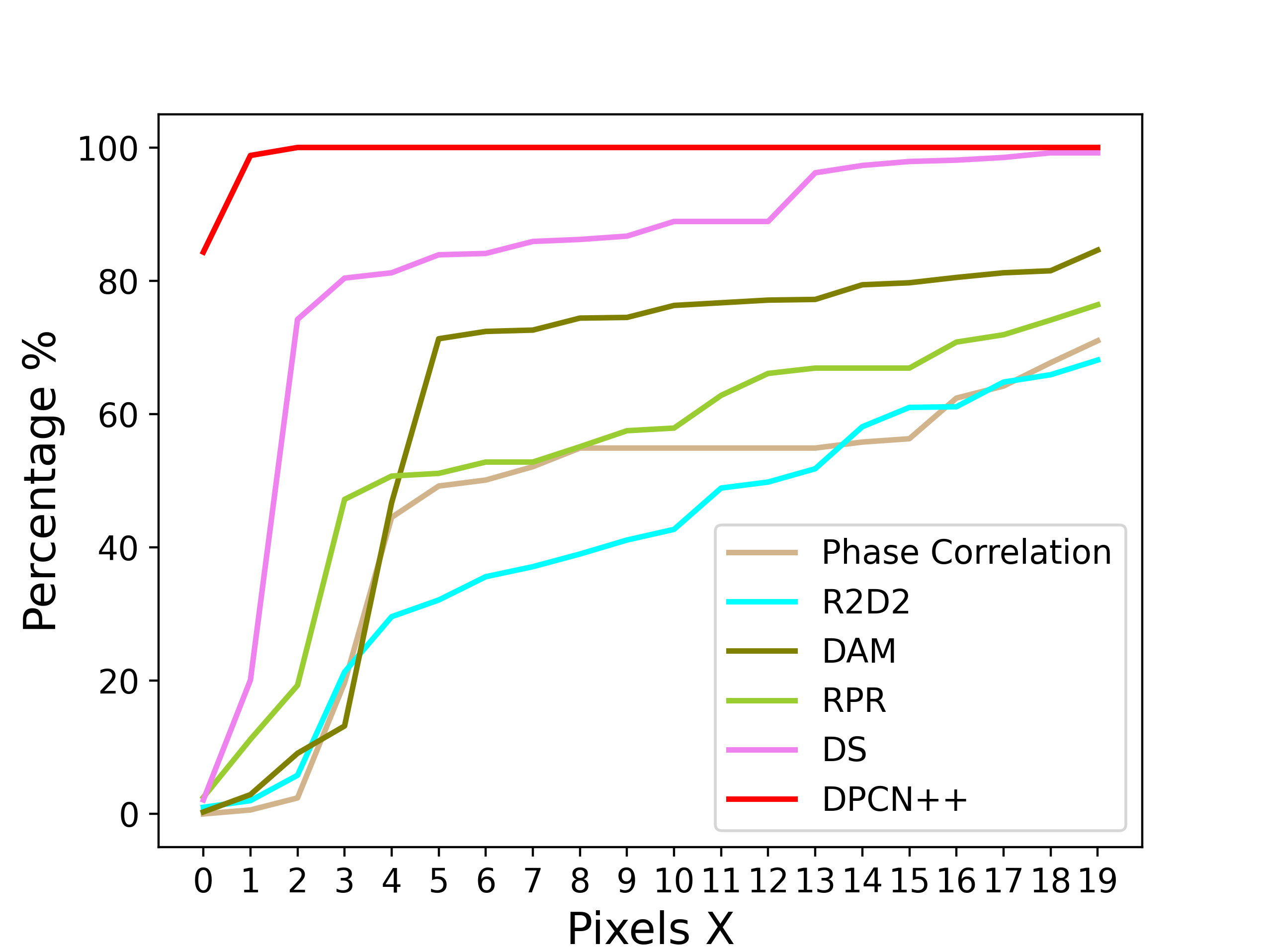}}
  \subfloat[Estimation of $y$ in Heterogeneous w/ Outlier dataset\label{Dynamic_y}]{%
        \includegraphics[width=0.25\linewidth]{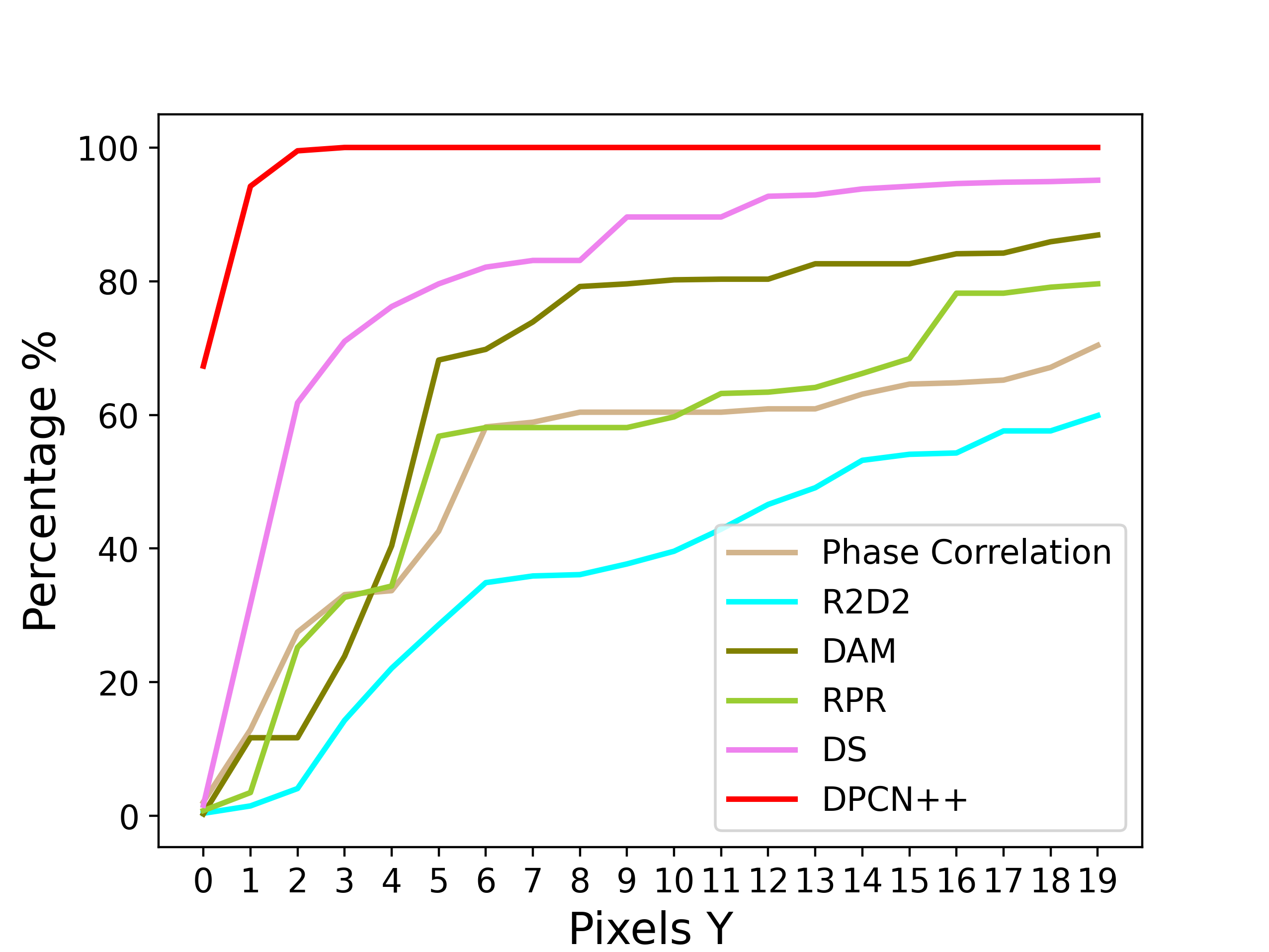}}
  \caption{$Acc_{0\:to\:19}$ of translation estimation in simulation dataset. }
  \label{Acc simulation}
\end{figure*}

Furthermore, the threshold of estimation in experiments is elaborated by the means of graphs. Figure \ref{Acc simulation} shows the $Acc_{0\:to\:19}$ of translation estimation in simulation dataset.

% The results show that for the homogeneous pairs of images, our method maintains an equivalent performance with the conventional phase correlation pipeline in accuracy with faster speed and outperformed the rest of the baselines.

\subsection{Elaboration on 2D Registration: AeroGround Dataset}
\label{subsec: append_Elaboration on 2D AG}

\begin{figure}[h]
    \captionsetup[subfigure]{justification=centering}
    \centering
  \subfloat[Estimation of $x$ in ``LiDAR to Drone" scene(a)\label{qsdjt_l2a_x}]{%
      \includegraphics[width=0.25\linewidth]{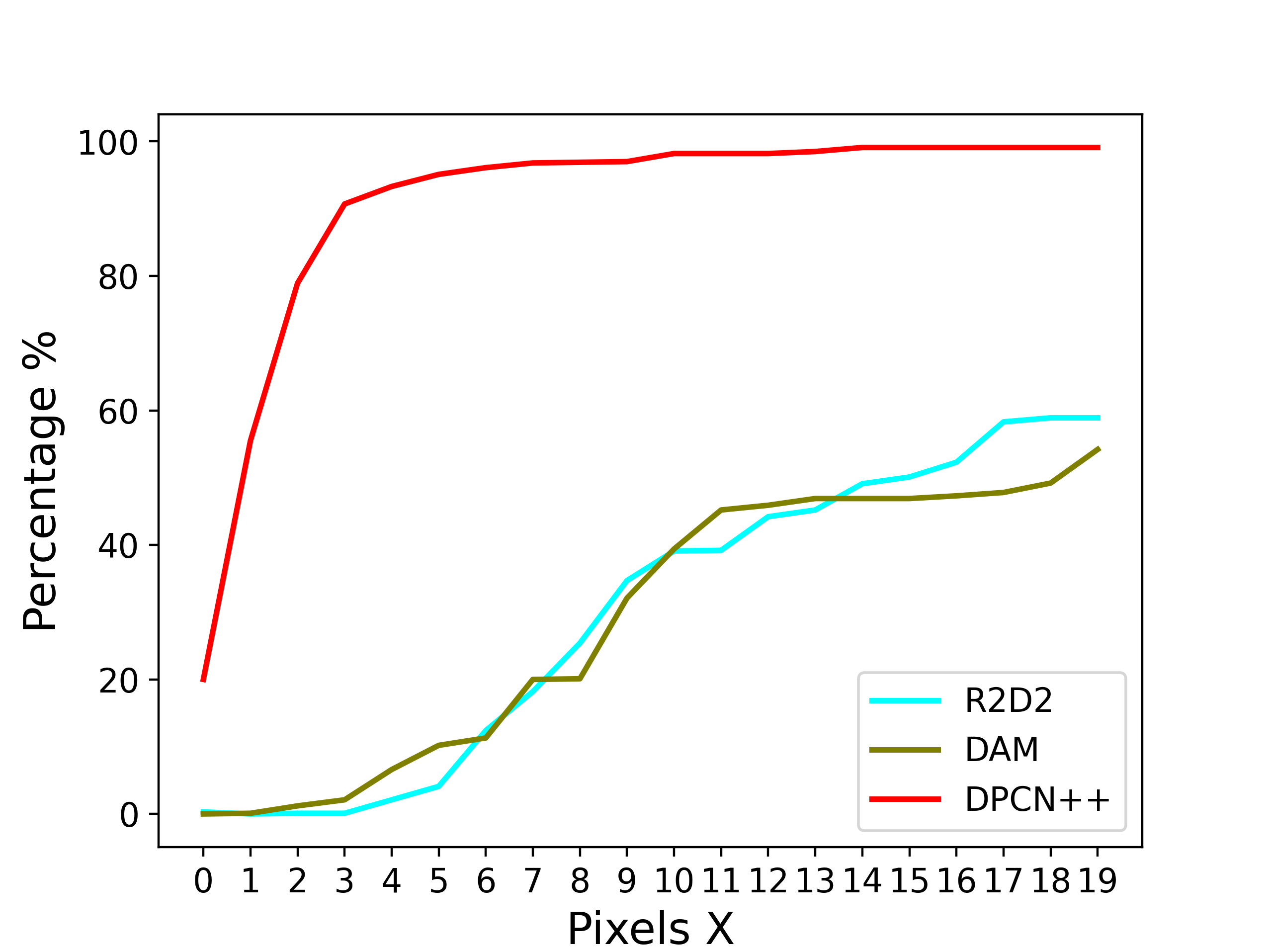}}
  \subfloat[Estimation of $y$ in ``LiDAR to Drone" scene(a)\label{qsdjt_l2a_y}]{%
        \includegraphics[width=0.25\linewidth]{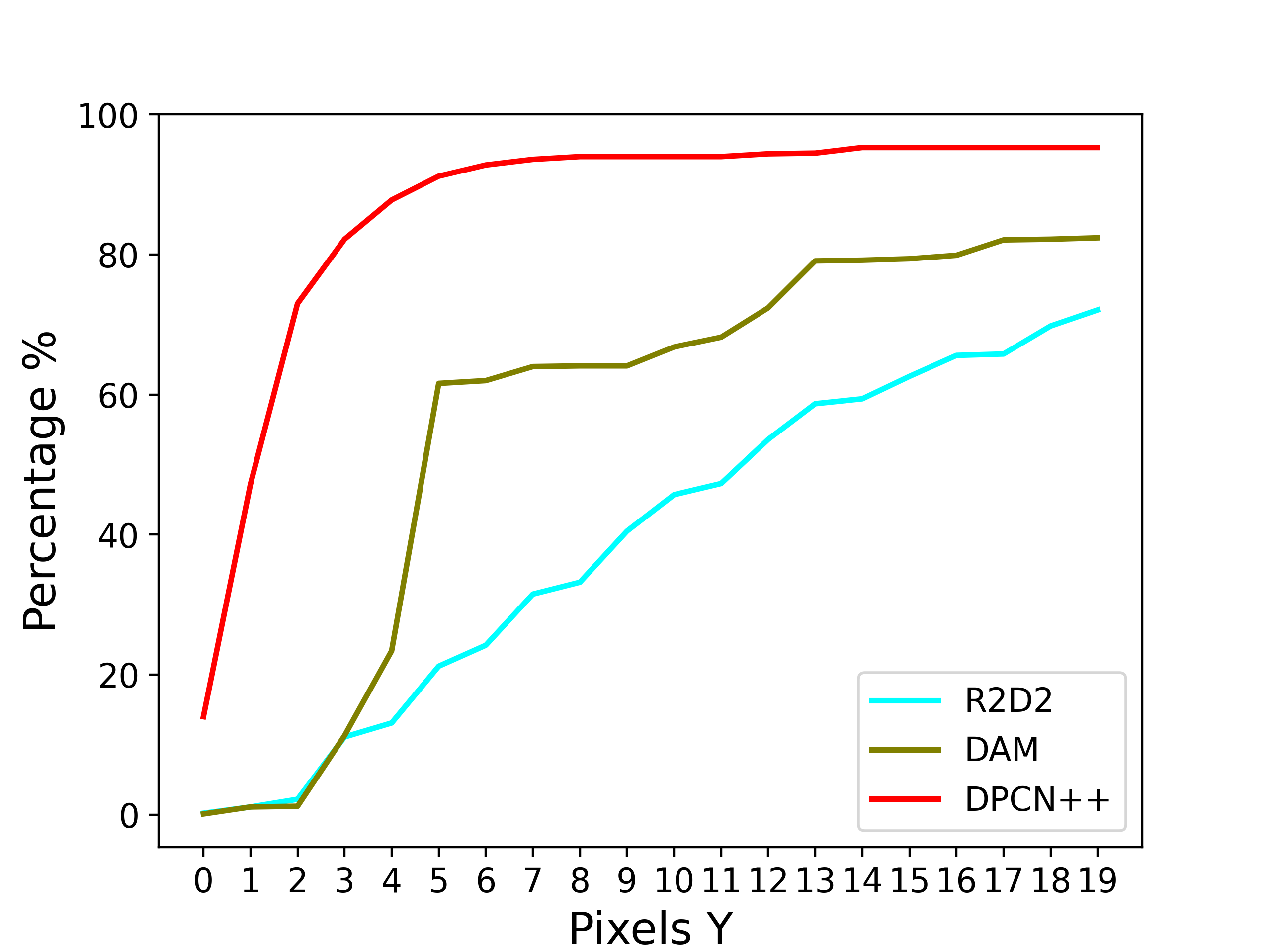}}
  \subfloat[Estimation of $x$ in ``LiDAR to Satellite" scene(a)\label{qsdjt_l2s_x}]{%
        \includegraphics[width=0.25\linewidth]{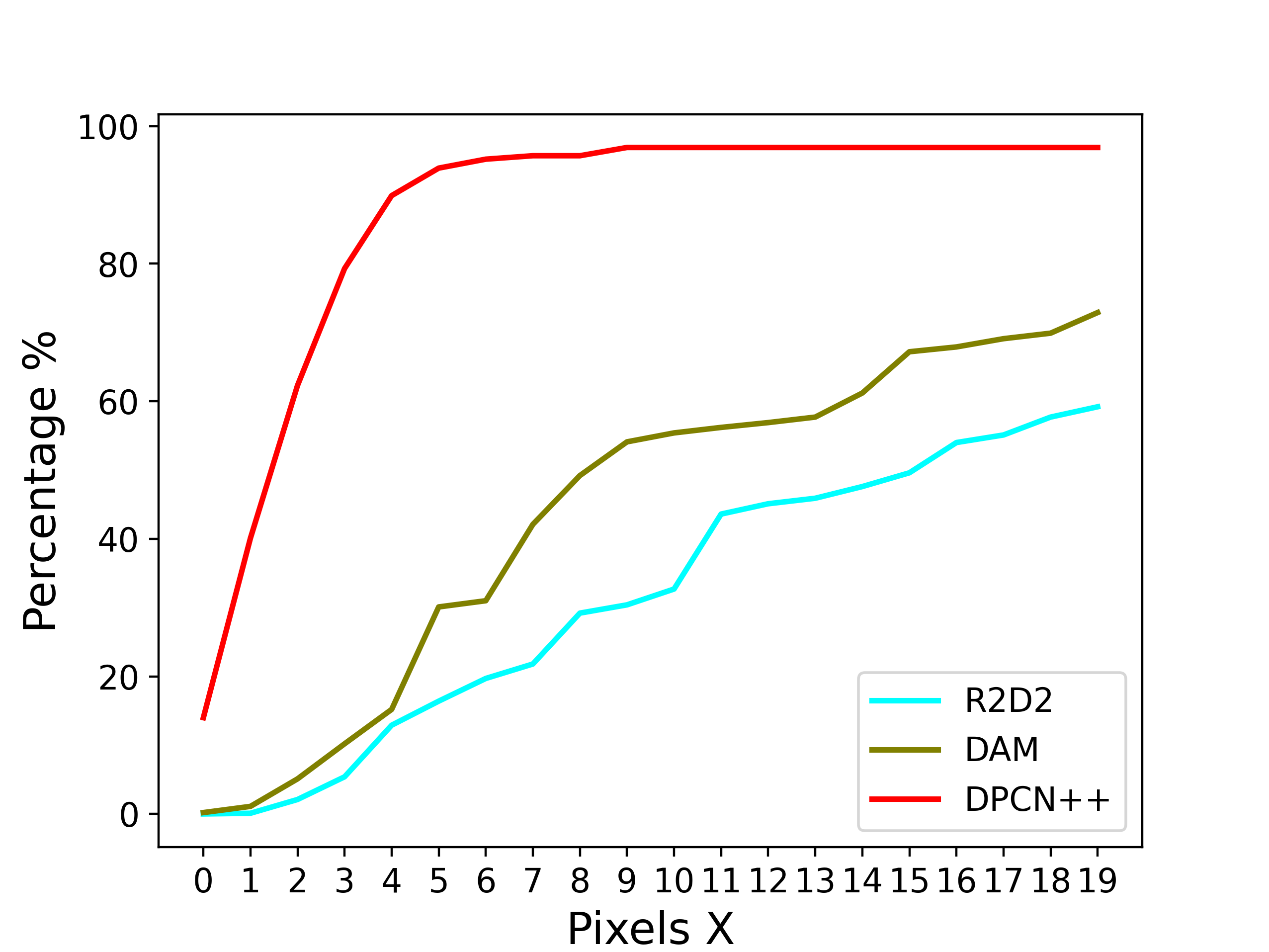}}
  \subfloat[Estimation of $y$ in ``LiDAR to Satellite" scene(a)\label{qsdjt_l2s_y}]{%
        \includegraphics[width=0.25\linewidth]{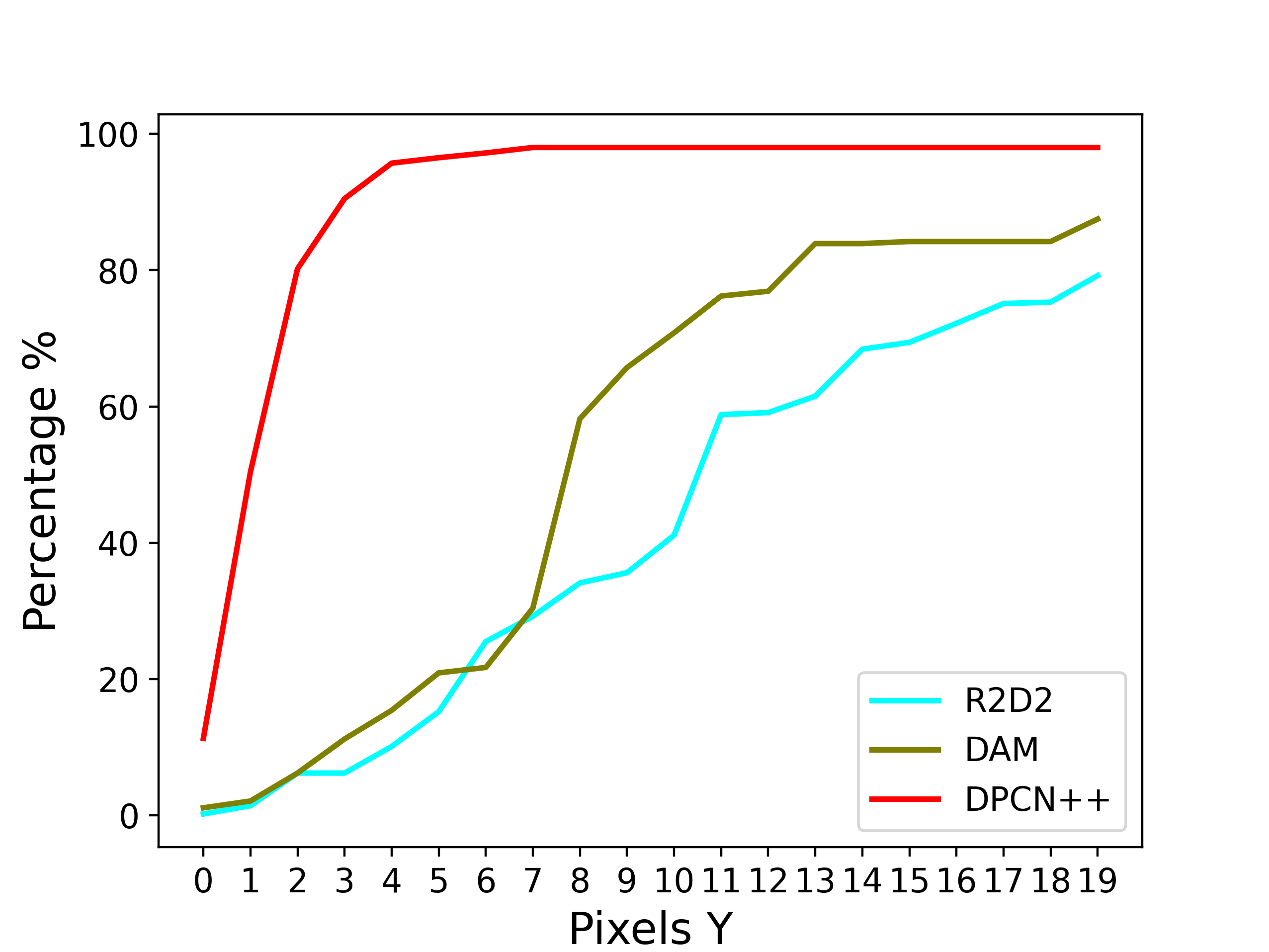}}
        \\
  \subfloat[Estimation of $x$ in ``Stereo to Drone" scene(a)\label{qsdjt_s2a_x}]{%
      \includegraphics[width=0.25\linewidth]{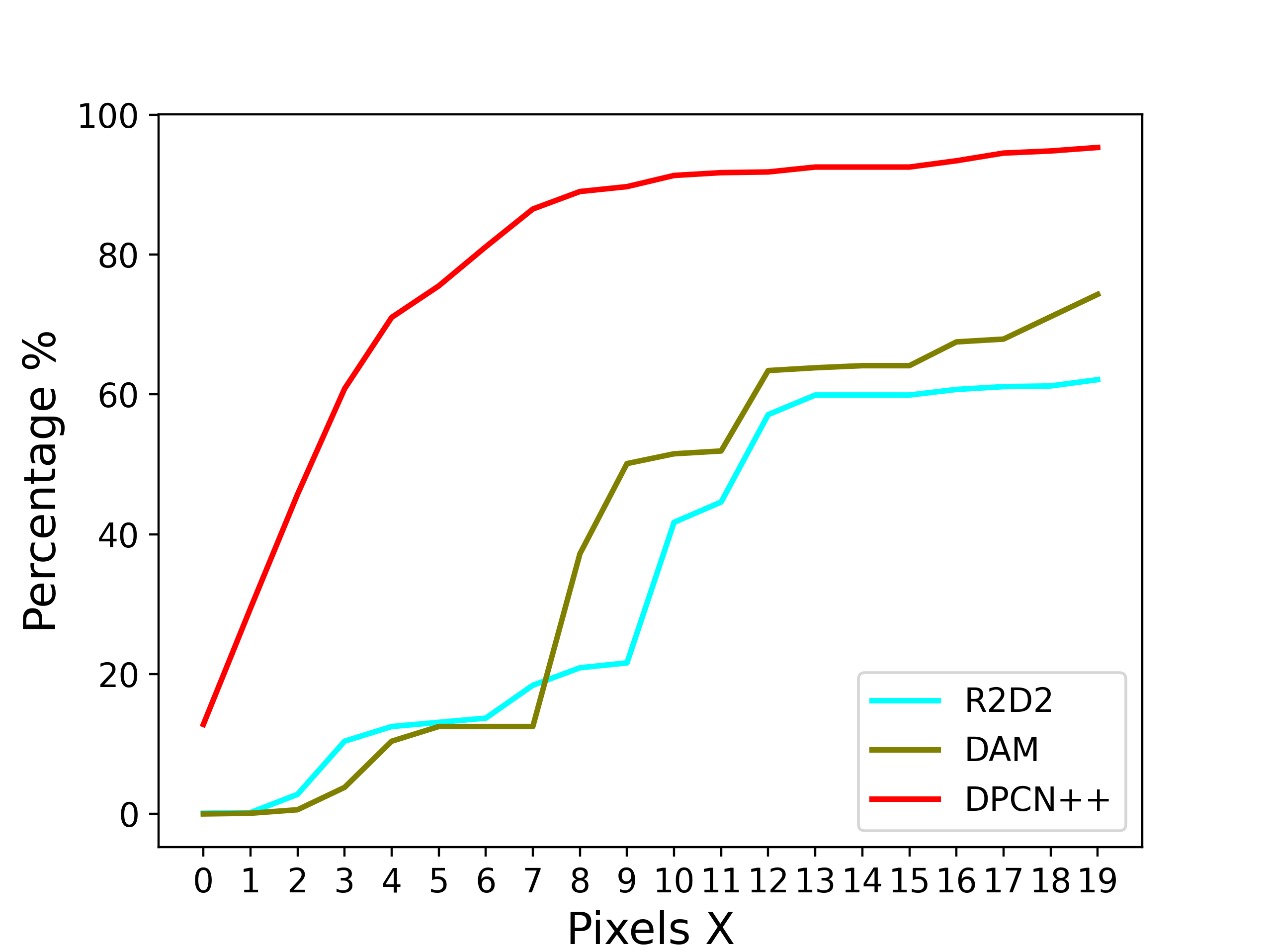}}
  \subfloat[Estimation of $y$ in ``Stereo to Drone" scene(a)\label{qsdjt_s2a_y}]{%
        \includegraphics[width=0.25\linewidth]{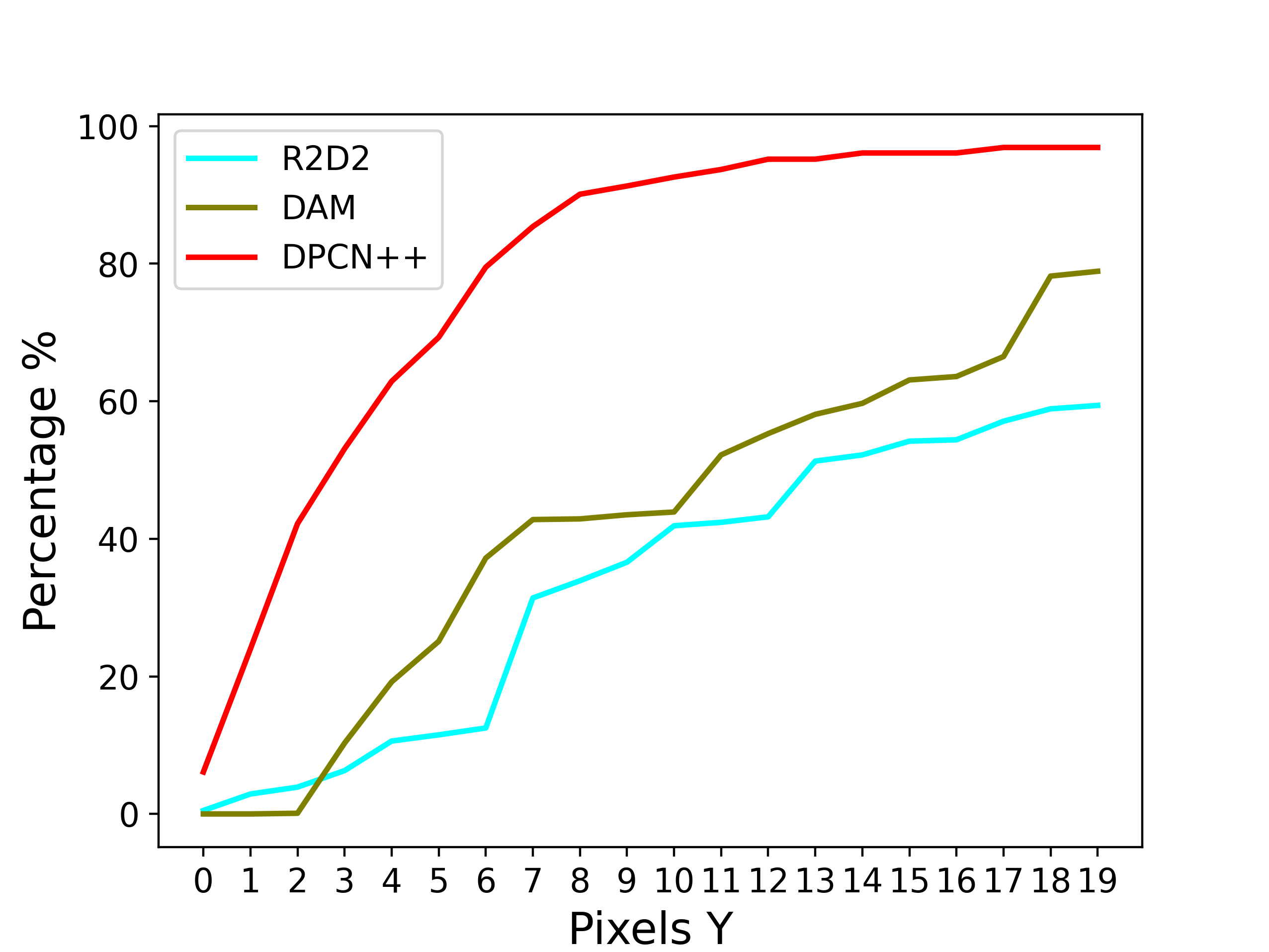}}
  \subfloat[Estimation of $x$ in ``Stereo to Satellite" scene(a)\label{qsdjt_s2s_x}]{%
        \includegraphics[width=0.25\linewidth]{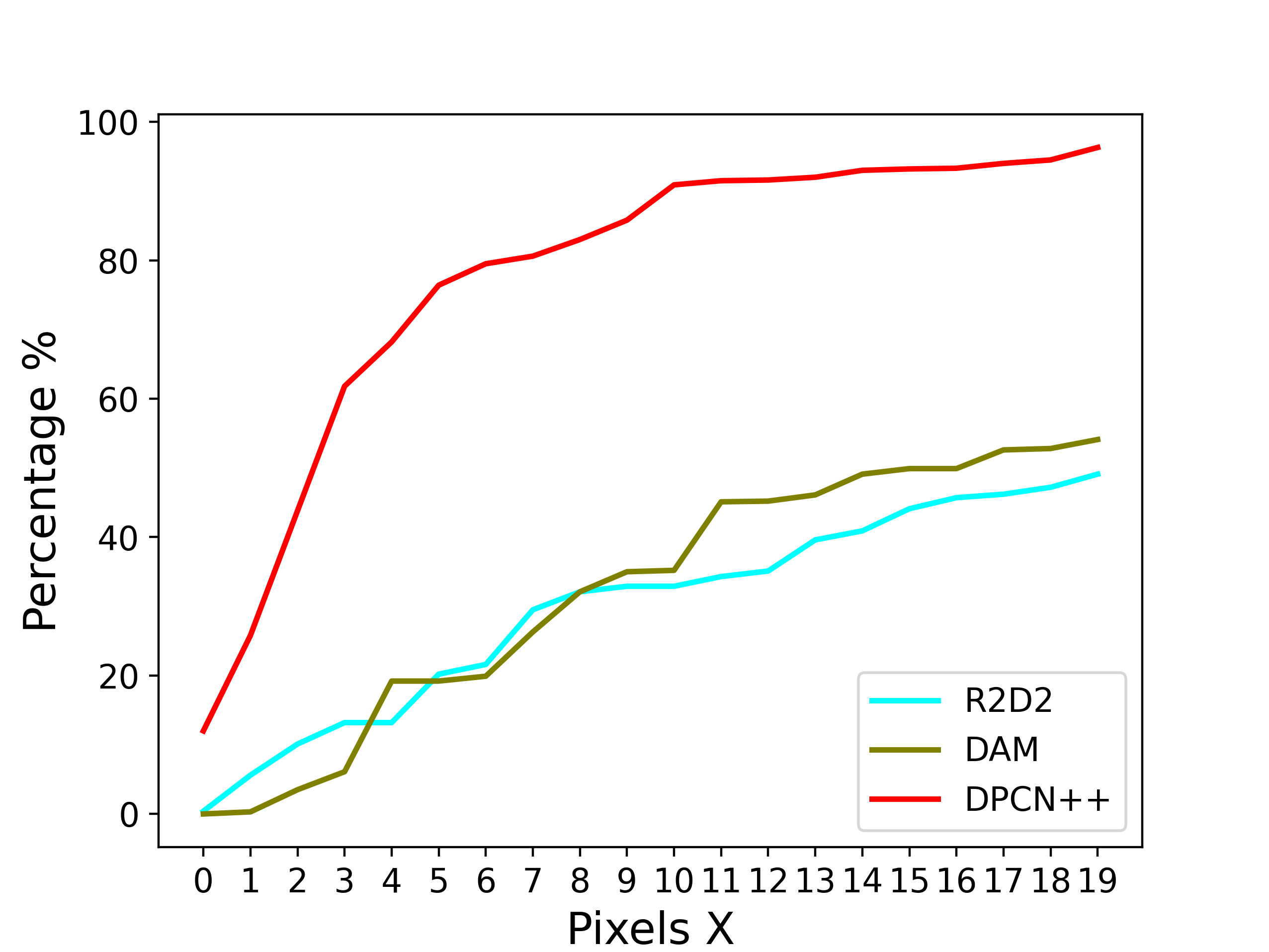}}
  \subfloat[Estimation of $y$ in ``Stereo to Satellite" scene(a)\label{qsdjt_s2s_y}]{%
        \includegraphics[width=0.25\linewidth]{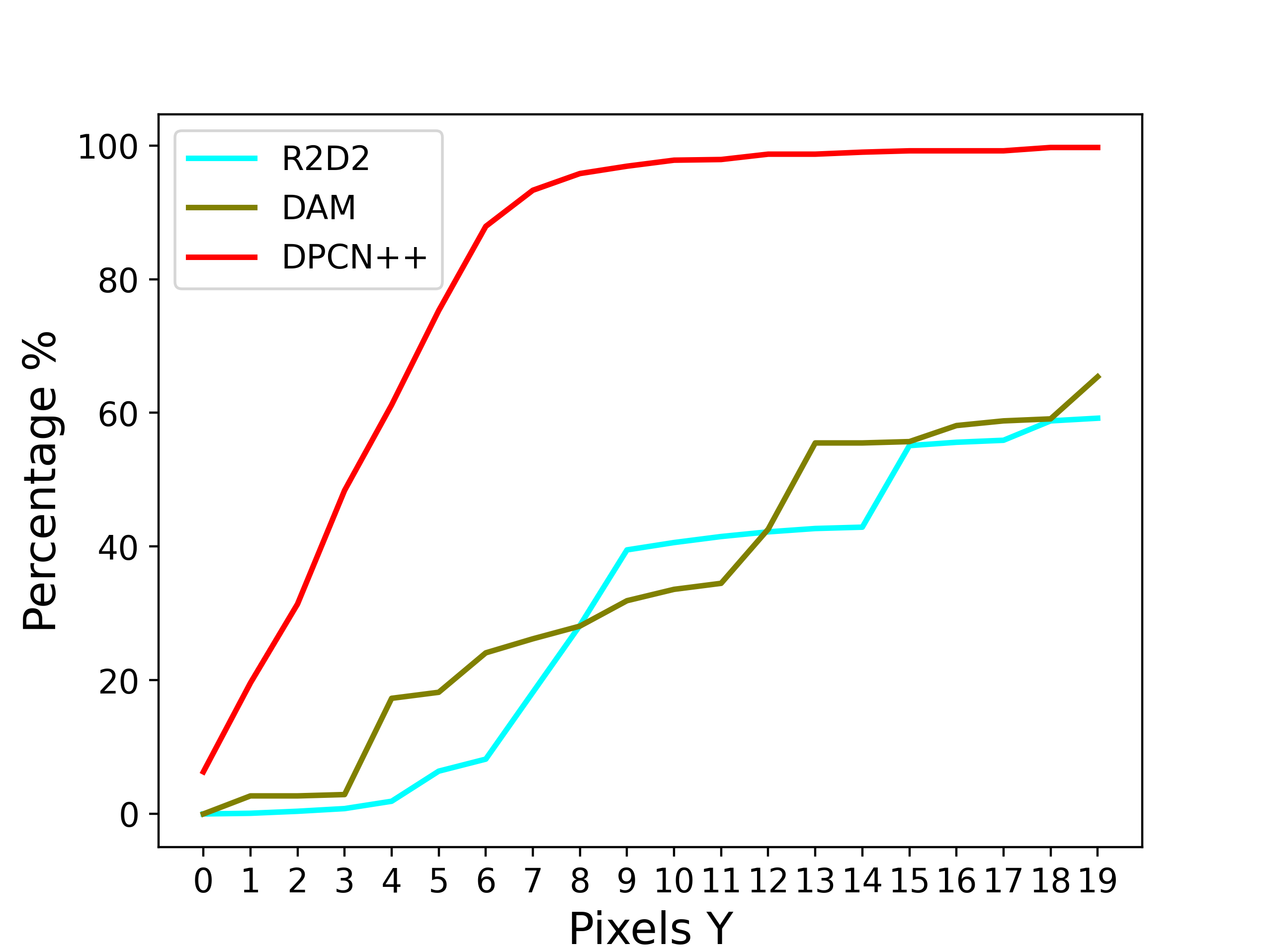}}
        \\
  \subfloat[Estimation of $x$ in ``LiDAR to Drone" scene(b)\label{gym_l2a_x}]{%
        \includegraphics[width=0.25\linewidth]{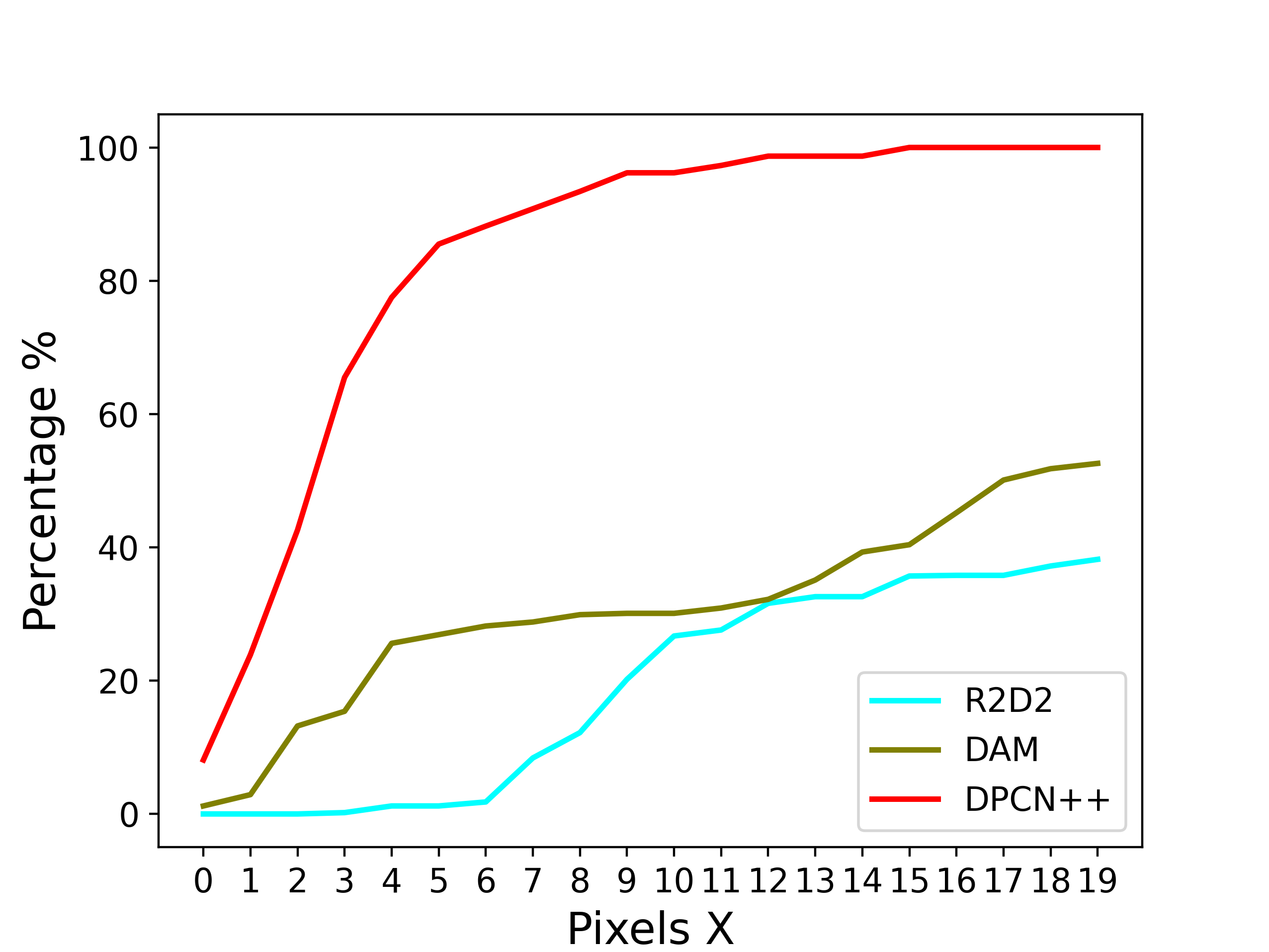}}
  \subfloat[Estimation of $y$ in ``LiDAR to Drone" scene(b)\label{gym_l2a_y}]{%
        \includegraphics[width=0.25\linewidth]{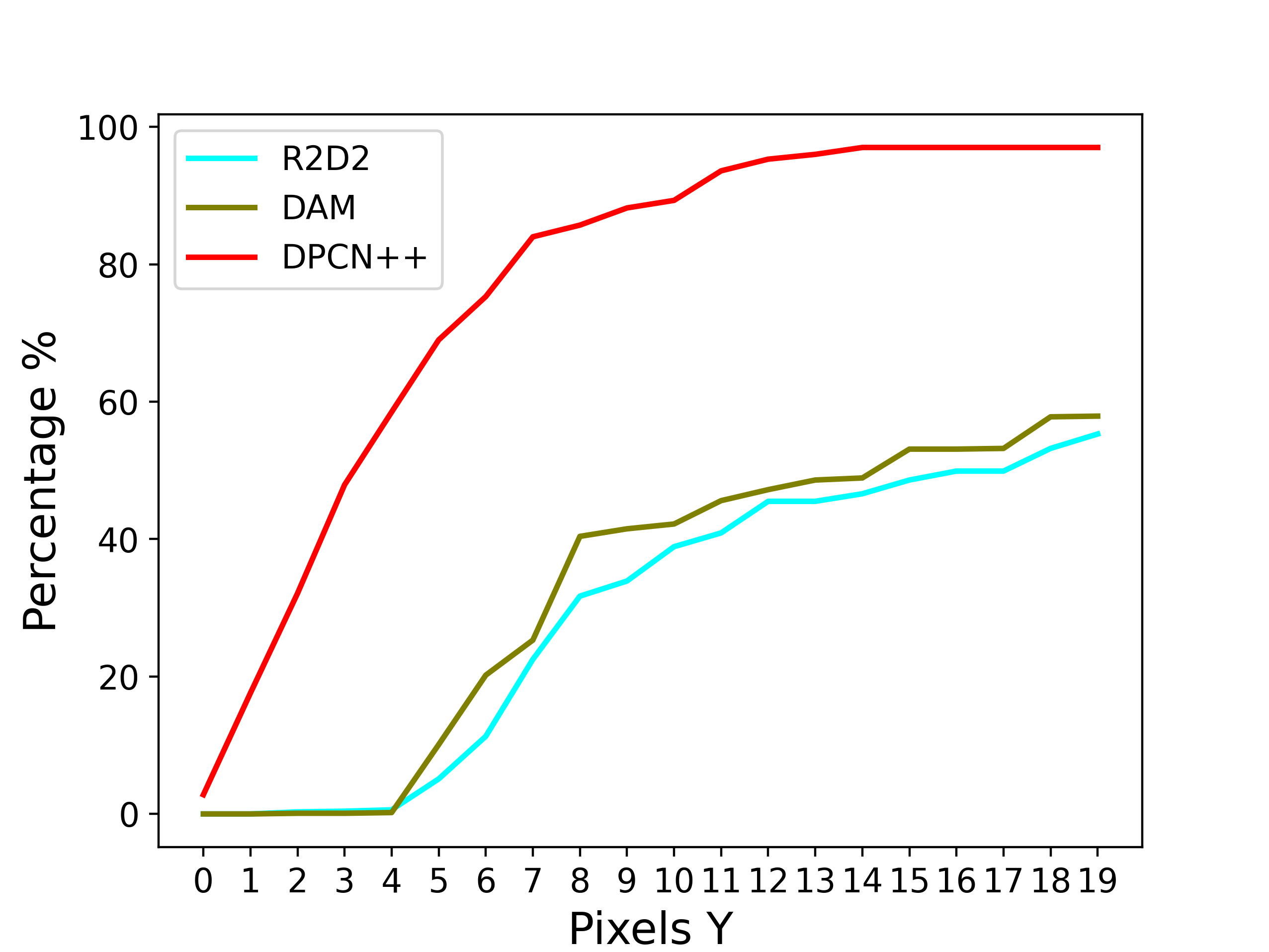}}
  \subfloat[Estimation of $x$ in ``Stereo to Drone" scene(b)\label{gym_s2a_x}]{%
        \includegraphics[width=0.25\linewidth]{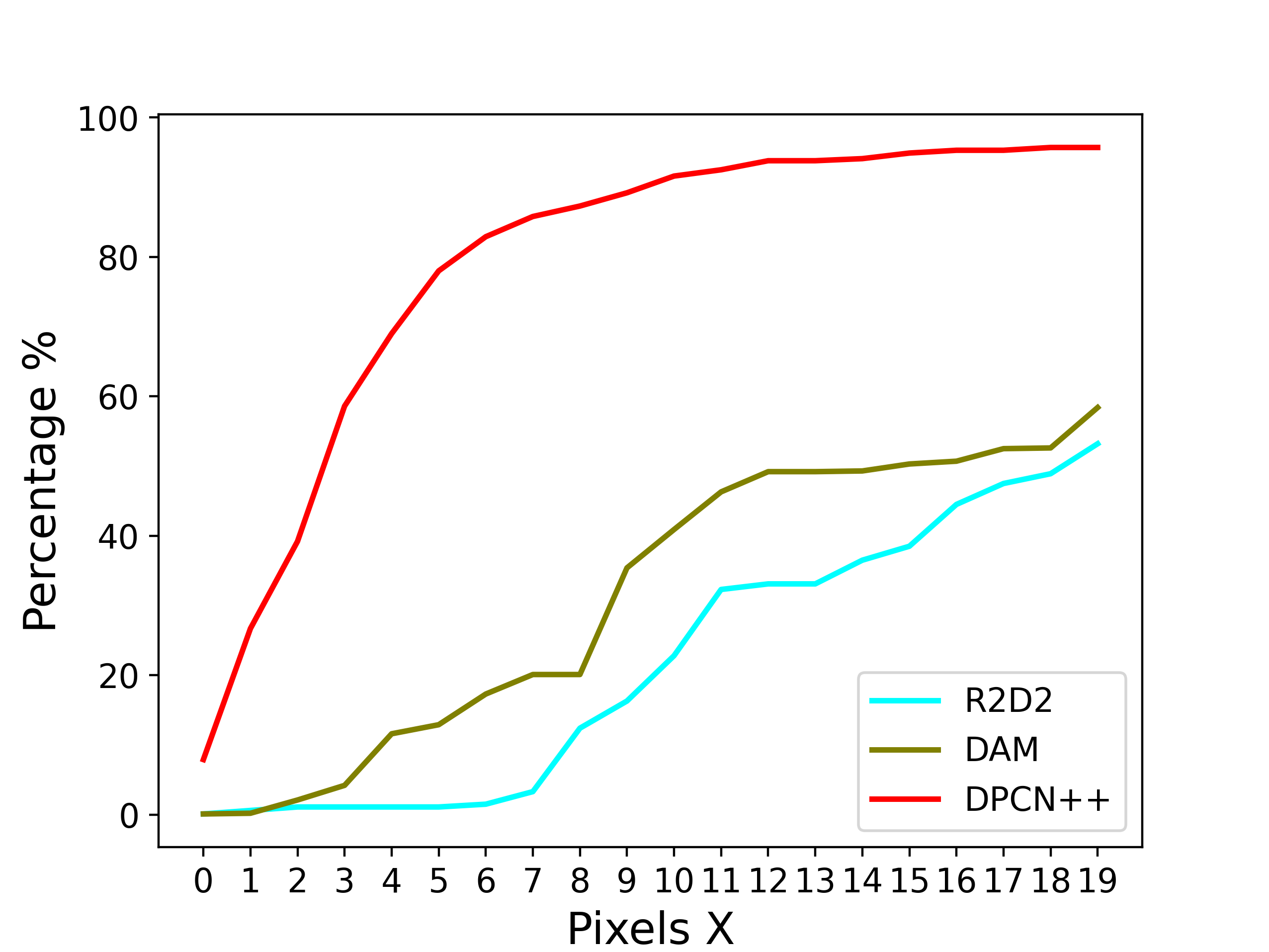}}
  \subfloat[Estimation of $y$ in ``Stereo to Drone" scene(b)\label{gym_s2a_y}]{%
        \includegraphics[width=0.25\linewidth]{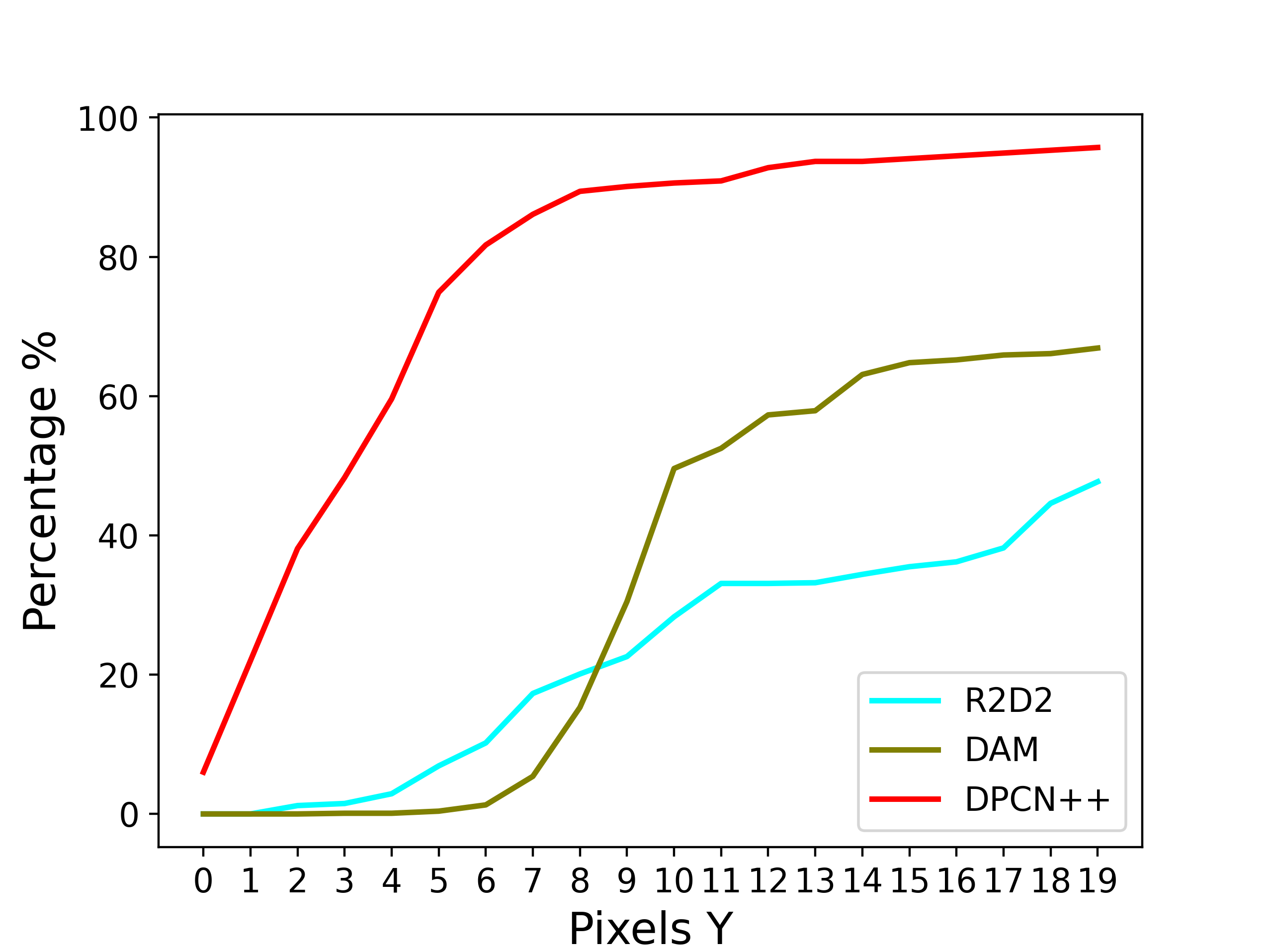}}
  \caption{$Acc_{0\:to\:19}$ of translation estimation in Aero-Ground dataset. }
  \label{Acc AG1}
\end{figure}

In this subsection, we give the results of evaluation with the MSE on AeroGround Dataset. The results are shown in Tab.~\ref{tab: append2d Aero-Ground Dataset}. Furthermore, the threshold of estimation in experiments is elaborated by the means of graphs. Figure \ref{Acc AG1} shows the $Acc_{0\:to\:19}$ of translation estimation in simulation dataset.

\begin{table*}[h]
% \captionsetup{justification=raggedright,singlelinecheck=false}
\caption{Results of the heterogeneous 2D registration in simulation and in scene (a) and (b) of the AG dataset. Note: ``l2sat'', ``l2d'', ``s2sat'', ``s2d'' are the abbreviation for 
    ``LiDAR Local Map" to ``Satellite Map",
    ``LiDAR Local Map" to ``Drone's Birds-eye Camera",
    ``Stereo Local Map" to ``Satellite Map",
    ``Stereo Local Map" to ``Drone's Birds-eye Camera", respectively.}
\label{tab: append2d Aero-Ground Dataset}
\renewcommand\arraystretch{1.15}
\centering
\resizebox{\textwidth}{!}{
\begin{tabular}{@{}ccllllllllll@{}}
\toprule[1pt]
Scene & \multicolumn{1}{l}{Method} & Exp. & $E_x$ & $Acc_{x_{10}}$ & $E_{y}$ & $Acc_{y_{10}}$ & $E_{r}$ & $Acc_{r_{1}}$ & $E_{\mu}$ & $Acc_{\mu_{0.2}}$ & Runtime(ms) \\ \midrule

\multirow{13}{*}{(a)} & R2D2~\cite{revaud2019r2d2}    & l2sat & 521.448 & 32.7 & 500.821 & 41.1 & 213.692 & 37.6 & 0.213 & 71.2 & 244.1 \\
                     &                      & l2d   & 569.729 & 39.1 & 488.721 & 45.7 & 238.579 & 30.1 & 0.228 & 73.5 & 244.1 \\
                     &                      & s2sat & 612.823 & 32.9 & 530.865 & 40.6 & 245.733 & 27.6 & 0.261 & 69.5 & 244.3 \\
                     &                      & s2d   & 503.763 & 41.7 & 480.554 & 41.9 & 193.838 & 32.9 & 0.207 & 75.9 & 244.2 \\
                     & DAM~\cite{Park2020}      & l2sat & \underline{507.194} & \underline{55.4} & \underline{208.866} & \underline{70.8} & \underline{44.2139} & \underline{37.8} & $\setminus$ & $\setminus$ & \underline{110.6} \\
                     &                      & l2d   & \underline{690.178} & \underline{39.4} & \underline{301.119} & \underline{66.8} & \underline{96.5603} & \underline{22.5} & $\setminus$ & $\setminus$ & \underline{117.3} \\
                     &                      & s2sat & \underline{740.688} & \underline{35.2} & \underline{732.416} & \underline{33.6} & \underline{105.1678} & \underline{24.1} & $\setminus$ & $\setminus$ & \underline{114.4} \\
                     &                      & s2d   & \underline{536.502} & \underline{51.5} & \underline{616.404} & \underline{43.9} & \underline{68.1288} & \underline{33.9} & $\setminus$ & $\setminus$ & \underline{114.2} \\
                     & \textbf{DPCN++}        & l2sat & \textbf{40.556} & \textbf{96.9} & \textbf{4.817} & \textbf{98.0} & \textbf{0.117} & \textbf{99.2} & \textbf{0.0034} & \textbf{95.5} & \textbf{24.75} \\
                     &        & l2d   & \textbf{15.530} & \textbf{98.2} & \textbf{6.453} & \textbf{94.0} & \textbf{0.041} & \textbf{99.2} & \textbf{0.012} & \textbf{94.2} & \textbf{26.37} \\
                     &                      & s2sat & \textbf{65.373} & \textbf{90.9} & \textbf{15.592} & \textbf{97.8} & \textbf{0.107} & \textbf{97.4} & \textbf{0.005} & \textbf{93.7} & \textbf{23.61} \\
                     &          & s2d   & \textbf{32.731} & \textbf{91.3} & \textbf{14.493} & \textbf{92.6} & \textbf{0.227} & \textbf{99.3} & \textbf{0.007} & \textbf{93.5} & \textbf{24.72} \\ \midrule
\multirow{6}{*}{(b)}
                     & R2D2~\cite{revaud2019r2d2}                 & l2d   & 1035.830 & 26.7 & 666.928 & 39.4 & 313.624 & 22.4 & 0.266 & 64.3 & 244.2 \\
                     &                      & s2d   & 941.489 & 22.8 & 531.762 & 28.3 & 294.753 & 25.1 & 0.239 & 66.8 & 244.0 \\
                     & DAM~\cite{Park2020}                   & l2d   & \underline{972.822} & \underline{30.1} & \underline{588.412} & \underline{42.2} & \underline{61.334} & \underline{35.1} & $\setminus$ & $\setminus$ & \underline{113.9} \\
                     &                      & s2d   & \underline{633.279} & \underline{40.9} & \underline{484.362} & \underline{49.6} & \underline{85.343} & \underline{27.4} & $\setminus$ & $\setminus$ & \underline{116.5} \\
                     & \textbf{DPCN++}        & l2d   & \textbf{8.004} & \textbf{96.2} & \textbf{102.359} & \textbf{89.2} & \textbf{0.005} & \textbf{99.7} & \textbf{0.001} & \textbf{99.7} & \textbf{24.51}\\
                     &                 & s2d   & \textbf{88.742} & \textbf{91.6} & \textbf{61.086} & \textbf{90.6} & \textbf{0.763} & \textbf{99.4} & \textbf{0.004} & \textbf{95.0} & \textbf{25.63} \\
\bottomrule[1pt]
\end{tabular}
}
\end{table*}

\subsection{Elaboration on 2D Registration: Generalization}
\label{subsec: append_Elaboration on 2D generalization}

\begin{figure*}[ht]
    \captionsetup[subfigure]{justification=centering}
    \centering
  \subfloat[Estimation of $x$ in simulation generalization\label{Generalization Sim_x}]{%
      \includegraphics[width=0.3\linewidth]{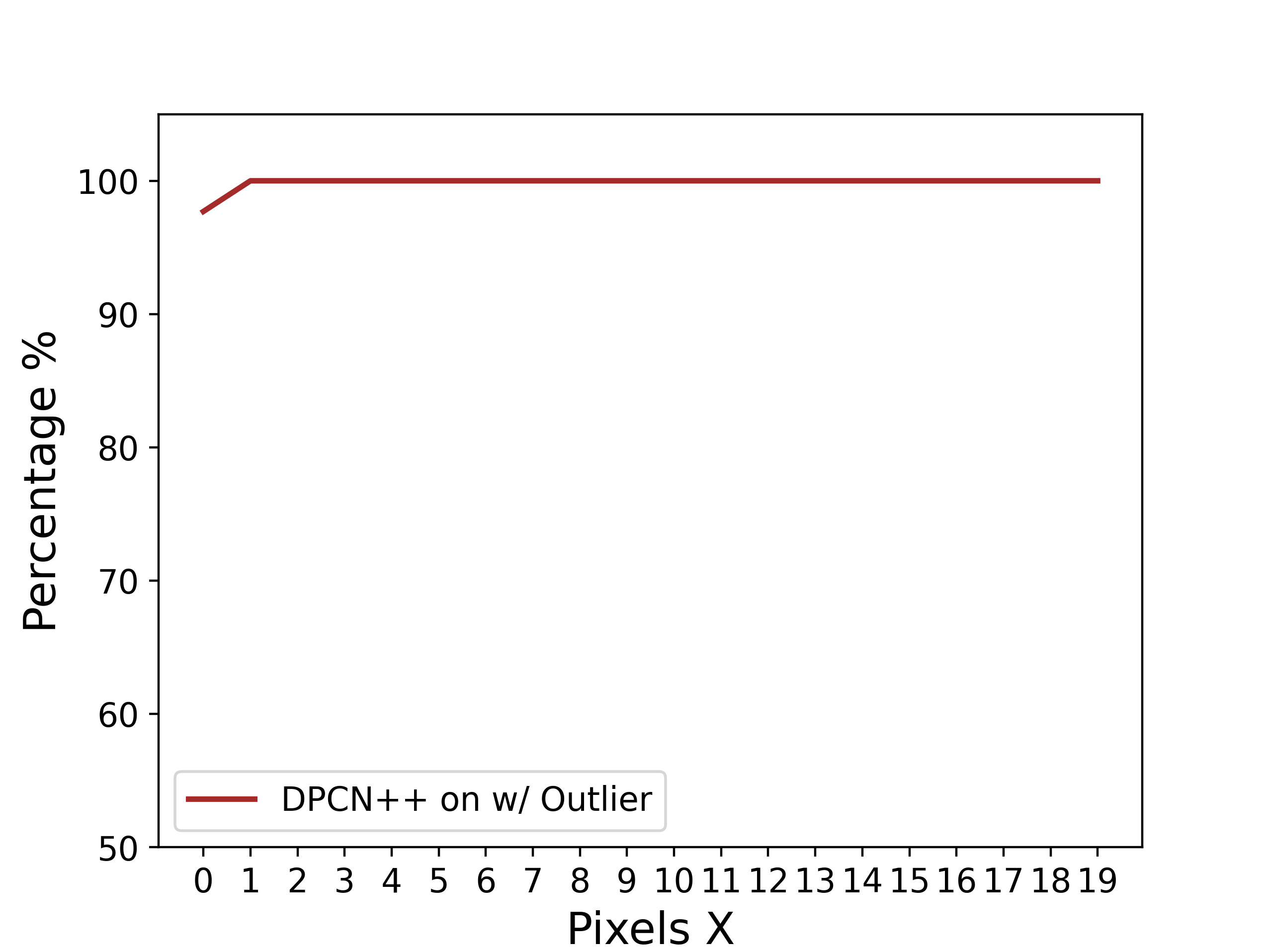}}
  \subfloat[Estimation of $y$ in simulation generalization\label{Generalization Sim_y}]{%
        \includegraphics[width=0.3\linewidth]{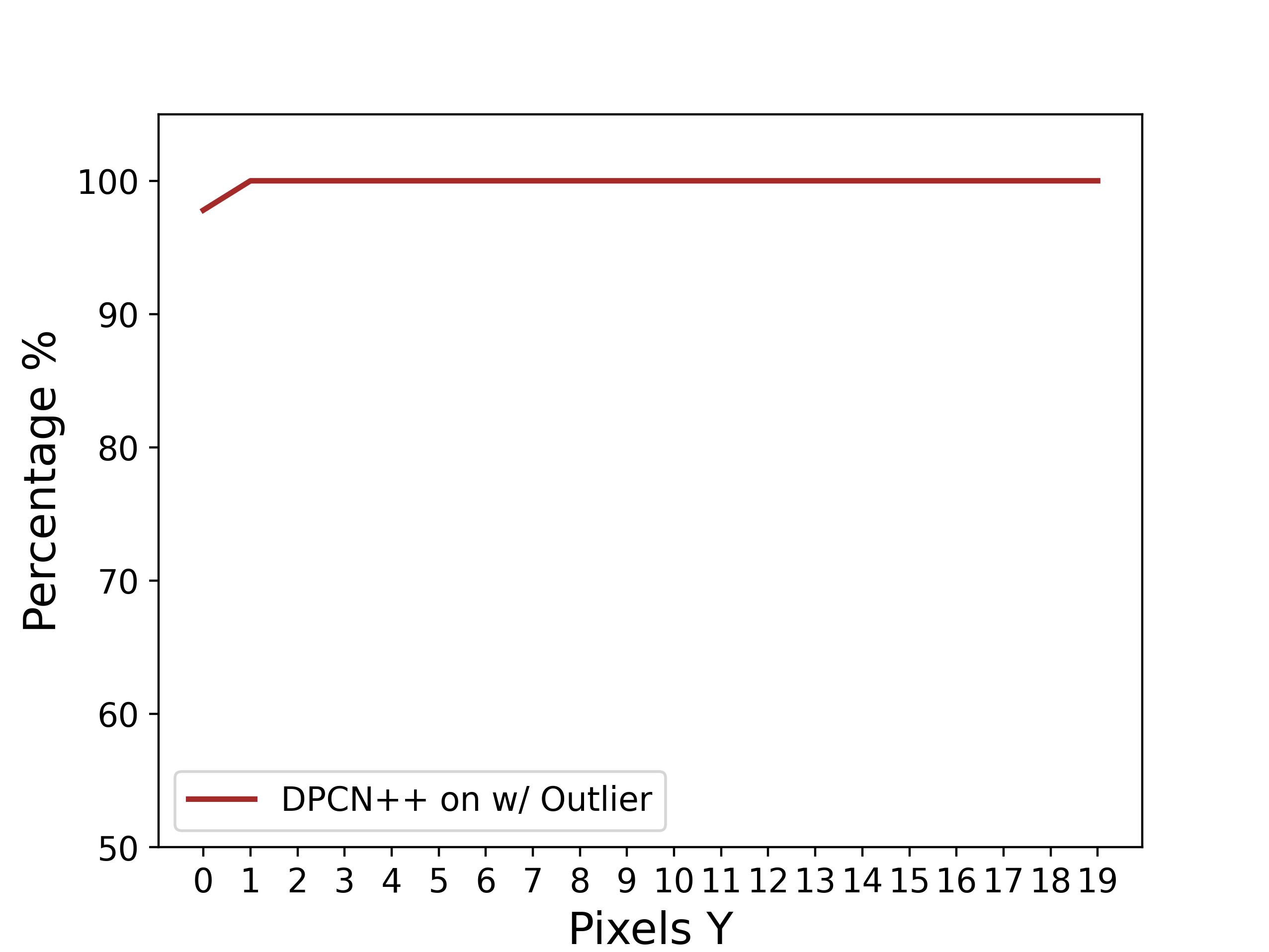}}

  \caption{$Acc_{0\:to\:19}$ of translation estimation in generalization on 2D simulation dataset. }
  \label{fig:append_Acc Generalization}
\end{figure*}

Tab.~\ref{tab: 2d_generalizationSimulation} and Fig.~\ref{fig:append_Acc Generalization} verify the generalizing ability of our approach in the simulation. The model is trained on the ``Heterogeneous" set and is evaluated on ``Heterogeneous w/ Outliers" in simulation datasets. The results show that with models not specifically trained, it can still maintain a high rate of accuracy in all $4DoF$.

\begin{table*}[h]
\centering
\renewcommand\arraystretch{1.2}
\captionsetup{justification=raggedright}
\caption{Results of generalization experiments with simulation dataset.}
\resizebox{0.9\textwidth}{!}{
\begin{tabular}{c ccccccccc ccccccccc }
\toprule[1pt]
Exp. & $E_x$ & $Acc_{x_5}(\%)$ & $E_y$ & $Acc_{y_5}(\%)$ & $E_{r}$ & $Acc_{r_{1}}(\%)$ & $E_{\mu}$ & $Acc_{\mu_{0.2}}(\%)$ \\ \midrule
Heterogeneous w/ Outlier  &  0.0276  & 100   &  0.105   &   100  &  0.039  &  100   &  0.003   &  100\\
\bottomrule[1pt]
\end{tabular}
}
\label{tab: 2d_generalizationSimulation}
\end{table*}

\begin{figure*}[ht]
    \captionsetup[subfigure]{justification=centering}
    \centering
  \subfloat[Estimation of $Acc_{x_{0\sim19}}$ \label{Generalization AG_x}]{%
        \includegraphics[width=0.33\linewidth]{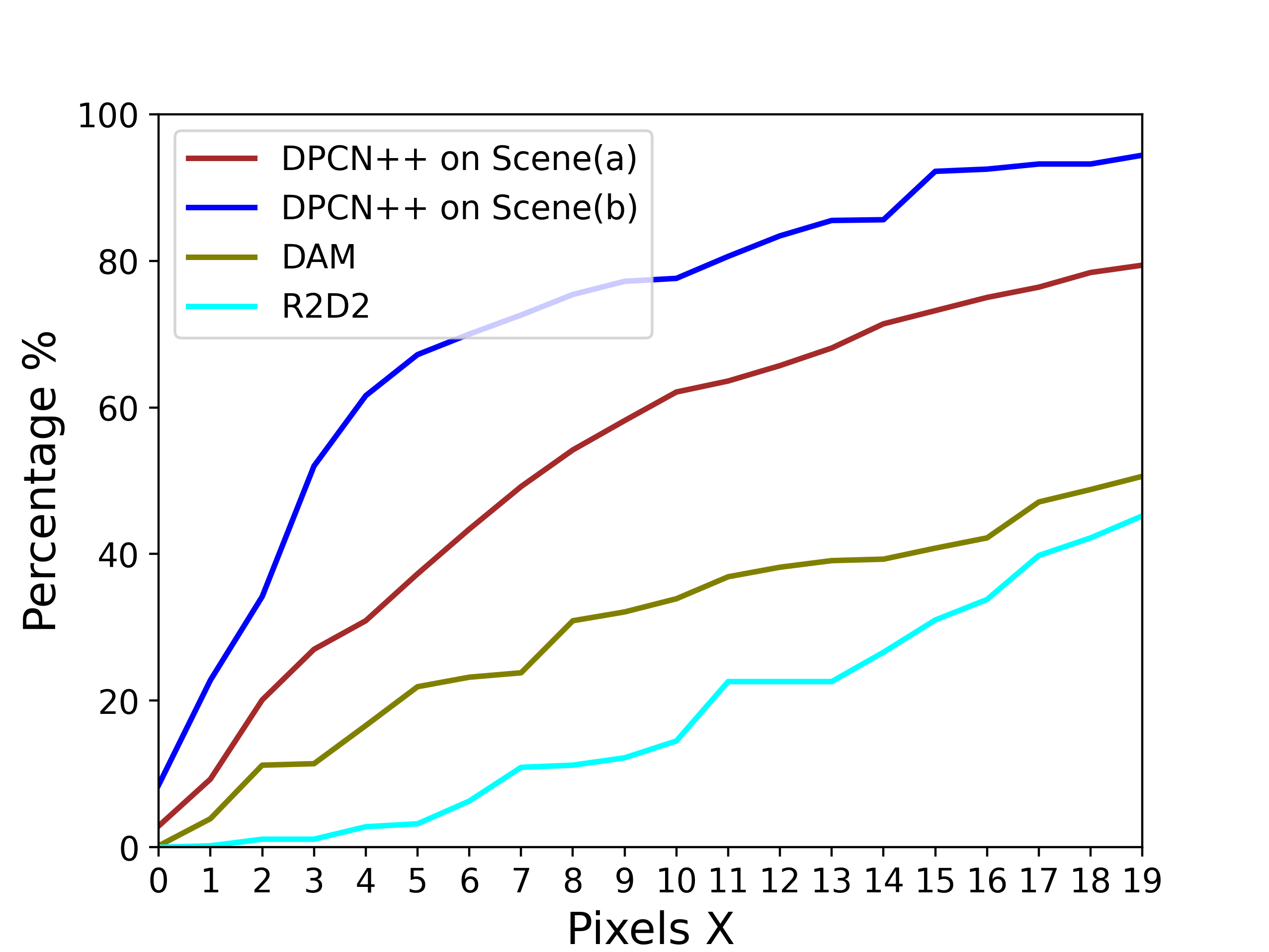}}
  \subfloat[Estimation of $Acc_{y_{0\sim19}}$\label{Generalization AG_y}]{%
        \includegraphics[width=0.33\linewidth]{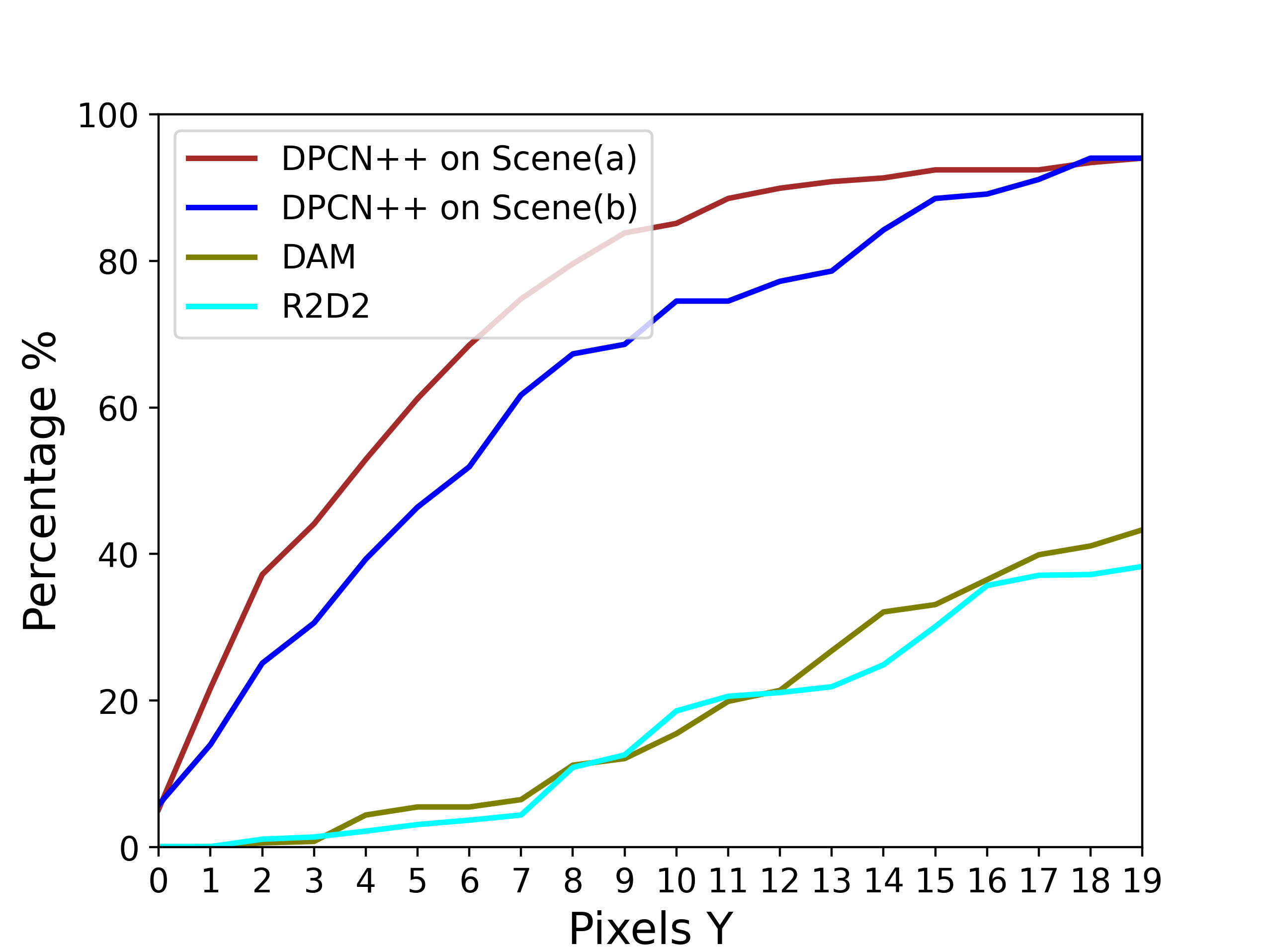}}
  \subfloat[Estimation of $Acc_{r_{0\sim19}}$\label{Generalization AG_rot}]{%
        \includegraphics[width=0.33\linewidth]{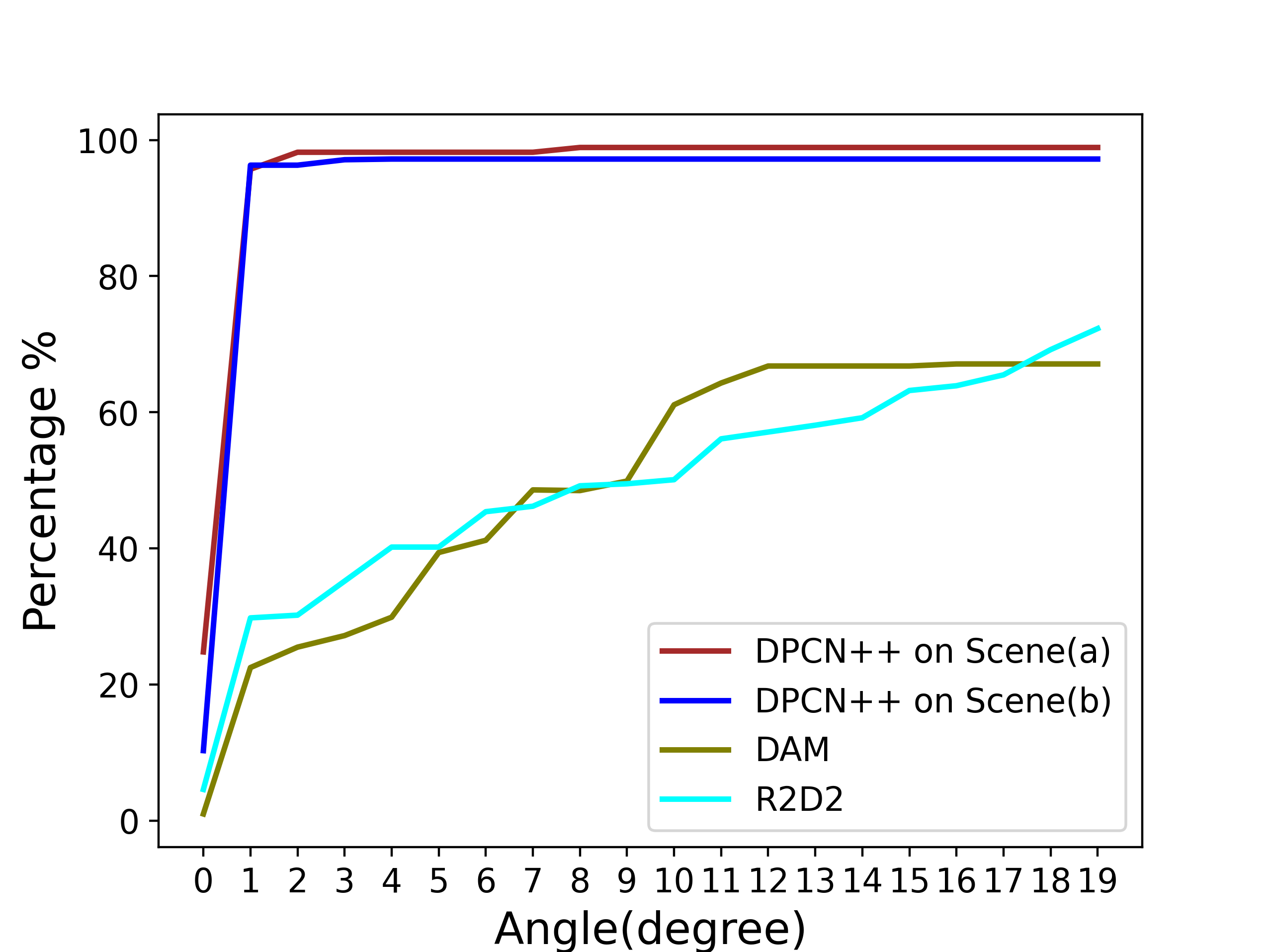}}
  \caption{Quantitative results of transformation estimation in generalization on scene (c) with baselines trained on scene (a) and scene (b).}
  \label{fig:AG Generalization}
  \vspace{0pt}
\end{figure*}

Tab. \ref{results:generalizationAG} and Fig.~\ref{fig:AG Generalization} show the generalization experiments on Aero-Ground Dataset. The models are trained in scene (a) and scene (b) and tested in scene (c).

\begin{table*}[ht]
\renewcommand\arraystretch{1.4}
\caption{Results of generalization experiments with Aero-Ground Dataset. Experiments are conducted with the input type of stereo camera and drone's birds-eye, therefore, the model applied in these experiments are trained on the ``s2d" dataset in scene (a) and (b). For generalization, we choose the threshold error of $15 \:pixels$ for translation, $1^{\circ}$ for rotation and $0.2\times$ for scale.}
\centering
\begin{small}
\resizebox{\textwidth}{!}{
\begin{tabular}{c ccccccccc ccccccccc }
\toprule[1pt]
Model & $E_x$ & $Acc_{x_{15}}(\%)$ & $E_y$ & $Acc_{y_{15}}(\%)$ & $E_{rot}$ & $Acc_{rot_{1}}(\%)$ & $E_{scale}$ & $Acc_{scale_{0.2}}(\%)$ \\ \midrule
DPCN in (a)       &  232.464  &  73.2   &  29.925   &  92.4   &  89.094  &  95.7   &  0.008  & 95.0\\
DPCN in (b)       &  31.207  &  92.2   &  138.545   &  88.5   &  2.879  &  96.3   &  0.015  & 93.3\\
DAM       &  602.8490 &  40.8   &  720.9244   &  33.1   &  88.724  &  22.5   &  $\setminus$  & $\setminus$\\
R2D2       &  873.648  &  31.0   &  922.831   &  30.1  &  194.382  &  29.8  &  0.251  & 59.7\\
\bottomrule[1pt]
\end{tabular}
}
\end{small}
\label{results:generalizationAG}
\end{table*}

\section{Additional Demonstration for 3D Registration}
\label{subsec: append_Additional Demonstration 3d}

\subsection{Failure Case in 3DMatch}
\label{subsec: append Visual on 3D Cases}
\begin{figure*}[t]
\centering
\includegraphics[width=0.8\textwidth]{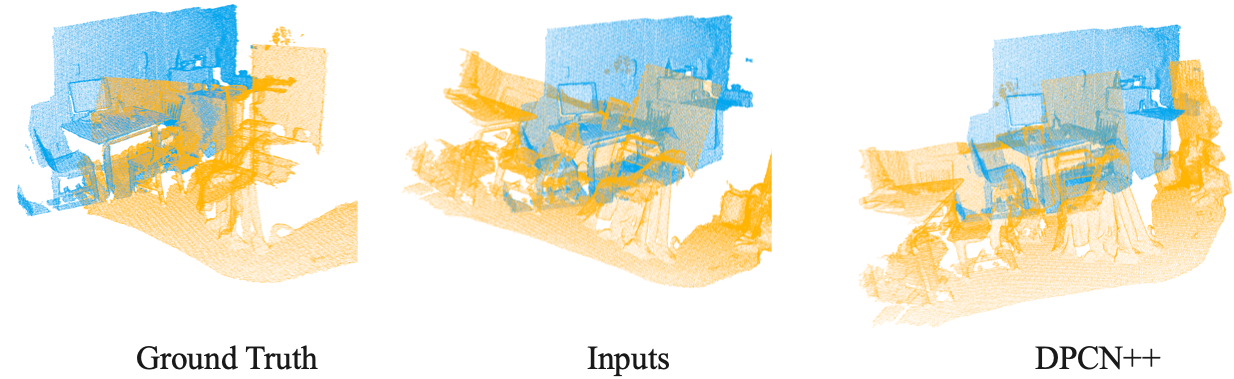}
\caption{One example of the failure cases in 3DMatch.}
\label{fig:3dmatch failure}
\end{figure*}

One particular case where DPCN++ fails in the 3DMatch dataset is showcased in Fig.~\ref{fig:3dmatch failure}.

\subsection{Quantitative Elaborations on 3DMatch}
\label{subsec: append Quant on 3DMatch}
\begin{figure*}[t]
\centering
\includegraphics[width=\textwidth]{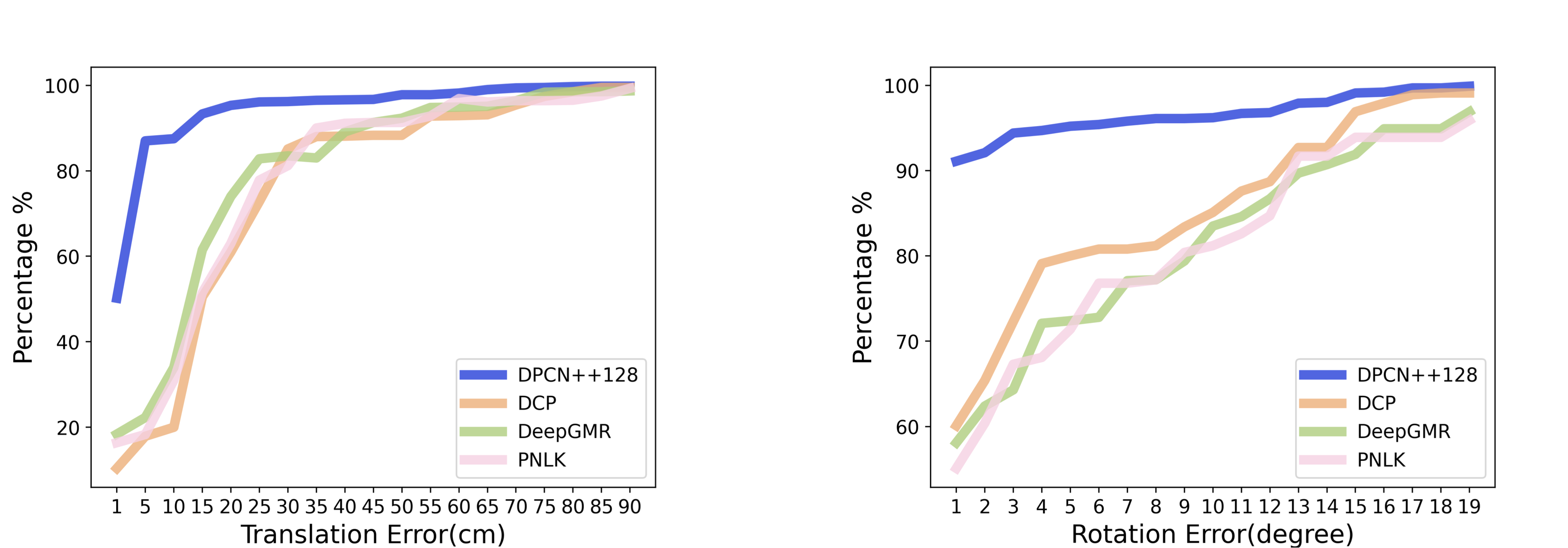}
\caption{The elaboration of the performances of the baselines on 3DMatch, when considering different performances on detailed thresholds.}
\label{fig:3dmatch_thesh}
\end{figure*}

We report the success rate considering different thresholds for baselines evaluated in the 3DMatch dataset in Fig.~\ref{fig:3dmatch_thesh}.

\subsection{Quantitative Elaborations on 3D Heterogeneous Registration}
\label{subsec: append Quant on mvp}
\begin{figure*}[t]
\centering
\includegraphics[width=\textwidth]{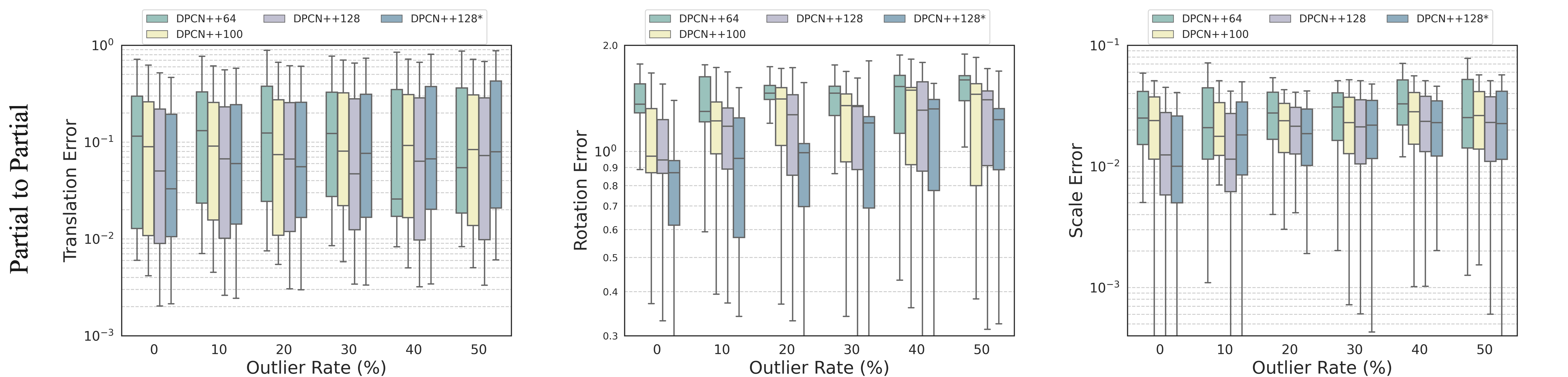}
\caption{The elaboration on qualitative performance point-mesh registration.}
\label{fig:point-mesh-p2p}
\end{figure*}

We further  The quantitative results on partial to partial point to mesh registration are shown in  Fig.~\ref{fig:point-mesh-p2p}.

\end{document}